\documentclass[twoside,11pt]{article}

\usepackage{blindtext}
\usepackage{multirow}
\usepackage{paralist,amsmath, bm}
\usepackage{algorithm,algorithmic}
\usepackage{dsfont}
            
\usepackage[preprint]{jmlr2e}

\usepackage{lastpage}
\jmlrheading{1}{2025}{1-\pageref{LastPage}}{XX/XX}{}{21-0000}{Lijun Zhang, Wenhao Yang, Guanghui Wang, Wei Jiang, Zhi-Hua Zhou}

\ShortHeadings{Universal Algorithms for Minimizing the Adaptive Regret of Convex Functions}{Zhang, Yang, Wang, Jiang, and Zhou}
\firstpageno{1}

\makeatletter
\AtBeginDocument{\Hy@breaklinkstrue}
\makeatother

\def \y {\mathbf{y}}

\def \EC {\mathcal{E}}
\def \x {\mathbf{x}}
\def \g {\mathbf{g}}

\def \u {\mathbf{u}}

\def \w {\mathbf{w}}
\def \R {\mathbb{R}}

\def \P {\mathcal{P}}

\def \G {\mathcal{G}}

\def \N {\mathbb{N}}
\def \A {\mathcal{A}}

\def \I {\mathcal{I}}

\def \wh {\widehat{\w}}

\def \S {\mathcal{S}}
\def \up {\Delta}
\newtheorem{ass}{Assumption}
\newtheorem{thm}{Theorem}

\DeclareMathOperator*{\Reg}{Regret}
\DeclareMathOperator*{\WAReg}{WA-Regret}
\DeclareMathOperator*{\SAReg}{SA-Regret}
\DeclareMathOperator*{\SAComReg}{Comp-SA-Regret}

\DeclareMathOperator*{\tr}{tr}

\begin{document}

\title{Dual Adaptivity: Universal Algorithms for Minimizing the Adaptive Regret of Convex Functions}

\author{\name Lijun Zhang\textsuperscript{\rm 1,2} \email zhanglj@lamda.nju.edu.cn 
       \AND
       \name Wenhao Yang\textsuperscript{\rm 1,2} \email yangwh@lamda.nju.edu.cn 
       \AND
       Guanghui Wang\textsuperscript{\rm 3} \email gwang369@gatech.edu 
       \AND 
       Wei Jiang\textsuperscript{\rm 1} \email jiangw@lamda.nju.edu.cn 
       \AND
       Zhi-Hua Zhou\textsuperscript{\rm 1,2} \email zhouzh@nju.edu.cn \\
       \addr \textsuperscript{\rm 1}National Key Laboratory for Novel Software Technology, Nanjing University, Nanjing, China \\
       \textsuperscript{\rm 2}School of Artificial Intelligence, Nanjing University, Nanjing, China \\
       \textsuperscript{\rm 3}College of Computing, Georgia Institute of Technology, Atlanta, USA
}

\editor{My editor}

\maketitle

\begin{abstract}
To deal with changing environments, a new performance measure---adaptive regret, defined as the maximum static regret over any interval, was  proposed in online learning. Under the setting of online convex optimization, several algorithms have been successfully developed to minimize the adaptive regret. However, existing algorithms lack universality in the sense that they can only handle one type of convex functions and need apriori knowledge of parameters, which hinders their application in real-world scenarios. To address this limitation, this paper investigates universal algorithms with dual adaptivity, which automatically adapt to the property of functions (convex, exponentially concave, or strongly convex), as well as the nature of environments (stationary or changing). Specifically, we propose a meta-expert framework for dual adaptive algorithms, where multiple experts are created dynamically and aggregated by a meta-algorithm. The meta-algorithm is required to yield a second-order bound, which can accommodate unknown function types. We further incorporate the technique of sleeping experts to capture the changing environments. For the construction of experts, we introduce two strategies (increasing the number of experts or enhancing the capabilities of  experts) to achieve universality. Theoretical analysis shows that our algorithms are able to minimize the adaptive regret for multiple types of convex functions simultaneously, and also allow the type of functions to switch between rounds. Moreover, we extend our meta-expert framework to online composite optimization, and develop a universal algorithm for minimizing the adaptive regret of  composite functions. 
\end{abstract}

\begin{keywords}
Online Convex Optimization, Adaptive Regret, Strongly Convex Functions, Exponentially Concave Functions, Online Composite Optimization
\end{keywords}

\section{Introduction}

Online learning aims to make a sequence of accurate decisions given knowledge of answers to previous tasks and possibly additional information \citep{Online:suvery}. It is performed in a sequence of consecutive rounds, where at round $t$ the learner is asked to select a decision $\w_t$ from a domain $\Omega$. After submitting the answer, a loss function $f_t:\Omega \mapsto \R$ is revealed and the learner suffers a loss $f_t(\w_t)$. The standard performance measure is the regret \citep{bianchi-2006-prediction}:
\begin{equation*}
    \Reg(T)=\sum_{t=1}^T f_t(\w_t) - \min_{\w \in \Omega} \sum_{t=1}^T f_t(\w)
\end{equation*}
defined as the difference between the cumulative loss of the online learner and that of the best decision chosen in hindsight. When both the domain $\Omega$ and the loss $f_t(\cdot)$ are convex, it becomes online convex optimization (OCO) \citep{zinkevich-2003-online}.

In the literature, there exists plenty of algorithms to minimize the regret under the setting of OCO \citep{Intro:Online:Convex}. However, when the environment is non-stationary, regret may not be the best performance measurement. That is because regret chooses a fixed comparator, and for the same reason, it is also referred to as \emph{static} regret. To avoid this limitation,  \citet{Adaptive:Hazan} introduce the concept of adaptive regret, which measures the performance with respect to a changing comparator. Later, \citet{Adaptive:ICML:15} propose a refined notation---strongly adaptive regret, defined as the maximum static regret over intervals of length $\tau$: 
\begin{equation} \label{eqn:strong:adaptive}
\SAReg(T,\tau) = \max_{[p, p+\tau -1] \subseteq [T]} \left(\sum_{t=p}^{p+\tau -1}  f_t(\w_t)  - \min_{\w \in \Omega} \sum_{t=p}^{p+\tau -1} f_t(\w) \right).
\end{equation}
Since the seminal work of \citet{Adaptive:Hazan}, several algorithms have been successfully developed to attain $O(\sqrt{\tau \log T})$, $O(d\log \tau \log T)$ and $O(\log \tau \log T)$ strongly  adaptive regret  for general convex, exponentially concave (abbr.~exp-concave) and strongly convex functions \citep{Hazan:2009:ELA,Improved:Strongly:Adaptive,Dynamic:Regret:Adaptive} respectively, where $d$ is the dimensionality. However, existing methods can only handle one type of convex functions. Furthermore, when facing exp-concave functions and strongly convex functions, they need to know the moduli of exp-concavity and strong convexity. The lack of universality hinders their application to real-world problems. 

On the other hand, there do exist universal methods for OCO, such as MetaGrad \citep{NIPS2016_6268} and USC \citep{ICML:2022:Zhang}, that attain optimal static regret for multiple types of convex functions simultaneously. This observation motivates us to ask whether it is possible to design a single algorithm to minimize the adaptive regret of multiple types of convex functions, which means that the algorithm needs to enjoy \emph{dual adaptivity}, adaptive to the function type and adaptive to the environment. In this paper, we provide an affirmative answer by proposing a meta-expert framework for dual adaptive algorithms, as detailed below. 

\paragraph{The Meta-expert Framework for Dual Adaptivity.} Our proposed meta-expert framework contains $3$ key components: 
\begin{compactitem}
  \item Expert-algorithms, which are able to minimize the static regret;
  \item A set of intervals, each of which is associated with one or multiple experts that minimize the regret of that interval;
  \item A meta-algorithm, which combines the predictions of active experts in each round.
\end{compactitem}
Inspired by the recent development of universal algorithms for static regret \citep{ICML:2022:Zhang}, we choose Adapt-ML-Prod \citep{pmlr-v35-gaillard14} as the meta-algorithm, and extend it to support sleeping experts---experts that are active only during specific periods. The resulting meta-algorithm achieves a second-order bound, allowing it to automatically exploit the properties of functions and attain small meta-regret. Following prior work~\citep{Adaptive:ICML:15}, we employ geometric covering (GC) intervals to define the lifetimes of experts. To construct experts operating on these intervals, we propose two strategies: the first increases the number of experts, while the second enhances their  capabilities. In the following, we describe both types of algorithms. 

\paragraph{A Two-layer Universal Algorithm} In the first strategy, we introduce a two-layer Universal algorithm for Minimizing the Adaptive regret (UMA$2$). Compared to existing adaptive algorithms, we create a \emph{larger} set of experts over each interval to handle the uncertainty of the type of functions and (possibly) the associated parameters. The decisions of experts are then aggregated using the aforementioned meta-algorithm, forming a two-layer architecture. Notably, although our meta-algorithm is inspired by \citet{ICML:2022:Zhang}, the construction of experts is substantially different. Specifically, we introduce surrogate losses parameterized by distinct learning rates~\citep{NIPS2016_6268}, which are minimized by individual experts, in contrast to their method, where each expert directly optimizes the original loss. As a result, our approach eliminates the need for multiple gradient estimations and avoids the assumption on bounded parameters. Theoretical analysis shows that UMA$2$ can minimize the adaptive regret of general convex functions, and automatically take advantage of easier functions whenever possible. Specifically, UMA$2$ attains $O(\sqrt{\tau \log T})$, $O(\frac{d}{\alpha}\log \tau \log T)$ and $O(\frac{1}{\lambda}\log \tau \log T)$ strongly adaptive regret for general convex, $\alpha$-exp-concave and $\lambda$-strongly convex functions respectively, where $d$ is the dimensionality. All of these bounds match state-of-the-art results on adaptive regret \citep{Improved:Strongly:Adaptive,Dynamic:Regret:Adaptive} exactly.  Furthermore, UMA$2$ can also handle the case that the type of functions changes between rounds. For example, suppose the online functions are general convex during interval $I_1$, then become $\alpha$-exp-concave in  $I_2$, and finally switch to $\lambda$-strongly convex in $I_3$. When facing this function sequence, UMA$2$ achieves $O(\sqrt{|I_1|\log T})$, $O(\frac{d}{\alpha}\log |I_2| \log T)$ and $O(\frac{1}{\lambda}\log |I_3| \log T)$ regret over intervals $I_1$, $I_2$ and $I_3$, respectively.

\paragraph{A Three-layer Universal Algorithm} In the second strategy, we develop a three-layer Universal algorithm for Minimizing the Adaptive regret (UMA$3$). Unlike existing adaptive algorithms that rely on single-purpose experts, we enhance the capability of the expert, enabling it to handle a broader class of convex functions. Specifically, we use Maler~\citep{Adaptive:Maler}, an existing universal method for minimizing the static regret, as the expert-algorithm. Then, we apply the same meta-algorithm as UMA$2$ to dynamically aggregate experts’ decisions. Since Maler itself is a two-layer algorithm, our approach forms a three-layer architecture. In contrast to UMA$2$, UMA$3$ treats the existing universal algorithm as a black-box subroutine, thereby simplifying both the algorithm design and the theoretical analysis. It achieves the same order of strongly adaptive regret bounds as UMA$2$, and also allows the type of functions to switch between rounds. 

\paragraph{Online Composite Optimization}  We further investigate online composite optimization, where the loss function $F_t(\w)\triangleq f_t(\w)+r(\w)$ is defined as the sum of a time-varying function $f_t(\cdot)$ and a fixed convex regularizer $r(\cdot)$. Our goal is to design a universal algorithm for minimizing the adaptive regret in terms of  composite functions:
\begin{equation}\label{eqn:SA-Regret-Comp}
 \SAComReg(T,\tau) = \max_{[p, p+\tau -1] \subseteq [T]} \left(\sum_{t=p}^{p+\tau -1}  F_t(\w_t)  - \min_{\w \in \Omega} \sum_{t=p}^{p+\tau -1} F_t(\w) \right).   
\end{equation}
To this end, a straightforward idea is to directly pass the composite function $F_t(\w)$ to UMA$2$ or UMA$3$. However, this approach cannot attain tight adaptive regret for exp-concave functions, as the sum of an exp-concave function and a convex regularizer does not preserve exp-concavity \citep{AISTATS:2018:Yang}. To address this problem, we develop a meta-expert framework for online composite optimization, which uses Optimistic-Adapt-ML-Prod \citep{NIPS:2016:Wei} as the meta-algorithm. Following the optimism setting of \citet{zhang2024universal}, we show that our framework can yield second-order bounds in terms of the time-varying functions. To handle diverse function classes, we can either employ a large number of specialized experts or a small number of more powerful ones. For simplicity, we adopt the latter method by leveraging universal algorithms for composite functions as experts. Since the existing method~\citep{zhang2024universal} relies on the assumption of bounded moduli,  we propose a novel universal method for composite functions that avoids this constraint. By deploying an expert on each interval, our algorithm achieves $O(\sqrt{\tau \log T})$, $O(\frac{d}{\alpha}\log \tau \log T)$ and $O(\frac{1}{\lambda}\log \tau \log T)$ strongly adaptive regret for three types of convex $f_t(\cdot)$ respectively in the composite setting.

\paragraph{Comparisons with the Conference Version} A preliminary version of this paper, published at the 35th Annual Conference on Neural Information Processing System \citep{NEURIPS2021_d1588e68}, developed a two-layer algorithm as an extension of MetaGrad. In this paper, we have significantly enriched the preliminary version in the following three aspects: 
\begin{compactitem}
  \item \textbf{The Meta-algorithm:} While the conference version uses TEWA~\citep{NIPS2016_6268} as the meta-algorithm, we adopt an algorithm with a second-order regret bound to serve this role. The meta-algorithm in this paper is more flexible, as it allows experts to  operate on either original  or surrogate losses. Furthermore, the meta-algorithm offers the advantage of adapting to other online settings, such as online composite optimization. 
  \item \textbf{Constructions of Experts:} The preliminary version only increases the number of experts to handle the uncertainty of functions. In  contrast, we propose two strategies for constructing experts in this paper: increasing the number of experts (two-layer algorithms) or enhancing their capabilities (three-layer algorithms). 
  \item \textbf{Extensions to Online Composite Optimization:} We  extend our meta-expert framework to support composite functions by choosing Optimistic-Adapt-ML-Prod as the meta-algorithm. First, we develop a novel universal method for static regret of composite functions, which removes the assumption on bounded moduli imposed in \citet{zhang2024universal}. Second, by employing this method as the expert-algorithm, we introduce a universal algorithm for adaptive regret of composite functions. 
\end{compactitem}

\paragraph{Organization} The rest is organized as follows. Section~\ref{sec:related}  review related work. Section~\ref{sec:framewok} presents our meta-expert framework for dual adaptive algorithms. Section~\ref{sec:UMA2+3} introduces the specific universal algorithms for minimizing the adaptive regret. Section~\ref{sec:Comp} extends our meta-expert framework to online composite optimization. Section~\ref{sec:analysis} presents the analysis of all theorems and lemmas. Section~\ref{sec:conclusion} concludes this paper and discusses future work. 

\section{Related Work}\label{sec:related}
In this section, we briefly review related work in OCO, including static regret, adaptive regret, and online composite optimization. 

\subsection{Static Regret}
To minimize the static regret of general convex functions, online gradient descent (OGD) with step size $\eta_t=O(1/\sqrt{t})$ achieves an  $O(\sqrt{T})$  regret bound \citep{zinkevich-2003-online}. If all the online functions are $\lambda$-strongly convex, OGD with step size $\eta_t=O(1/[\lambda t])$ attains an $O(\frac{1}{\lambda} \log T)$ bound \citep{ICML_Pegasos}. When the functions are $\alpha$-exp-concave, online Newton step (ONS), with knowledge of $\alpha$, enjoys an $O(\frac{d}{\alpha} \log T)$ bound, where $d$ is the dimensionality \citep{ML:Hazan:2007}.  These regret bounds are minimax optimal for the corresponding types of functions \citep{Lower:bound:Portfolio,Minimax:Online}, but choosing the optimal algorithm for a specific problem requires domain knowledge.

The study of universal algorithms for OCO stems from the adaptive online gradient descent (AOGD) \citep{NIPS2007_3319} and its proximal extension \citep{icml2009_033}. The key idea of AOGD is to add a quadratic regularization term to the loss. It has been proven that AOGD is able to interpolate between the $O(\sqrt{T})$  regret bound of general convex functions and the $O(\log T)$ regret bound of strongly convex functions. Furthermore, it allows the online function to switch between general convex and strongly convex. However, AOGD has two restrictions:
\begin{compactitem}
  \item It needs to calculate the modulus of strong convexity on the fly, which is a nontrivial task. 
  \item It does not support exp-concave functions explicitly, and thus can only achieve suboptimal $O(\sqrt{T})$ regret for this type of functions.
\end{compactitem}

Another milestone is the multiple eta gradient algorithm (MetaGrad) \citep{NIPS2016_6268,pmlr-v99-mhammedi19a,JMLR:v22:20-1444}, which adapts to a much broader class of functions, including convex functions and exp-concave functions. MetaGrad's main feature is that it simultaneously considers multiple learning rates  and  does not need to know the modulus of exp-concavity.
MetaGrad achieves $O(\sqrt{T \log\log T})$ and $O(\frac{d}{\alpha} \log T)$ regret bounds for general convex and $\alpha$-exp-concave functions, respectively. However, MetaGrad treats strongly convex functions as exp-concave, and thus only gives  suboptimal $O(\frac{d}{\lambda} \log T)$ regret for $\lambda$-strongly convex functions. To address this problem, \citet{Adaptive:Maler} develop a universal algorithm named as multiple sub-algorithms and learning rates (Maler). It attains $O(\sqrt{T})$, $O(\frac{d}{\alpha} \log T)$ and $O(\frac{1}{\lambda} \log T)$ regret for general convex, $\alpha$-exp-concave, and $\lambda$-strongly convex functions, respectively. Furthermore, \citet{AAAI:2020:Wang} extend Maler to make use of smoothness.

Most of universal algorithms discussed above require constructing surrogate losses specifically for the expert-algorithms. \citet{ICML:2022:Zhang} present a simple strategy that does not need surrogate losses. In particular, their universal algorithm allows experts to operate on the original loss functions, while a meta-algorithm is applied to the \emph{linearized} losses. Crucially, the meta-algorithm is required to yield a second-order bound to automatically exploit strong convexity and exp-concavity. Based on this framework, \citet{NeurIPS:2024:Yang} proposed a projection-efficient universal algorithm, reducing the number of projections from $O(\log T)$ to $1$ per round. 

\subsection{Adaptive Regret} 
Adaptive regret has been studied in the setting of prediction with expert advice  \citep{LITTLESTONE1994212,Freund:1997:UCP,Adamskiy2012,Track_Large_Expert,pmlr-v40-Luo15} and OCO \citep{Adaptive:Hazan,Adaptive:ICML:15,Improved:Strongly:Adaptive}. In this section, we focus on the related work in the latter one.

Adaptive regret is firstly introduced by \citet{Adaptive:Hazan}, and later refined  by \citet{Adaptive:ICML:15}. To distinguish between them,
we refer to the definition of  \citeauthor{Adaptive:Hazan} as weakly adaptive regret:
\[
\WAReg(T)=  \max_{[p, q] \subseteq [T]} \left(\sum_{t=p}^{q} f_t(\w_t) - \min_{\w \in \Omega} \sum_{t=p}^{q} f_t(\w)\right).
\]
For $\alpha$-exp-concave functions, \citet{Adaptive:Hazan} propose an adaptive algorithm named as Follow-the-Leading-History (FLH). FLH restarts a copy of ONS in each round as an expert, and chooses the best one using expert-tracking algorithms. The meta-algorithm used to track the best expert is inspired by the Fixed-Share algorithm \citep{Herbster1998}. While FLH is equipped with $O(\frac{d}{\alpha} \log T)$ weakly adaptive regret, it is computationally expensive since it needs to maintain $t$ experts in the $t$-th iteration. To reduce the computational cost,  \citet{Adaptive:Hazan} further prune the number of experts based on a data streaming algorithm. In this way, FLH only keeps $O(\log t)$ experts, at the price of $O(\frac{d}{\alpha} \log^2 T)$ weakly adaptive regret.  Notice that the efficient version of FLH essentially creates and removes experts dynamically. As pointed out by \citet{Adamskiy2012}, this behavior can be modeled by the sleeping expert setting \citep{Freund:1997:UCP}, in which the expert can be ``asleep'' for certain rounds and does not make any advice.

For general convex functions, we can use OGD as the expert-algorithm in FLH. \citet{Adaptive:Hazan} prove that FLH and its efficient variant attain $O(\sqrt{T \log T})$ and $O(\sqrt{T \log^3 T})$ weakly adaptive regret, respectively. This result reveals a limitation of weakly adaptive regret---it does not respect short intervals well. For example, the $O(\sqrt{T \log T})$ regret bound is meaningless for intervals of length $O(\sqrt{T})$. To address this limitation, \citet{Adaptive:ICML:15} introduce the strongly adaptive regret  which takes the interval length as a parameter, as shown in (\ref{eqn:strong:adaptive}), and propose a novel algorithm named as Strongly Adaptive Online Learner (SAOL).  SAOL carefully constructs a set of intervals, then runs an instance of low-regret algorithm in each interval as an expert, and finally combines active experts' outputs by a variant of multiplicative weights method \citep{v008a006}. SAOL also maintains $O(\log t)$ experts in the $t$-th round, and achieves $O( \sqrt{\tau} \log T )$ strongly adaptive regret for convex functions.  Later, \citet{Improved:Strongly:Adaptive} develop a new meta-algorithm named as sleeping coin betting (SCB), and improve the strongly adaptive regret bound to $O(\sqrt{\tau \log T})$. \citet{pmlr-v119-cutkosky20a} has established problem-dependent bounds for strongly adaptive regret, which can guarantee the $O(\sqrt{\tau \log T})$ rate in the worst case, while achieving tighter results when the square norms of gradients are small. When we have prior knowledge about the change of environments, it is also possible to improve the logarithmic factor in the adaptive regret \citep{Adaptive:Short}.

For $\lambda$-strongly convex functions,  \citet{Dynamic:Regret:Adaptive} point out that we can replace ONS with OGD, and obtain $O(\frac{1}{\lambda} \log T)$ weakly adaptive regret. They also demonstrate that the number of active experts can be reduced from $t$ to $O(\log t)$, at a cost of an additional $\log T$ factor in the regret. All the aforementioned adaptive algorithms need to query the gradient of the loss function at least $\Theta(\log t)$ times in the $t$-th iteration. Based on surrogate losses, \citet{Adaptive:One:Gradient} show that the number of gradient evaluations per round can be reduced to $1$ without affecting the performance.

\subsection{Online Composite Optimization} 
Under the setting of online composite optimization, the online learner suffers a composite loss in each round $t$, which is formulated as: 
\begin{equation}\label{eqn:composite}
    F_t(\w)=f_t(\w)+r(\w),
\end{equation}
where $f_t(\cdot)\colon \Omega \rightarrow \R$ is a time-varying function, and $r(\cdot) \colon \Omega \rightarrow \R$ is a fixed convex regularizer, such as the $\ell_1$-norm for sparse vectors \citep{tibshirani1996regression} and the trace norm for low-rank matrices \citep{toh2010accelerated}. 

In the literature, there has been extensive explorations into minimizing the static regret of composite functions. Pioneering work \citep{DBLP:journals/jmlr/DuchiS09} proposes the forward backward splitting (FOBOS) method to achieve $O(\sqrt{T})$ and $O(\frac{1}{\lambda}\log T)$ regret bounds for general convex and $\lambda$-strongly convex $f_t(\w)$, respectively. Subsequently, \citet{DBLP:conf/nips/Xiao09} introduces the regularized dual averaging (RDA) method, which achieves regret bounds of the same order as those of FOBOS. Later, \citet{DBLP:conf/colt/DuchiSST10} propose a
generalized version of FOBOS, named as composite objective mirror descent (COMID). When the time-varying function $f_t(\w)$ is $\alpha$-exp-concave, \citet{DBLP:journals/ijon/YangTCWS24} develop the proximal online Newton step (ProxONS) to attain an $O(\frac{d}{\alpha}\log T)$ regret bound. Very recently, \citet{zhang2024universal} extend their universal strategy to support online composite optimization by choosing Optimistic-Adapt-ML-Prod as the meta-algorithm and setting appropriate parameters. Their algorithm achieves $O(\sqrt{T})$, $O(\frac{1}{\lambda}\log T)$ and $O(\frac{d}{\alpha}\log T)$ regret bounds for general convex $f_t(\cdot)$, $\lambda$-strongly convex $f_t(\cdot)$, and $\alpha$-exp-concave $f_t(\cdot)$, respectively. However, existing methods, which primarily focus on minimizing the static regret, are unable to deal with changing environments. Therefore, designing algorithms for minimizing the adaptive regret of online composite optimization remains open. 

\section{The Meta-expert Framework}\label{sec:framewok}
In this section, we first introduce necessary assumptions and definitions. Then, we outline a meta-expert framework for universal algorithms that minimize the adaptive regret. 

\subsection{Preliminaries}
First, we start with two common assumptions used in the study of OCO \citep{Intro:Online:Convex}. 

\begin{ass}\label{ass:1} The diameter of the domain $\Omega$ is bounded by $D$, i.e.,
\begin{equation}\label{eqn:domain}
\max_{\x, \y \in \Omega} \|\x -\y\| \leq D.
\end{equation}
\end{ass}
\begin{ass}\label{ass:2} The gradients of all the online functions are bounded by $G$, i.e.,
\begin{equation}\label{eqn:gradient}
\max_{\w \in \Omega}\|\nabla f_t(\w)\| \leq G, \ \forall t \in[T].
\end{equation}
\end{ass}

Next, we state definitions of strong convexity and exp-concavity \citep{Convex-Optimization,bianchi-2006-prediction}.
\begin{definition} \label{def:strong} A function $f: \Omega \mapsto \R$ is $\lambda$-strongly convex if
\begin{equation}\label{eqn:def:str}
    f(\y) \geq f(\x) +  \langle \nabla f(\x), \y -\x  \rangle + \frac{\lambda}{2} \|\y -\x \|^2,  \  \forall \x, \y \in \Omega. 
\end{equation}
\end{definition}
\begin{definition} \label{def:exp} A function $f: \Omega \mapsto \R$ is $\alpha$-exp-concave if $\exp(-\alpha f(\cdot))$ is concave over $\Omega$.
\end{definition}
The following property of exp-concave functions  will be used later \citep[Lemma 3]{ML:Hazan:2007}.
\begin{lemma} \label{lem:exp} For a function $f:\Omega \mapsto \R$, where $\Omega$ has diameter $D$, such that $\forall \w \in \Omega$, $\|\nabla f(\w)\|\leq G$ and $\exp(-\alpha f(\cdot))$ is concave, the following holds for $\beta = \frac{1}{2} \min\{\frac{1}{4GD}, \alpha \}$:
\begin{equation}\label{eqn:lem:exp}
    f(\y) \geq f(\x)+  \langle \nabla f(\x), \y -\x  \rangle + \frac{\beta}{2}   \langle \nabla f(\x), \y -\x  \rangle^2,  \  \forall \x, \y \in \Omega.
\end{equation}
\end{lemma}
\begin{figure*}
\centering
\begin{tabular}{@{}c@{\hspace{0.9ex}}*{17}{@{\hspace{0.5ex}}c}@{\hspace{0.9ex}}c@{}}
$t$ & 1 & 2 & 3 & 4 & 5 & 6 & 7 & 8 & 9 & 10 &11 &12 & 13 & 14 &15 & 16 & 17  &$\cdots$ \\
 $\I_0$ & [\quad ] & [\quad ] &  [\quad  ] & [\quad ] & [\quad ] & [\quad ] & [\quad ] & [\quad ] & [\quad ] & [\quad ] & [\quad ] &[\quad ] &[\quad ] & [\quad ] & [\quad ] & [\quad ]& [\quad ]& $\cdots$   \\
 $\I_1$ &  & [\quad  \phantom{]}& \phantom{[}\quad ] & [\quad \phantom{]} & \phantom{[}\quad ] & [\quad \phantom{]} & \phantom{[}\quad ] & [\quad \phantom{]} & \phantom{[}\quad ] & [\quad \phantom{]} & \phantom{[}\quad ] &[\quad \phantom{]} &\phantom{[}\quad ] & [\quad \phantom{]} & \phantom{[}\quad ] & [\quad \phantom{]} & \phantom{[}\quad ]& $\cdots$   \\
 $\I_2$  &  & &  & [\quad \phantom{]} & \phantom{[}\quad \phantom{]} & \phantom{[}\quad \phantom{]} & \phantom{[}\quad ] & [\quad \phantom{]} & \phantom{[}\quad \phantom{]} & \phantom{[}\quad \phantom{]} & \phantom{[}\quad ] &[\quad \phantom{]} &\phantom{[}\quad \phantom{]} & \phantom{[}\quad \phantom{]} & \phantom{[}\quad ] & [\quad \phantom{]} & \phantom{[}\quad \phantom{]} &$\cdots$   \\
  $\I_3$   &  & &  &  &  &  &  & [\quad \phantom{]} & \phantom{[}\quad \phantom{]} & \phantom{[}\quad \phantom{]} & \phantom{[}\quad \phantom{]} &\phantom{[}\quad \phantom{]} &\phantom{[}\quad \phantom{]} & \phantom{[}\quad \phantom{]} & \phantom{[}\quad ] & [\quad \phantom{]} & \phantom{[}\quad \phantom{]}& $\cdots$   \\
$\I_4$   &  & &  &  &  &  &  &  &  &  &  & & &  &  & [\quad \phantom{]} & \phantom{[}\quad \phantom{]}& $\cdots$   \\
\end{tabular}
\caption{Geometric covering (GC) intervals of \citet{Adaptive:ICML:15}.}
\label{fig:interval:saol}
\end{figure*}

\subsection{A Meta-expert Framework for Dual Adaptive Algorithms}
Most of existing adaptive algorithms~\citep{Adaptive:Hazan,Adaptive:ICML:15,Improved:Strongly:Adaptive,Dynamic:Regret:Adaptive} adopt a meta-expert framework, where multiple experts are created dynamically and aggregated by a meta-algorithm. Our universal algorithms are also built upon this framework, and we  subsequently detail the key components, including a set of intervals, the expert-algorithm and the meta-algorithm.  

\paragraph{GC Intervals}  To capture changing environments, we utilize the technique of sleeping experts \citep{Freund:1997:UCP}, where experts are active only at certain times and inactive otherwise. To determine the lifetime of experts,  we construct the geometric covering (GC) intervals~\citep{Adaptive:ICML:15}:
 \[
 \I= \bigcup_{k \in \N \cup \{0\}} \I_k,
 \]
where
  \[
  \I_k=\left\{ [ i \cdot 2^k, (i+1) \cdot 2^k -1]: i \in \N\right\}, \  k \in \N \cup \{0\}.
   \]
 A graphical illustration of GC intervals is given in Fig.~\ref{fig:interval:saol}. We observe that each $\I_k$ is a partition of $\N \setminus \{1,\cdots, 2^k-1 \}$ to consecutive intervals of length $2^k$.   The GC intervals can be generated on the fly, so we do not need to fix the horizon $T$. We note that similar intervals have been proposed by \citet{6543068}. 

\begin{algorithm}[tb]
   \caption{A Meta-expert Framework for Dual Adaptive Algorithms}
   \label{alg:framework}
\begin{algorithmic}[1]
\STATE Initialize the active expert set: $\A_0=\emptyset$ 
   \FOR{$t=1$ {\bfseries to} $T$}
   \STATE Update the active set: $\A_t = \A_{t-1}$
   \FORALL{$I=[r,s] \in \I$ that starts from $t$ }
    \STATE Construct one or multiple experts through $\mathcal{E}=\texttt{Construct-Experts}(I)$
    \FORALL{$E_i \in \mathcal{E}$}
    \STATE Set its ending time: $e_i = s$
    \STATE Initialize the associated parameters as $x_{t-1,i}=1$, $\gamma_i = 4 s^2$ and $L_{t-1,i}=0$ 
    \ENDFOR
    \STATE Add experts to the active set: $\A_t=\A_t \cup \mathcal{E}$
   \ENDFOR
   \STATE Set the learning rate and calculate the weight by \eqref{eqn:framework:xp} for each expert $E_i\in\A_t$
   \STATE Receive output $\w_{t,i}$ from each expert $E_i\in \A_t$ 
   \STATE Submit $\w_t$ in \eqref{eqn:framework:meta}
   \STATE Observe the loss $f_t(\cdot)$ and evaluate the gradient $\nabla f_t(\w_t)$
   \STATE Construct the normalized linearized loss $\ell_{t,i}$ by \eqref{eqn:framework:ellI} for each expert $E_i\in\A_t$
   \STATE Calculate the meta loss: $\ell_t = \sum_{E_i\in\A_t} p_{t,i}\ell_{t,i}$
   \FORALL{$E_i \in \A_t$}
   \STATE Update $L_{t,i}$ and $x_{t,i}$ by \eqref{eqn:framework:xtJ}
   \ENDFOR
   \STATE Remove experts whose ending times are $t$ from $\A_t$
   \ENDFOR
\end{algorithmic}
\end{algorithm}

\paragraph{Expert-algorithm} We construct experts by running appropriate expert-algorithms over each GC interval $I=[r,s]\in \I$. These experts become active in round $r$ and will be removed forever after round $s$. To deal with multiple types of functions, we propose two strategies: increasing the number of experts or enhancing the capabilities of experts. In the first strategy, we create multiple experts simultaneously over each interval to address the uncertainty of functions. In the second one, we create one expert over each interval by employing an universal algorithm for static regret.

\paragraph{Meta-algorithm} Inspired by \citet{ICML:2022:Zhang}, our meta-algorithm  chooses the linearized loss to measure the performance of experts, i.e., $l_t(\w)=\langle \nabla f_t(\w_t),\w-\w_t\rangle$, and makes use of second-order bounds to control the meta-regret. In this way, the meta-regret is small for exp-concave functions and strongly convex functions, and is also tolerable for convex functions. In this paper, we choose Adapt-ML-Prod \citep{pmlr-v35-gaillard14} over linearized loss as the meta-algorithm. As demonstrated by \citet{NeurIPS'22:efficient}, we can extend Adapt-ML-Prod to support sleeping experts. 

\paragraph{Overall Procedure} Our meta-expert framework for dual adaptive algorithms is summarized in Algorithm~\ref{alg:framework}. In the $t$-th round, for each interval $I=[r,s]\in\I$, we create one or multiple experts using the subroutine algorithm $\texttt{Construct-Experts}(I)$, which produces a set  consisting of  experts. From Steps~6 to 9, we set the ending time and the updating parameters for each expert. Then, we add the created experts to the active set in Step~10. In Step~12,  we set the learning rate for each expert in the active set and calculate the weight: 
\begin{equation}\label{eqn:framework:xp}
    \up_{t-1,i} = \left\{\frac{1}{2}, \sqrt{\frac{\ln \gamma_i}{1+L_{t-1,i}}}\right\}, \quad p_{t,i} = \frac{\up_{t-1,i}x_{t-1,i}}{\sum_{E_i\in\A_t}\up_{t-1,i}x_{t-1,i}}. 
\end{equation}
In Step~13, our framework collects the predictions of all the active experts, and aggregate them in Step~14:
\begin{equation}\label{eqn:framework:meta}
    \w_t = \sum_{E_i\in\A_t} p_{t,i} \w_{t,i}. 
\end{equation}
In Step~15, the framework observes the loss $f_t(\cdot)$ and evaluates the gradient $\nabla f_t(\w_t)$. In Step~16, we construct the normalized linearized loss for all the active experts:
\begin{equation}\label{eqn:framework:ellI}
    \ell_{t,i}=\frac{\langle\nabla f_t(\w_t),\w_{t,i}-\w_t\rangle+GD}{2GD}\in [0,1].
\end{equation}
In Step~17, we calculate the weighted average of $\ell_{t,i}$ as the loss of the meta-algorithm suffered in the $t$-th round. Finally, we update the the parameter $L_{t,i}$ and $x_{t,i}$ for all the active experts according to the rule of Adapt-ML-Prod (Steps~18 to 20):
\begin{equation}\label{eqn:framework:xtJ}
    L_{t,i}=L_{t-1,i} + (\ell_t-\ell_{t,i})^2,\quad x_{t,i} = \left(x_{t-1,i}\left(1+ \up_{t-1,i}(\ell_{t}-\ell_{t,i})\right)\right)^{\frac{\up_{t,i}}{\up_{t-1,i}}}. 
\end{equation}
In Step~19, the framework removes experts whose ending times are $t$ from $\A_t$. 

The meta-algorithm of our proposed framework satisfies the following theoretical guarantee, which is an informal version of Lemma~\ref{lem:TUMA:meta-regret}. 
\begin{lemma}
\label{lem:framework:meta-regret}
    \textnormal{(Informal)} Under Assumptions~\ref{ass:1} and \ref{ass:2}, for any GC interval $I=[r,s]\in \I$, the meta-regret of our framework in Algorithm~\ref{alg:framework} with respect to expert $E_i$ satisfies
    \begin{equation}\label{eqn:meta-expert:lemma}
        \sum_{t=r}^s \langle \nabla f_t(\w_t),\w_t-\w_{t,i} \rangle \leq \sqrt{\Xi_1 \sum_{t=r}^s \langle \nabla f_t(\w_t),\w_t-\w_{t,i} \rangle^2} + \Xi_2
    \end{equation}
    where $\Xi_1$ and $\Xi_2$ denote small constants that depend on the number of experts.  
\end{lemma}
\paragraph{Remark 1} Lemma~\ref{lem:framework:meta-regret} shows that, when functions are $\alpha$-exp-concave, we can make use of Lemma~\ref{lem:exp} and AM-GM inequality to obtain small meta-regret over any interval $[r,s]$,  
\begin{equation*}
\begin{aligned}
    & \sum_{t=r}^s f_t(\w_t) -  \sum_{t=r}^s f_t(\w_{t,i}) 
    \overset{\eqref{eqn:lem:exp}}{\leq}  \sum_{t=r}^s \langle \nabla f_t(\w_t),\w_t-\w_{t,i} \rangle - \frac{\beta}{2}  \sum_{t=r}^s\langle \nabla f_t(\w_t),\w_t-\w_{t,i} \rangle^2 \\
        \overset{\eqref{eqn:meta-expert:lemma}}{\leq} {} & \sqrt{\Xi_1 \sum_{t=r}^s \langle \nabla f_t(\w_t),\w_t-\w_{t,i} \rangle^2} + \Xi_2 - \frac{\beta}{2}  \sum_{t=r}^s\langle \nabla f_t(\w_t),\w_t-\w_{t,i} \rangle^2 
        \leq \frac{\Xi_1}{2\beta} +\Xi_2.
\end{aligned}
\end{equation*}
A similar derivation also holds for $\lambda$-strongly convex functions. For  convex functions, we can derive $O(\sqrt{s-r})$ meta-regret, which is optimal in the worst case. Based on this framework, we will elaborate on details of the expert construction in the following section.

\section{Universal Algorithms for Minimizing the Adaptive Regret}\label{sec:UMA2+3}
In this section, we present two kinds of universal algorithms for minimizing the adaptive regret, including two-layer approaches by increasing the number of experts and three-layer approaches by enhancing the capabilities of
experts. 

\subsection{A Two-layer Universal Algorithms Based on the Original Loss}\label{sec:UMA2:opt1}
The first two-layer method can be considered as an extension of the universal algorithm of \citet{ICML:2022:Zhang} from static regret to adaptive regret. The basic idea is to decompose the regret over any interval $I=[r,s]\subseteq [T]$ into the sum of the meta-regret and the expert-regret, which is formulated as,
\begin{equation}\label{eqn:decom}
    \sum_{t=r}^s f_t(\w_t) - \sum_{t=r}^s f_t(\w) = \underbrace{\sum_{t=r}^s f_t(\w_t) -\sum_{t=r}^s f_t(\w_{t,i})}_{\texttt{meta-regret}}+\underbrace{\sum_{t=r}^s f_t(\w_{t,i})-\sum_{t=r}^s f_t(\w)}_{\texttt{expert-regret}},
\end{equation}
where $\w_{t,i}$ denotes the output of an expert. According to our discussion in Remark~1, our meta-expert framework ensures small meta-regret for exp-concave functions and strongly convex functions, and manageable meta-regret for convex functions. As a result, we turn our attention to bounding the expert-regret. Following \citet{ICML:2022:Zhang}, we construct an expert for each type of convex functions and its possible modulus, achieving universality by increasing the number of experts. Specifically, we utilize OGD \citep{zinkevich-2003-online} and ONS \citep{ML:Hazan:2007} to handle three types of convex functions. When facing unknown moduli of strong convexity and exponential concavity, we assume they are both upper bounded and lower bounded, and discretize them by constructing a geometric series to cover the range of their values. Taking $\alpha$-exp-concave functions as an example, we assume $\alpha\in [1/T,1]$. Based on this interval, we set $\P_{exp}$ to be an exponentially spaced grid with a ratio of $2$:
\begin{equation}\label{eqn:Pexp}
    \P_{exp} = \left\{\frac{1}{T},\frac{2}{T},\frac{2^2}{T},\cdots \frac{2^N}{T}\right\},\quad N=\lceil \log_2 T\rceil.
\end{equation} 
In this way, $\P_{exp}$ can approximate $\alpha$ well in the sense that for any $\alpha\in [1/T,1]$, there must exist a $\hat{\alpha}\in \P_{exp}$ such that $\hat{\alpha}\leq \alpha\leq 2\hat{\alpha}$. Also, we can construct a similar set $\P_{str}$ for $\lambda$-strongly convex functions by assuming $\lambda\in [1/T,1]$: 
\begin{equation}\label{eqn:Pstr}
    \P_{str} = \left\{\frac{1}{T},\frac{2}{T},\frac{2^2}{T},\cdots \frac{2^N}{T}\right\},\quad N=\lceil \log_2 T\rceil.
\end{equation} 

\begin{algorithm}[t]
    \caption{$\texttt{Construct-Experts}(I)$}
    \label{alg:expert:option1}
 \begin{algorithmic}[1]
 \STATE Initialize the expert set $\mathcal{E}=\emptyset$
 \STATE Create an expert $E_I$ by running an instance of OGD to minimize $f_t(\cdot)$ during $I$, and add it to the expert set: $\mathcal{E}=\mathcal{E}\cup \{E_I\}$
 \FORALL{$\hat{\alpha} \in \P_{exp}$}
     \STATE Create an expert $E_I^{\hat{\alpha}}$ by running an instance of ONS to minimize $f_t(\cdot)$ with parameter $\hat{\alpha}$ during $I$, and add it to the expert set: $\mathcal{E}=\mathcal{E}\cup \{ E_I^{\hat{\alpha}} \}$
     \ENDFOR
     \FORALL{$\hat{\lambda}\in \P_{str}$}
      \STATE Create an expert $\widehat{E}_I^{\hat{\lambda}}$ by running an instance of OGD to minimize $f_t(\cdot)$ with parameter $\hat{\lambda}$ during $I$, and add it to the expert set: $\mathcal{E}=\mathcal{E}\cup \{ E_I^{\hat{\lambda}} \}$
    \ENDFOR 
 \STATE \textbf{Return:} Expert set $\mathcal{E}$ 
 \end{algorithmic}
 \end{algorithm}

Our first method for constructing experts is summarized in Algorithm~\ref{alg:expert:option1}. At the beginning, we create an expert to deal with general convex functions in Step~2, and add it to the expert set $\mathcal{E}$. For each parameter $\hat{\alpha}\in \P_{exp}$, we create an expert by running an instance of ONS with $\hat{\alpha}$ as the modulus of exponential concavity in Step~4, and also add it to the expert set. Similarly, for each parameter $\hat{\lambda}\in \P_{str}$, we also create an expert  by running an instance of OGD with $\hat{\lambda}$ as the modulus of strong convexity, and add it to the expert set (Steps~6 to 8).  

Combining the meta-expert framework in Algorithm~\ref{alg:framework} and \texttt{Construct-Experts} in Algorithm~\ref{alg:expert:option1}, our two-layer Universal algorithm for Minimizing the Adaptive regret of convex functions (UMA$2$)  enjoys the following theoretical guarantee.
\begin{thm}\label{thm:UMA2-op1}
    Under Assumptions~\ref{ass:1} and \ref{ass:2}, for any interval $I=[p,q]\subseteq [T]$ and any $\w\in\Omega$, UMA$2$ with Algorithm~\ref{alg:expert:option1} achieves $\SAReg(T,\tau) = O(\frac{d}{\alpha}\log \tau\log T)$, $O(\frac{1}{\lambda}\log \tau \log T)$, and $O(\sqrt{\tau\log T})$ for $\alpha$-exp-concave functions  with $\alpha\in[1/T,1]$, $\lambda$-strongly convex functions  with $\lambda\in [1/T,1]$, and general convex functions, respectively.
\end{thm}
\paragraph{Remark 2} Theorem~\ref{thm:UMA2-op1} shows that UMA$2$ with Algorithm~\ref{alg:expert:option1} is able to minimize the adaptive regret for three types of convex functions simultaneously. Because of dual adaptivity, our algorithm can handle the tough case that the type of functions switches or the parameter of functions changes. However, this algorithm exhibits two unfavorable characteristics: (i) it requires bounded moduli for $\alpha$-exp-concave functions and $\lambda$-strongly convex functions, and (ii) it necessitates multiple gradient estimations per round since each expert is required to process the original loss. In the following subsection, we resolve these two problems. 

\subsection{A Two-layer Universal Algorithms Based on the Surrogate Loss}\label{sec:two:UMA2}
To avoid the two limitations of UMA$2$ in Section~\ref{sec:UMA2:opt1}, we propose an alternative method, which draws inspiration from MetaGrad~\citep{NIPS2016_6268}. Instead of decomposing the regret in terms of the original loss in \eqref{eqn:decom}, we provide a novel regret decomposition based on the surrogate loss. As an example, consider  $\alpha$-exp-concave functions, for which we have 
\begin{equation}\label{eqn:two-layer:decom}
    \sum_{t=r}^s f_t(\w_t) - \sum_{t=r}^s f_t(\w) \overset{\eqref{eqn:lem:exp}}{\leq} \sum_{t=r}^s \langle \nabla f_t(\w_{t}), \w_t-\w\rangle - \frac{\beta}{2} V_{r,s} 
\end{equation}
where $V_{r,s}=\sum_{t=r}^s \langle \nabla f_t(\w_{t}), \w_t-\w\rangle^2$. Then, we decompose the linearized loss from \eqref{eqn:two-layer:decom} in the following way:
\begin{equation*}
    \sum_{t=r}^s \langle \nabla f_t(\w_{t}), \w_t-\w\rangle  =  \underbrace{\sum_{t=r}^s \langle \nabla f_t(\w_{t}), \w_t-\w_{t,i}\rangle}_{\texttt{meta-regret}} + \sum_{t=r}^s \langle \nabla f_t(\w_{t}), \w_{t,i}-\w\rangle.
\end{equation*}
Applying Lemma~\ref{lem:framework:meta-regret} to bound the meta-regret, we have 
\begin{equation}\label{eqn:two-layer:decom:2}
    \begin{aligned}
        &\sum_{t=r}^s \langle \nabla f_t(\w_{t}), \w_t-\w\rangle  \\
        \leq {} & \sqrt{\Xi_1\sum_{t=r}^s \langle \nabla f_t(\w_{t}), \w_t-\w_{t,i}\rangle^2} +\Xi_2 + \sum_{t=r}^s \langle \nabla f_t(\w_{t}), \w_{t,i}-\w\rangle  \\
        \leq  {} & \eta \sum_{t=r}^s \langle \nabla f_t(\w_{t}), \w_t-\w_{t,i}\rangle^2 + \frac{\Xi_1}{\eta} +\Xi_2 + \sum_{t=r}^s \langle \nabla f_t(\w_{t}), \w_{t,i}-\w\rangle  \\
        =   {} & \frac{1}{\eta} \underbrace{\sum_{t=r}^s (\ell_t^\eta (\w_{t,i}) - \ell_t^\eta (\w))}_{\texttt{expert-regret}} +\eta V_{r,s} +\frac{\Xi_1}{\eta} +\Xi_2 
    \end{aligned}
\end{equation}
where  the second inequality follows from AM-GM inequality, and the surrogate loss $\ell_t^\eta (\cdot)$  parameterized by a learning rate $\eta$ is defined as
\begin{equation} \label{eqn:metagrad:3}
\ell_t^\eta(\w)= -\eta\langle \nabla f_t(\w_t),  \w_t -\w \rangle + \eta^2 \langle \nabla f_t(\w_t),  \w_t -\w \rangle^2. 
\end{equation}
To bound the expert-regret in \eqref{eqn:two-layer:decom:2}, we can employ the slave algorithm of MetaGrad\footnote{As proven in \citet[Lemma~2]{Adaptive:Maler}, $\ell_t^\eta(\cdot)$ is $1$-exp-concave. Thus, we can also use ONS expert to minimize \eqref{eqn:metagrad:3}. Here, we choose the slave algorithm of MetaGrad as the expert-algorithm, as it offers slightly better guarantees than ONS. } to minimize~\eqref{eqn:metagrad:3}, and achieve  tight expert-regret $O(d\log (s-r))$. 
Next, to bound the remaining terms of \eqref{eqn:two-layer:decom:2}, we have two choices: 
\begin{compactitem}
    \item We set an appropriate $\eta\leq \frac{\beta}{2}$ to offset the second-order term $\eta V_{r,s}$ by the negative term $-\frac{\beta}{2} V_{r,s}$ from \eqref{eqn:two-layer:decom}.  To achieve this, we need to maintain multiple learning rates $\eta$ to account for all possible values of $\beta$. Therefore, it requires assuming that $\beta$ is bounded, which inherits the same limitation discussed in Section~\ref{sec:UMA2:opt1}. 
    \item We set an appropriate $\eta=\eta^*=\sqrt{(\Xi_1+O(d\log(s-r)))/V_{r,s}}$ in \eqref{eqn:two-layer:decom:2} to obtain a second-order bound:
    \begin{equation}\label{eqn:second:exp}
        \sum_{t=r}^s \langle \nabla f_t(\w_{t}), \w_t-\w\rangle 
\overset{\eqref{eqn:two-layer:decom:2}}{\leq}  2\sqrt{(\Xi_1 +O(d\log (s-r)))V_{r,s}} +\Xi_2.
\end{equation}
As revealed by the analysis of MetaGrad~~\citep{NIPS2016_6268}, $\eta^*$ is both upper and lower bounded, so we can construct the following discrete set to approximate all possible values of $\eta^*$:
    \begin{equation} \label{eqn:metagrad:4}
\S(I)=\left\{ \frac{2^{-i}}{5 DG} \ \left |\ i=0,1,\ldots, \left\lceil \frac{1}{2} \log_2 (s-r+1)\right\rceil \right. \right\}. 
\end{equation} 
Combining \eqref{eqn:second:exp} with \eqref{eqn:two-layer:decom} and applying  AM-GM inequality, we  obtain 
\begin{equation*}
    \sum_{t=r}^s f_t(\w_t) - \sum_{t=r}^s f_t(\w) \leq O\left(\frac{d}{\alpha}\log (s-r)\right).
\end{equation*}
We observe that an algorithm enjoying a second-order bound in \eqref{eqn:second:exp} can minimize the adaptive regret without knowing the value of $\alpha$. Compared to the first approach, this method does not require the assumption of bounded moduli, and we therefore adopt it. 
\end{compactitem}
To handle strongly convex functions, we propose a similar surrogate loss for each $\eta\in \S(I)$, which is inspired by Maler~\citep{Adaptive:Maler}, 
\begin{equation} \label{eqn:ell:hat}
\hat{\ell}_t^\eta(\w)= - \eta \langle \nabla f_t(\w_t),  \w_t -\w \rangle + \eta^2 G^2 \|\w_t -\w \|^2. 
\end{equation}
Since $\hat{\ell}_t^\eta(\cdot)$ is $2\eta G^2$-strongly convex~\citep[Lemma~2]{Adaptive:Maler}, we can employ OGD to minimize it. In this way, we obtain a similar second-order bound:
\begin{equation*}
    \sum_{t=r}^s \langle \nabla f_t(\w_t),\w_t-\w\rangle \leq 2G\sqrt{\left(\Xi_1+O(\log (s-r))\right) \sum_{t=r}^s \Vert \w_t-\w\Vert^2} + \Xi_2
\end{equation*}
which delivers the desired regret bound for strongly convex functions. Notably, the above bound can also provide a favorable regret bound for general convex functions, and thus we do not need to construct surrogate losses for the general convex case. 

Our second method for constructing experts is summarized in Algorithm~\ref{alg:expert:option2}. For each learning rate $\eta\in \S (I)$, we construct surrogate losses in \eqref{eqn:metagrad:3} and \eqref{eqn:ell:hat}, and employ the slave algorithm of MetaGrad and OGD to minimize them. 

\begin{algorithm}[t]
   \caption{$\texttt{Construct-Experts}(I)$}
   \label{alg:expert:option2}
\begin{algorithmic}[1]
\STATE Initialize the expert set $\mathcal{E}=\emptyset$
\FORALL{$\eta \in \S(|I|)$}
    \STATE Create an expert $E_I^{\eta}$ by running an instance of the slave algorithm of MetaGrad to minimize $\ell_t^\eta(\cdot)$ during $I$
     \STATE Create an expert $\widehat{E}_I^{\eta}$ by running an instance of OGD to minimize $\hat{\ell}_t^\eta(\cdot)$ during $I$
     \STATE Add the created experts into the set $\mathcal{E}=\mathcal{E}\cup \{E_I^{\eta},\widehat{E}_I^{\eta}\}$
   \ENDFOR 
\STATE \textbf{Return:} Expert set $\mathcal{E}$
\end{algorithmic}
\end{algorithm}

Combining the meta-expert framework in Algorithm~\ref{alg:framework} and  \texttt{Construct-Experts} in Algorithm~\ref{alg:expert:option2}, our two-layer Universal algorithm for Minimizing the Adaptive regret (UMA$2$) enjoys the following theoretical guarantee.

\begin{thm} \label{thm:main}
Under Assumptions~\ref{ass:1} and \ref{ass:2}, for any interval $[p,q] \subseteq [T]$ and any $\w \in \Omega$, UMA$2$ with Algorithm~\ref{alg:expert:option2} satisfies
\begin{align}
\sum_{t=p}^q\langle \nabla f_t(\w_t),  \w_t -\w \rangle &\leq \tau (p,q) b(p,q) + \frac{3}{2} \sqrt{a(p,q) b(p,q)} \sqrt{\sum_{t=p}^q\langle \nabla f_t(\w_t),  \w_t -\w \rangle^2},\label{eqn:UMA:inequality:0}\\
 \sum_{t=p}^q\langle \nabla f_t(\w_t),  \w_t -\w \rangle &\leq    \hat{\tau}(p,q) b(p,q) + \frac{3}{2} G\sqrt{\hat{a}(p,q) b(p,q)} \sqrt{\sum_{t=p}^q \| \w_t -\w \|^2}, \label{eqn:UMA:inequality:1}\\
\sum_{t=p}^q\langle \nabla f_t(\w_t),  \w_t -\w \rangle &\leq  \hat{\tau}(p,q) b(p,q) + \frac{21}{2} DG\sqrt{\hat{a}(p,q)(q-p+1)} \label{eqn:UMA:inequality:2}
\end{align}
where 
\begin{align} 
a(p,q)&= \frac{c(q)}{4}+\frac{1}{2}+\frac{d}{2} \ln \left(1+\frac{2}{25 d}(q-p+1)\right), \label{eqn:a} \\
b(p,q)&=  2\lceil\log_2 (q-p+2)\rceil, \label{eqn:b} \\
c(q)&= 32\ln (2q), \label{eqn:c} \\
\hat{a}(p,q) &= \frac{c(q)}{4}+1 + \log (q-p+1), \label{eqn:a:hat} \\
\tau (p,q) &= 2GD (5a(p,q)+2c(q)), \quad \hat{\tau} (p,q) = 2GD (5\hat{a}(p,q)+2c(q)). 
\end{align}
If all the online functions are $\alpha$-exp-concave, we have
\[
\sum_{t=p}^q f_t(\w_t) - \sum_{t=p}^q f_t(\w)  \leq \left(\frac{9}{8\beta}a(p,q)+\tau (p,q) \right) b(p,q)= O\left( \frac{d\log q \log(q-p)}{\alpha} \right).
\]
If all the online functions are $\lambda$-strongly convex, we have
\[
\sum_{t=p}^q f_t(\w_t) - \sum_{t=p}^q f_t(\w)  \leq \left( \frac{9G^2}{8\lambda}\hat{a}(p,q) +\hat{\tau}(p,q) \right) b(p,q)= O\left( \frac{\log q \log(q-p)}{\lambda} \right).
\]
\end{thm}
\paragraph{Remark 3} Theorem~\ref{thm:main} demonstrate that UMA$2$ with Algorithm~\ref{alg:expert:option2} is equipped with second-order regret bounds over any interval, i.e., \eqref{eqn:UMA:inequality:0} and \eqref{eqn:UMA:inequality:1}, leading to tight regret for exp-concave functions and strongly convex functions. Furthermore, \eqref{eqn:UMA:inequality:2} manifests that UMA attains $O(\sqrt{\tau \log T})$ strongly adaptive regret for general convex functions. In terms of the adaptive regret, UMA$2$ with Algorithm~\ref{alg:expert:option2} achieves the same theoretical guarantee as UMA$2$ with Algorithm~\ref{alg:expert:option1} while offering two advantages: (i) it removes the assumption of bounded moduli, and (ii) it estimates the gradient only once per round.

\subsection{A Three-layer Universal Algorithm for Minimizing the Adaptive Regret} \label{sec:three:UMA3}
In this subsection, we discuss a three-layer method that utilizes more powerful  experts. In previous two-layer methods, the meta-algorithm manages both function variations and changing environments. In contrast, in the three-layer method, we let the expert-algorithm to handle function variations. To bound the expert-regret in \eqref{eqn:decom}, we utilize Maler \citep{Adaptive:Maler}, an existing universal algorithm for static regret as the expert-algorithm. Consequently, the expert-regret over any GC interval can be bounded by the theoretical guarantee of Maler, allowing us to construct a smaller number of experts. Since Maler itself is a two-layer algorithm, combining it with the meta-algorithm transforms the overall algorithm into a three-layer architecture. Our method for constructing more powerful experts is summarized in Algorithm~\ref{alg:expert:UMA3}. Specifically, we create an expert by running an instance of Maler to minimize the original function, and return it. 

\begin{algorithm}[t]
   \caption{$\texttt{Construct-Experts}(I)$}
   \label{alg:expert:UMA3}
\begin{algorithmic}[1]
\STATE Create an expert $E_I$ by running an instance of Maler to minimize $f_t(\cdot)$ during $I$
\STATE \textbf{Return:} Expert set $\{E_I\}$
\end{algorithmic}
\end{algorithm}

Combining the meta-expert framework in Algorithm~\ref{alg:framework} and  \texttt{Construct-Experts} in Algorithm~\ref{alg:expert:UMA3}, our three-layer Universal algorithm for Minimizing the Adaptive regret (UMA$3$) enjoys the following theoretical guarantee.
\begin{thm}\label{thm:TUMA}
    Under Assumptions~\ref{ass:1} and \ref{ass:2}, for any interval $[p,q]\subseteq [T]$ and any $\w\in\Omega$, if all the online functions are $\alpha$-exp-concave, UMA$3$ satisfies
    \begin{equation*}
        \begin{aligned}
            \sum_{t=p}^q f_t(\w_t) - \sum_{t=p}^q f_t(\w) &\leq \left(10GD+\frac{9}{2\beta}\right) b(p,q) \left(c(q)+\Xi(p,q)+10d\log (q-p+1)\right) \\
            &= O\left(\frac{d\log q \log (q-p)}{\alpha}\right)
        \end{aligned}
    \end{equation*}
    where $\beta=\frac{1}{2} \{\frac{1}{4GD},\alpha\}$, $b(\cdot,\cdot)$ and $c(\cdot)$ are given in \eqref{eqn:b} and \eqref{eqn:c} respectively, and 
    \begin{equation*}
        \Xi(p,q) = 2\ln \left(\frac{\sqrt{3}}{2}\log_2 (q-p+1)+3\sqrt{3}\right).
    \end{equation*}
    If all the online functions are $\lambda$-strongly convex, UMA$3$ satisfies
    \begin{equation*}
        \begin{aligned}
            \sum_{t=p}^q f_t(\w_t) - \sum_{t=p}^q f_t(\w) &\leq \left(10GD+\frac{9G^2}{2\lambda}\right) b(p,q) \left(c(q)+\Xi(p,q)+10d\log (q-p+1)\right) \\
            &= O\left(\frac{\log q \log (q-p)}{\lambda}\right)
        \end{aligned}
    \end{equation*}
    If all the online functions are general convex, UMA$3$ satisfies
    \begin{equation*}
        \sum_{t=p}^q f_t(\w_t) - \sum_{t=p}^q f_t(\w) \leq 2GDc(q)b(p,q) + GD \left(\sqrt{c(q)} +7 \right)\sqrt{q-p+1}=O\left(\sqrt{(q-p) \log q}\right). 
    \end{equation*}
\end{thm}
\paragraph{Remark 4} Theorem~\ref{thm:TUMA} demonstrates that UMA$3$ is able
to minimize the adaptive regret for three types
of convex functions  simultaneously. Specifically, it achieves $\SAReg(T,\tau)= O(\frac{d}{\alpha}\log\tau\log T)$, $O(\frac{1}{\lambda}\log \tau \log T)$, and $O(\sqrt{\tau\log T})$ for $\alpha$-exp-concave, $\lambda$-strongly convex, and general convex functions, respectively. Furthermore, UMA$3$ also enjoys dual adaptivity, which can manage changes in the type of functions and the parameter of functions. Since Maler does not require the bounded moduli assumption, UMA$3$ does not require it either.

\section{A Universal Algorithm for Minimizing the Adaptive Regret of Online Composite Optimization} \label{sec:Comp}
In this section, we extend our universal algorithms to online composite optimization, where the online learner suffers a composite loss $F_t(\cdot)\triangleq f_t(\cdot)+r(\cdot)$ in the $t$-th round.

\subsection{A Meta-expert Framework for Online Composite Optimization}
First, we introduce two standard assumptions in online composite optimization \citep{DBLP:journals/jmlr/DuchiS09,DBLP:conf/colt/DuchiSST10}. 
\begin{ass}\label{ass:3}
    The regularization function $r(\cdot)$ in \eqref{eqn:composite} is convex over $\Omega$. 
\end{ass}
\begin{ass}\label{ass:4}
    The regularization function $r(\cdot)$ in \eqref{eqn:composite} is non-negative and bounded by a constant $C$, i.e., $\forall \w\in\Omega$, $0\leq r(\w)\leq C$. 
\end{ass}
Similar to Algorithm~\ref{alg:framework}, we also adopt the meta-expert framework. 
For the key components of the framework, GC intervals can be directly utilized to capture changing environments. The difference lies in the design of the meta-algorithm and the expert-algorithm. 
\paragraph{Meta-algorithm} Inspired by \citet{zhang2024universal}, we choose Optimistic-Adapt-ML-Prod \citep{NIPS:2016:Wei} with suitable configurations as the meta-algorithm, which can control the meta-regret by eliminating the influence of $r(\cdot)$. To deal with changing environments, we extend Optimistic-Adapt-ML-Prod to support sleeping experts. 
\paragraph{Expert-algorithm} Recall that we propose two strategies for constructing experts in Section~\ref{sec:UMA2+3}. In the composite setting, both strategies can be similarly applied. For simplicity, we choose to construct more powerful universal experts. Although there exists a universal algorithm for static regret of composite functions, it requires the moduli of exp-concave functions and strongly convex functions to be constrained within the range of $[1/T,1]$ \citep{zhang2024universal}. Therefore, we propose a novel universal algorithm for online composite optimization in Section~\ref{sec:UMS-Comp}, which avoids the assumption on bounded moduli. 

\paragraph{Overall Procedure} Our meta-expert framework for dual adaptive algorithms of online composite optimization is summarized in Algorithm~\ref{alg:framework:composite}. The procedure is similar to that of Algorithm~\ref{alg:framework}, with the incorporation of an optimistic estimation (also called optimism) into the meta-algorithm. Specifically, in the $t$-th round, we create one or multiple experts for each interval $I=[r,s]\in\I$ by using the subroutine algorithm $\texttt{Construct-Experts}$, which produces an expert set. In Step~14, we compute the optimistic estimation of each expert:
\begin{equation}\label{eqn:framework:UMA-Comp:mtI}
    m_{t,i} = \frac{1}{GD} \left( \sum_{E_j\in\A_t} p_{t,j}r(\w_{t,j}) - r(\w_{t,i}) \right). 
\end{equation}
In Step~15, our framework sets the learning rate and calculates the weight as follows:
\begin{equation}\label{eqn:framework:UMA-Comp:etatI}
    \up_{t-1,i} = \min \left\{ \frac{1}{4},\sqrt{\frac{\gamma_i}{1+L_{t-1,i}}}\right\},\quad p_{t,i} = \frac{\up_{t-1,i}\widetilde{x}_{t-1,i}}{\sum_{E_j\in\A_t} \up_{t-1,j}\widetilde{x}_{t-1,j}},
\end{equation}
where $\widetilde{x}_{t-1,i}=x_{t-1,i}\exp (\up_{t-1,i}m_{t,i})$. 

\begin{algorithm}[tb]
    \caption{A Meta-expert Framework for Dual Adaptive Algorithms of Online Composite Optimization}
    \label{alg:framework:composite}
 \begin{algorithmic}[1]
 \STATE Initialize  the active expert set: $\A_0=\emptyset$
 \STATE Observe the convex regularizer $r(\cdot)$
    \FOR{$t=1$ {\bfseries to} $T$}
    \STATE Update the active set: $\A_t=\A_{t-1}$
    \FORALL{$I \in \I$ that starts from $t$ }
     \STATE Construct one or multiple experts through $\mathcal{E}=\texttt{Construct-Experts}(I)$
     \FORALL{$E_i\in\mathcal{E}$}
     \STATE Set tis ending time: $e_i=s$
     \STATE Initialize $x_{t-1,i}=1$, $\gamma_i = 4s^2$ and $L_{t-1,i}=0$ 
     \ENDFOR
     \STATE Add experts to the active set: $\A_t=\A_t \cup \mathcal{E}$
    \ENDFOR
    \STATE Receive output $\w_{t,i}$ from each expert $E_i\in \A_t$ 
    \STATE Compute the optimism $m_{t,i}$ of each expert $E_i$ by \eqref{eqn:framework:UMA-Comp:mtI}
    \STATE Set the learning rate and calculate the weight by \eqref{eqn:framework:UMA-Comp:etatI} for each expert $E_i\in\A_t$
    \STATE Submit $\w_t$ in \eqref{eqn:framework:UMA-Comp:wt}
    \STATE Observe the loss $f_t(\cdot)$ and evaluate the gradient $\nabla f_t(\w_t)$
    \STATE Construct the normalized linearized loss $\ell_{t,i}$ by \eqref{eqn:framework:UMA-Comp:elltI} for each expert $E_i\in\A_t$
    \STATE Calculate the meta loss: $\ell_t = \sum_{E_i\in\A_t} p_{t,i}\ell_{t,i}$
    \FORALL{$E_i \in \A_t$}
    \STATE Update $L_{t,i}$ and $x_{t,i}$ by \eqref{eqn:framework:UMA-Comp:xtI}
    \ENDFOR
    \STATE Remove experts whose ending times are $t$ from $\A_t$
    \ENDFOR
 \end{algorithmic}
 \end{algorithm}

Here, we would like to clarify that the term $\sum_{E_j\in\A_t} p_{t,j}r(\w_{t,j})$ of $m_{t,i}$ in \eqref{eqn:framework:UMA-Comp:mtI} could be computed before the weights $p_{t,i}$ (which also depends on $m_{t,i}$) are assigned~\citep{zhang2024universal}. The basic idea is to treat $\gamma=\sum_{E_j\in\A_t} p_{t,j}r(\w_{t,j})$ as the fixed point of a continuous function. To find the value of $\gamma$, we can deploy the binary-search strategy, which only suffer $1/T$ error in $\log T$  iterations and therefore, does not affect the regret bound. For details, please refer to~\citet{NIPS:2016:Wei}. 

In Step~16, our framework submits the following solution:
\begin{equation}\label{eqn:framework:UMA-Comp:wt}
    \w_t = \sum_{E_i\in\A_t} p_{t,i} \w_{t,i}. 
\end{equation}
Since Optimistic-Adapt-ML-Prod requires $|\ell_t-\ell_{t,i}-m_{t,i}|\leq 2$, we construct the normalized linearized loss in Step~18:
\begin{equation}\label{eqn:framework:UMA-Comp:elltI}
    \ell_{t,i} = \frac{1}{GD} \left( \langle\nabla f_t(\w_t),\w_{t,i}-\w_t \rangle + r(\w_{t,i}) \right). 
\end{equation}
Finally, we update the parameter $L_{t,i}$ and $x_{t,i}$ for all the active experts according to the rule of Optimistic-Adapt-ML-Prod:
\begin{equation}\label{eqn:framework:UMA-Comp:xtI}
\begin{aligned}
    L_{t,i} &=L_{t-1,i} + (\ell_t-\ell_{t,i}-m_{t,i})^2, \\
    x_{t,i} &= \left(x_{t-1,i}\exp \left(x_{t-1,i} \left(\ell_{t}-\ell_{t,i}\right) - \left( x_{t-1,i}(\ell_{t}-\ell_{t,i} - m_{t,i})\right)^2  \right)\right)^{\frac{\up_{t,i}}{\up_{t-1,i}}}. 
\end{aligned}
\end{equation}
The meta-algorithm of our framework with appropriate optimism estimation in \eqref{eqn:framework:UMA-Comp:mtI} satisfies the following theoretical guarantee, which is an informal version of Lemma~\ref{lem:TUMA-Comp:meta-regret}. 
\begin{lemma}
\label{lem:framework:composite:meta-regret}
    \textnormal{(Informal)} Under Assumptions~\ref{ass:1}, \ref{ass:2}, \ref{ass:3} and \ref{ass:4}, for any interval $I=[r,s]\in \I$, the meta-regret of our framework in Algorithm~\ref{alg:framework} with respect to expert $E_i$ satisfies
    \begin{equation*}
        \sum_{t=r}^s \langle \nabla f_t(\w_t),\w_t-\w_{t,i} \rangle + \sum_{t=r}^s r(\w_t) - \sum_{t=r}^s r(\w_{t,i}) \leq \sqrt{\Xi_1 \sum_{t=r}^s \langle \nabla f_t(\w_t),\w_t-\w_{t,i} \rangle^2} + \Xi_2
    \end{equation*}
    where $\Xi_1$ and $\Xi_2$ denote some constants that depend on the number of experts.  
\end{lemma}
\paragraph{Remark 5} Lemma~\ref{lem:framework:composite:meta-regret} demonstrates that our meta-expert framework with appropriate estimations can deliver a second-order meta-regret bound that solely depends on the time-varying function $f_t(\cdot)$. Therefore, we can directly exploit the property of exp-concave functions and strongly convex functions to control the meta-regret. 

\subsection{A Universal Algorithm for Minimizing the Static Regret of Online Composite Optimization}\label{sec:UMS-Comp}
In this section, we propose a Universal algorithm for Minimizing the Static regret of online Composite optimization (UMS-Comp). UMS-Comp also adopts the meta-expert framework, similar to UMA$2$ in Section~\ref{sec:two:UMA2}. The main differences are as follows: (i) the regularizer is incorporated into the surrogate loss for the expert-algorithm, (ii) GC intervals are not required, and (iii) Optimistic-Adapt-ML-Prod is chosen as the meta-algorithm. In the following, we provide the details. 
\begin{algorithm}[t]
   \caption{Meta-algorithm of UMS-Comp}
   \label{alg:UMS-Comp:meta}
\begin{algorithmic}[1] 
\STATE Observe the convex regularizer $r(\cdot)$
\STATE Construct multiple experts through $\mathcal{E}=\texttt{Construct-Experts}(T)$
\STATE Initialize $x_{0,i}=1/\vert \mathcal{E} \vert$ and $L_{0,i}=0$ for each expert $E_i\in \mathcal{E}$
   \FOR{$t=1$ {\bfseries to} $T$}
   \STATE Receive output $\u_{t,i}$ from each expert $E_i$ in $\EC$
   \STATE Compute the optimism $m_{t,i}$ of each expert by \eqref{eqn:UMS-Comp:m}
   \STATE Set the learning rate and calculate the weight $p_{t,i}$ of each expert by \eqref{eqn:UMS-Comp:p}
   \STATE Output the weighted average $\u_t= \sum_{i=1}^{\vert \mathcal{E}\vert} p_{t,i} \u_{t,i}$
   \STATE Observe the loss $f_t(\cdot)$ and evaluate the gradient $\nabla f_t(\u_t)$
   \STATE Construct the normalized linearized loss $\ell_{t,i}$ by \eqref{eqn:UMS-Comp:meta:lti} for each expert $E_i \in \EC$
   \STATE Calculate the meta-loss: $\ell_t = \sum_{i=1}^{\vert \EC\vert} p_{t,i} \u_{t,i}$ 
   \FORALL{$E_i\in \EC$}
   \STATE Update $L_{t,i}$ and $x_{t,i}$ by \eqref{eqn:framework:UMA-Comp:xtI}
   \ENDFOR
   \ENDFOR
\end{algorithmic}
\end{algorithm}

\paragraph{Meta-algorithm} The meta-algorithm of UMS-Comp is summarized in Algorithm~\ref{alg:UMS-Comp:meta}. In the beginning, we construct multiple experts through the subroutine algorithm, and initialize the parameters (Steps~2 to 3). In the $t$-th round, we receive the output from each expert in Step~5, and compute the optimism of each expert in Step~6:
\begin{equation}\label{eqn:UMS-Comp:m}
    m_{t,i} = \frac{1}{GD} \left( \sum_{i=1}^{\vert\mathcal{E}\vert} p_{t,i} r(\u_{t,i})  - r(\u_{t,i}) \right).
\end{equation}
In Step~7, we set the learning rate and calculate the weight of each expert according to Optimistic-Adapt-ML-Prod~\citep{NIPS:2016:Wei}:
\begin{equation}\label{eqn:UMS-Comp:p}
    \up_{t-1,i} = \min \left\{ \frac{1}{4},\sqrt{\frac{\ln \vert\mathcal{E}\vert}{1+L_{t-1,i} }} \right\}, \quad p_{t,i} = \frac{\up_{t-1,i} \widetilde{x}_{t-1,i} }{\sum_{i=1}^{\vert\mathcal{E}\vert} \up_{t-1,i} \widetilde{x}_{t-1,i}}
\end{equation}
where $\widetilde{x}_{t-1,i}=x^\eta_{t-1}\exp (\up_{t-1,i}m_{t,i})$. Then, the meta-algorithm outputs the weighted average decision in Step~8. After observing the information of $f_t(\cdot)$, we construct the normalized linearized loss in Step~10:
\begin{equation}\label{eqn:UMS-Comp:meta:lti}
    \ell_{t,i} = \frac{1}{GD} \left( \langle \nabla f_t(\u_t),\u_{t,i}-\u_t\rangle+r(\u_{t,i}) \right).
\end{equation}
Finally, we update the parameter $L_{t,i}$ and $x_{t,i}$ for all experts by \eqref{eqn:framework:UMA-Comp:xtI}. 
\begin{algorithm}[t]
    \caption{$\texttt{Construct-Experts}(T)$}
    \label{alg:UMS-Comp:expert}
 \begin{algorithmic}[1]
 \STATE Initialize the expert set $\mathcal{E}=\emptyset$
 \STATE Create an expert $\widetilde{E}$ by running an instance of FOBOS to minimize the original composite function $f_t(\cdot)+r(\cdot)$, and add it into the set $\EC=\EC\cup \{\widetilde{E}\}$
 \FORALL{$\eta \in \S(|T|)$}
     \STATE Create an expert $E^{\eta}$ by running an instance of ProxONS to minimize $\ell_t^\eta(\cdot)$ in \eqref{eqn:comp:surr:exp}
      \STATE Create an expert $\widehat{E}^{\eta}$ by running an instance of FOBOS to minimize $\hat{\ell}_t^\eta(\cdot)$ in \eqref{eqn:comp:surr:str}
      \STATE Add the created experts into the set $\mathcal{E}=\mathcal{E}\cup \{E^{\eta},\widehat{E}^{\eta}\}$
    \ENDFOR 
 \STATE \textbf{Return:} Expert set $\mathcal{E}$
 \end{algorithmic}
 \end{algorithm}

\paragraph{Expert-algorithm} Following  UMA$2$ in Section~\ref{sec:two:UMA2}, we construct the following surrogate loss to handle exp-concavity:
\begin{equation}\label{eqn:comp:surr:exp}
    \ell_t^\eta (\u) = - \eta \langle \nabla f_t(\u_t),\u_t-\u\rangle+ \eta^2 \langle \nabla f_t(\u_t),\u_t-\u\rangle^2 + \eta r(\u) 
\end{equation}
for each learning rate $\eta$ in
\begin{equation} \label{eqn:ST}
\S(T)=\left\{ \frac{2^{-i}}{5 DG} \ \left |\ i=0,1,\ldots, \left\lceil \frac{1}{2} \log_2 T\right\rceil \right. \right\}. 
\end{equation} 
To minimize the loss in \eqref{eqn:comp:surr:exp}, we use an existing algorithm for exp-concave functions with regularizer, i.e., ProxONS~\citep{DBLP:journals/ijon/YangTCWS24}. Next, for strongly convex functions, we construct a similar surrogate loss: 
\begin{equation}\label{eqn:comp:surr:str}
    \hat{\ell}_t^\eta (\u) = -\eta \langle \nabla f_t(\u_t),\u_t-\u\rangle + \eta^2 G^2 \Vert \u_t-\u\Vert^2 + \eta r (\u),
\end{equation}
for each $\eta$ in \eqref{eqn:ST}, and use FOBOS~\citep{DBLP:journals/jmlr/DuchiS09} to minimize it. In Section~\ref{sec:two:UMA2}, we reuse the second-order bound for strongly convex functions to deal with general convex functions. However, for minimizing the static regret, this approach will result in a suboptimal $O(\sqrt{T\log T})$ bound for general convex functions. To resolve this issue, we create one additional expert by running an instance of  FOBOS~\citep{DBLP:journals/jmlr/DuchiS09} to minimize the original composite function. Our method for constructing experts  is summarized in Algorithm~\ref{alg:UMS-Comp:expert}.   

Combining the meta-algorithm in Algorithm~\ref{alg:UMS-Comp:meta} and \texttt{Construct-Experts} in Algorithm~\ref{alg:UMS-Comp:expert}, UMS-Comp enjoys the following theoretical guarantee. 
\begin{thm}
    \label{thm:UMS-Comp}
    Under Assumptions~\ref{ass:1}, \ref{ass:2}, \ref{ass:3} and \ref{ass:4}, for a 
$T$-round game and any $\u\in\Omega$, when the time-varying function $f_t(\cdot)$ is $\alpha$-exp-concave, UMS-Comp satisfies 
    \begin{equation*}
        \sum_{t=1}^T F_t(\u_t) - \sum_{t=1}^T F_t(\u)  \leq \left( \frac{9}{8\beta }+10GD \right) \cdot \left(4d\ln (T+1)+\phi_1+4\right) + 2GD \phi_2 
    \end{equation*}
    where $F_t(\cdot)\triangleq f_t(\cdot)+r(\cdot)$, and $\phi_1$ and $\phi_2$ are defined as
    \begin{equation}\label{eqn:phi}
        \begin{aligned}
            \phi_1 &= \frac{1}{4}\left(\ln \left( 3+\lceil \log_2 T\rceil \right)  + \ln \left( 1+\frac{3+\lceil \log_2 T\rceil}{e} (1+\ln (T+1)) \right)\right)^2 = O(\log\log T) \\
            \phi_2 &= 19 \ln \left( 3+\lceil \log_2 T\rceil \right) + \frac{1}{4} \ln \left( 1+\frac{3+\lceil \log_2 T\rceil}{e} (1+\ln (T+1)) \right)= O(\log\log T). 
        \end{aligned}
    \end{equation}
    When the time-varying function $f_t(\cdot)$ is $\lambda$-strongly convex, UMS-Comp satisfies 
    \begin{equation*}
        \sum_{t=1}^T F_t(\u_t) - \sum_{t=1}^T F_t(\u)  \leq \left( \frac{9G^2}{\lambda }+10GD \right) \cdot \left(7\log T+8+\phi_1 \right) + 2GD \phi_2. 
    \end{equation*}
    When the time-varying function $f_t(\cdot)$ is general convex, UMS-Comp satisfies 
    \begin{equation*}
        \sum_{t=1}^T F_t(\u_t) - \sum_{t=1}^T F_t(\u)  \leq   GD \phi_3 \sqrt{T}+ GD(\phi_2+1)
    \end{equation*}
    where 
    \begin{equation}\label{eqn:phi3}
        \phi_3=\sqrt{7}+\ln \left( 3+\lceil \log_2 T\rceil \right)  + \ln \left( 1+\frac{3+\lceil \log_2 T\rceil}{e} (1+\ln (T+1)) \right)=O(\log\log T).
    \end{equation} 
\end{thm}
\paragraph{Remark 6} Theorem~\ref{thm:UMS-Comp} demonstrates that UMS-Comp attains optimal static regret for three types of convex $f_t(\cdot)$ simultaneously. Specifically, for $\alpha$-exp-concave functions,  UMS-Comp achieves $O(\frac{d}{\alpha}\log T)$ static regret without knowing the value of $\alpha$. Moreover, for $\lambda$-strongly convex functions, UMS-Comp achieves $O(\frac{1}{\lambda}\log T)$ static regret without knowing the value of $\lambda$. Finally, for general convex functions,  UMS-Comp achieves $O(\sqrt{T})$ static regret. Compared to the existing universal algorithm for composite functions~\citep{zhang2024universal}, UMS-Comp avoids the assumption of bounded moduli of functions. 

\subsection{A Universal Algorithm for Minimizing the Adaptive Regret of Online Composite Optimization}
In this subsection, we introduce our universal algorithm for adaptive regret in the composite setting, similar to UMA$3$ in Section~\ref{sec:three:UMA3}. Following the regret decomposition in~\eqref{eqn:decom}, we decompose the regret of composite functions over interval $I=[r,s]$ into the sum of the meta-regret and the expert-regret: 
\begin{equation}\label{eqn:decom:comp}
\sum_{t=r}^s F_t(\w_t) - \sum_{t=r}^s F_t(\w) 
=  \underbrace{\sum_{t=r}^s F_t(\w_t) - \sum_{t=r}^s F_t(\w_{t,i})}_{\texttt{meta-regret}} +\underbrace{\sum_{t=r}^s F_t(\w_{t,i}) - \sum_{t=r}^s F_t(\w)}_{\texttt{expert-regret}}. 
\end{equation}
Based on our proposed meta-expert framework, we adopt UMS-Comp as the expert-algorithm, thereby bounding the expert-regret in \eqref{eqn:decom:comp}. Our method for constructing experts is summarized in Algorithm~\ref{alg:expert:UMS}. We create an expert by running an instance of UMS-Comp to minimize the composite function, and return it. 
\begin{algorithm}[t]
   \caption{$\texttt{Construct-Experts}(I)$}
   \label{alg:expert:UMS}
\begin{algorithmic}[1]
\STATE Create an expert $E_I$ by running an instance of UMS-Comp to minimize the 
 composite function $f_t(\cdot)+r(\cdot)$ during $I$
\STATE \textbf{Return:} Expert set $\{E_I\}$
\end{algorithmic}
\end{algorithm}

Combining the meta-expert framework in Algorithm~\ref{alg:framework:composite} and \texttt{Construct-Experts} in  Algorithm~\ref{alg:expert:UMS}, our Universal algorithm for Minimizing the Adaptive regret of online Composite optimization (UMA-Comp) enjoys the following theoretical guarantee. 
\begin{thm} \label{thm:UMA-Comp}
Under Assumptions~\ref{ass:1}, \ref{ass:2}, \ref{ass:3} and \ref{ass:4}, for any interval $[p,q]\subseteq [T]$ and any $\w\in\Omega$, if the time-varying function $f_t(\cdot)$ is $\alpha$-exp-concave, UMA-Comp satisfies
\begin{equation*}
\sum_{t=p}^q F_t(\w_t) - \sum_{t=p}^q F_t(\w)\leq \left(GD+\frac{1}{2\beta}\right) c(q)b(p,q)+ \varphi (p,q) b(p,q) = O\left(\frac{d\log q \log (q-p)}{\alpha}\right)
\end{equation*}
where $\varphi (p,q) = (\frac{9}{8\beta}+10GD)\cdot (4d\ln (q-p+2)+\phi_1+4)+2GD\phi_2$, $b(\cdot,\cdot)$ and $c(\cdot)$ are defined in \eqref{eqn:b} and \eqref{eqn:c} respectively, and $\phi_1$, $\phi_2$ are defined in \eqref{eqn:phi}. 

If the time-varying function $f_t(\cdot)$ is $\lambda$-strongly convex, UMA-Comp satisfies
\begin{equation*}
\sum_{t=p}^q F_t(\w_t) - \sum_{t=p}^q F_t(\w)\leq\left(GD+\frac{G^2}{2\lambda}\right) c(q)b(p,q)+\hat{\varphi}(p,q) b(p,q) = O\left(\frac{\log q \log (q-p)}{\lambda}\right)
\end{equation*}
where $\hat{\varphi}(p,q) = ( \frac{9G^2}{\lambda }+10GD ) \cdot (7\log (s-r+1)+\phi_1+8 ) + 2GD \phi_2$. 

If the time-varying function $f_t(\cdot)$ is general convex, UMA-Comp satisfies
\begin{equation*}
    \begin{aligned}
        \sum_{t=p}^q F_t(\w_t) - \sum_{t=p}^q F_t(\w) &\leq GD c(q)b(p,q) + GD \left(\sqrt{c(q)}+\phi_3\right) \sqrt{q-p+1} +GD(\phi_2+1)\\
        &= O\left(\sqrt{(q-p)\log q}\right)
    \end{aligned}
\end{equation*}
where $\phi_3$ is defined in \eqref{eqn:phi3}. 
\end{thm}
\paragraph{Remark 7} According to Theorem~\ref{thm:UMA-Comp}, UMA-Comp with appropriate configurations guarantees that the additional regularizer does not affect the adaptive regret. Therefore, our algorithm can deliver the same order of adaptive regret as UMA$2$ or UMA$3$ for three types of convex $f_t(\cdot)$. 

\section{Analysis}\label{sec:analysis}
Here, we present proofs of main theorems and lemmas. 
\subsection{Proof of Theorem~\ref{thm:UMA2-op1}}
First, we start with the meta-regret over the interval $[r,s]$. After combining the expert-regret, we extend it to any interval $[p,q]\subseteq [T]$. The following theoretical guarantee is a special case of Lemma~\ref{lem:TUMA-Comp:meta-regret}, which is the theoretical result of UMA-Comp in the composite setting, and we set $m_{t,i}=0$ to obtain the following lemma. 
\begin{lemma}\label{lem:UMA2-opt1:meta-regret}
Under Assumptions~\ref{ass:1} and \ref{ass:2}, for any interval $I=[r,s]\in \I$, the meta-regret of UMA$2$ with Algorithm~\ref{alg:expert:option1} with respect to any active expert $E_i$ satisfies
    \begin{equation*}
        \sum_{t=r}^s \ell_{t} - \sum_{t=r}^s \ell_{t,i} \leq \frac{\Gamma_i}{\sqrt{\gamma_i}} \sqrt{1+\sum_{t=r}^s (\ell_t-\ell_{t,i})^2} + 2\Gamma_i
    \end{equation*}
    where $\Gamma_i = 2\gamma_i+\ln N_s + \ln \ln \left(9+36s\right)$ and $N_s$ is the number of experts created till round $s$. 
\end{lemma}
According to the definition of $\ell_t$ and $\ell_{t,i}$ in \eqref{eqn:framework:ellI}, we have
\begin{equation}\label{eqn:UMA2-opt1:meta-regret:1}
\begin{aligned}
    \sum_{t=r}^s \langle\nabla f_t(\w_t),\w_t-\w_{t,i}\rangle &\leq  \frac{\Gamma_i}{\sqrt{\gamma_i}} \sqrt{4G^2D^2+\sum_{t=r}^s \langle\nabla f_t(\w_t),\w_t-\w_{t,i}\rangle^2} + 4GD\Gamma_i \\
    &\leq 2GD\left(\frac{\Gamma_i}{\sqrt{\gamma_i}}+2\Gamma_i \right) + \sqrt{\frac{\Gamma_i^2}{\gamma_i}\sum_{t=r}^s \langle\nabla f_t(\w_t),\w_t-\w_{t,i}\rangle^2}
\end{aligned}
\end{equation}
where the last step is due to $\sqrt{a+b}\leq \sqrt{a}+\sqrt{b}$. When functions are $\alpha$-exp-concave during the interval $[r,s]$, the meta-regret with respect to any expert $E_i$ is bounded by
\begin{equation}\label{eqn:meta-regret:exp}
    \begin{aligned}
    &\sum_{t=r}^s f_t(\w_t) - \sum_{t=r}^s f_t(\w_{t,i}) 
    \overset{\eqref{eqn:lem:exp}}{\leq}  \sum_{t=r}^s \langle\nabla f_t(\w_t),\w_t-\w_{t,i}\rangle - \frac{\beta}{2} \sum_{t=r}^s \langle\nabla f_t(\w_t),\w_t-\w_{t,i}\rangle^2 \\
    \overset{\eqref{eqn:UMA2-opt1:meta-regret:1}}{\leq} {} & 2GD\left(\frac{\Gamma_i}{\sqrt{\gamma_i}}+2\Gamma_i \right) + \sqrt{\frac{\Gamma_i^2}{\gamma_i}\sum_{t=r}^s \langle\nabla f_t(\w_t),\w_t-\w_{t,i}\rangle^2} - \frac{\beta}{2} \sum_{t=r}^s \langle\nabla f_t(\w_t),\w_t-\w_{t,i}\rangle^2 \\
    \leq {} & 2GD\left(\frac{\Gamma_i}{\sqrt{\gamma_i}}+2\Gamma_i \right) + \frac{\Gamma_i^2}{2\beta\gamma_i} 
    \end{aligned}
\end{equation}
where the last step is due to $\sqrt{ab}\leq \frac{a}{2}+\frac{b}{2}$. To bound $N_s$, we present the following lemma. 
\begin{lemma}\label{lem:Ns:logT}
    Due to the construction of experts, UMA$2$ with Algorithm~\ref{alg:expert:option1} satisfies  
    \begin{equation*}
    N_s\leq s (\lfloor \log_2 s\rfloor+1) \left(3+2\left\lceil \log_2  T\right\rceil\right)
\end{equation*}
where $N_s$ is the number of experts created till round $s$. 
\end{lemma}
According to the definition of $\Gamma_i$ and $\gamma_i=\ln (4s^2)\geq 1$, we derive the following upper bound
\begin{equation}\label{eqn:hsT}
    \begin{aligned}
        \frac{\Gamma_i}{\sqrt{\gamma_i}}+2\Gamma_i  &= \Gamma_i\cdot \left(2+\frac{1}{\sqrt{\gamma_i}}\right) \leq 3\Gamma_i = 6\gamma_i +3\ln N_s + 3\ln\ln (9+36s) \leq 9\gamma_i +3\ln N_s \\
       &\leq 18\ln (2s) + 3\ln (2s)+3\ln \left(3+2\lceil \log_2 T \rceil \right) \leq h(s,T) \\
       \frac{\Gamma_i^2}{\gamma_i} & = \frac{\left(2\gamma_i +\ln N_s +\ln\ln (9+36s)\right)^2}{\gamma_i} \leq \frac{(3\gamma_i +\ln N_s)^2}{\gamma_i} = 9\gamma_i + 6\ln N_s + \frac{(\ln N_s)^2}{\gamma_i} \\
       &\leq 24\ln (2s) + 6\ln \left(3+2\lceil \log_2 T\rceil\right) + \frac{(\ln (2s) + \ln (3+2\lceil \log_2 T \rceil))^2}{2\ln (2s)} \\
       &\leq 24\ln (2s) + 7\ln \left(3+2\lceil \log_2 T\rceil\right) + \ln^2 \left(3+2\lceil \log_2 T\rceil\right) = h(s,T)
    \end{aligned}
\end{equation}
where we utilize $N_s\leq 2s (3+2\left\lceil \log_2  T\right\rceil)^2$ and  $\ln\ln (9+36s)\leq \gamma_i$, and set
\begin{equation*}
    h(s,T) = 24\ln (2s) + 7\ln \left(3+2\lceil \log_2 T\rceil\right) + \ln^2 \left(3+2\lceil \log_2 T\rceil\right).
\end{equation*}
Therefore, \eqref{eqn:meta-regret:exp} implies 
\begin{equation}\label{eqn:UMA2-opt1:meta-regret:exp}
    \sum_{t=r}^s f_t(\w_t) - \sum_{t=r}^s f_t(\w_{t,i}) \leq \left(2GD +\frac{1}{2\beta}\right)h(s,T). 
\end{equation}
Recall that we create multiple ONS experts over each interval $I=[r,s]\in \I$. And, there exits an expert $E_i$ with modulus $\hat{\alpha}^*\in \P_{exp}$ that satisfies $\hat{\alpha}^*\leq \alpha \leq 2\hat{\alpha}^*$. Therefore, we can bound the expert-regret by the theoretical guarantee of ONS~\citep[Theorem~2]{ML:Hazan:2007}: 
\begin{equation}
    \label{eqn:ONS:expert-regret}
    \sum_{t=r}^s f_t(\w_{t,i}) -\sum_{t=r}^s f_t(\w) \leq 5\left(\frac{1}{\hat{\alpha}^*}+GD\right)d\log (s-r+1)\leq 5\left(\frac{2}{\alpha}+GD\right)d\log (s-r+1). 
\end{equation}
Combining \eqref{eqn:UMA2-opt1:meta-regret:exp} and \eqref{eqn:ONS:expert-regret}, we obtain
\begin{equation}\label{eqn:UMA2-opt1:exp-bound}
        \sum_{t=r}^s f_t(\w_t) - \sum_{t=r}^s f_t(\w) 
        \leq \left(2GD +\frac{1}{2\beta}\right)c(s) + 5\left(\frac{2}{\alpha}+GD\right)d\log (s-r+1)
\end{equation}
where $c(\cdot)$ is defined in \eqref{eqn:c}. Next, we introduce the following property of GC intervals \citep[Lemma 1.2]{Adaptive:ICML:15}.
\begin{lemma}\label{lem:GC:intervals} For any interval $[p,q] \subseteq [T]$, it can be partitioned into two sequences of disjoint and consecutive intervals, denoted by $I_{-m},\ldots,I_0 \in \I$ and $I_1,\ldots,I_n \in \I$, such that
\[
|I_{-i}|/ |I_{-i+1}| \leq 1/2, \ \forall i \geq 1
\]
and
\[
|I_i|/|I_{i-1}| \leq 1/2, \ \forall i \geq 2.
\]
\end{lemma}
Based on the lemma above, we extend the above bound to any interval $[p,q]\subseteq [T]$.  Specifically, from Lemma \ref{lem:GC:intervals}, we conclude that $n \leq \lceil\log_2(q-p+2)\rceil$ because otherwise
\[
|I_1|+\cdots+|I_n| \geq 1+2+\ldots + 2^{n-1}= 2^n -1 > q-p+1 = |I|.
\]
Similarly, we have $m+1 \leq \lceil\log_2(q-p+2)\rceil$. 
Combining with \eqref{eqn:UMA2-opt1:exp-bound}, we have
\begin{equation}\label{eqn:UMA2:exp}
    \begin{aligned}
        &\sum_{t=p}^q f_t(\w_t) - \sum_{t=p}^q f_t(\w) =\sum_{i=-m}^n \sum_{t\in I_i} \left( f_t(\w_t) -  f_t(\w)\right)  \\
        \leq & \left(2GD +\frac{1}{2\beta}\right)h(q,T)b(p,q) + 5\left(\frac{2}{\alpha}+GD\right)d\log (q-p+1)b(p,q),
    \end{aligned}
\end{equation}
where $b(p,q)=2\lceil\log_2(q-p+2)\rceil$. 

When functions are $\lambda$-strongly convex during the interval $[r,s]$, the meta-regret can be bounded by
\begin{equation}\label{eqn:UMA2-opt1:meta-regret:str}
    \begin{aligned}
    &\sum_{t=r}^s f_t(\w_t) - \sum_{t=r}^s f_t(\w_{t,i}) 
    \overset{\eqref{eqn:def:str}}{\leq}  \sum_{t=r}^s \langle\nabla f_t(\w_t),\w_t-\w_{t,i}\rangle - \frac{\lambda}{2} \sum_{t=r}^s \Vert\w_t-\w_{t,i}\Vert^2 \\
    \overset{\eqref{eqn:UMA2-opt1:meta-regret:1}}{\leq} & {} 2GD\left(\frac{\Gamma_i}{\sqrt{\gamma_i}}+2\Gamma_i \right) + \sqrt{\frac{\Gamma_i^2}{\gamma_i}\sum_{t=r}^s \langle\nabla f_t(\w_t),\w_t-\w_{t,i}\rangle^2} - \frac{\lambda}{2} \sum_{t=r}^s \Vert\w_t-\w_{t,i}\Vert^2 \\
    \overset{\eqref{eqn:gradient}}{\leq} & {}  2GD\left(\frac{\Gamma_i}{\sqrt{\gamma_i}}+2\Gamma_i \right) + \sqrt{\frac{\Gamma_i^2G^2}{\gamma_i}\sum_{t=r}^s \Vert\w_t-\w_{t,i}\Vert^2} - \frac{\lambda}{2} \sum_{t=r}^s \Vert\w_t-\w_{t,i}\Vert^2\\
    \leq& {} 2GD\left(\frac{\Gamma_i}{\sqrt{\gamma_i}}+2\Gamma_i \right) + \frac{\Gamma_i^2G^2}{2\lambda\gamma_i} 
    \overset{\eqref{eqn:hsT}}{\leq}  \left(2GD+\frac{G^2}{2\lambda}\right)h(s,T).
    \end{aligned}
\end{equation}
Recall that we run multiple OGD experts with different $\hat{\lambda}\in \P_{str}$ over each interval $I=[r,s]\in \I$. And, there exists an expert $E_i$ with modulus $\hat{\lambda}^*\in \P_{str}$ that satisfies  $\hat{\lambda}^*\leq \lambda \leq 2\hat{\lambda}^*$. Therefore, we can directly bound the expert-regret by the theoretical guarantee of OGD for $\hat{\lambda}^*$-strongly convex functions~\citep[Theorem~1]{ML:Hazan:2007}: 
\begin{equation}
    \label{eqn:OGD-str:expert-regret}
    \sum_{t=r}^s f_t(\w_{t,i}) -\sum_{t=r}^s f_t(\w) \leq \frac{G^2}{2\hat{\lambda}^*} \left(1+\log (s-r+1)\right) \leq \frac{G^2}{\lambda} \left(1+\log (s-r+1)\right). 
\end{equation}
Combining \eqref{eqn:UMA2-opt1:meta-regret:str} and \eqref{eqn:OGD-str:expert-regret}, we have
\begin{equation*}
    \sum_{t=r}^s f_t(\w_{t}) -\sum_{t=r}^s f_t(\w) \leq \left(2GD+\frac{G^2}{2\lambda}\right)h(s,T) + \frac{G^2}{\lambda} \left(1+\log (s-r+1)\right). 
\end{equation*}
Next, we extend the above bound to any interval $[p,q]\subseteq [T]$. Following the analysis of \eqref{eqn:UMA2:exp}, we apply Lemma~\ref{lem:GC:intervals} and obtain
\begin{equation*}
\sum_{t=p}^q f_t(\w_t) - \sum_{t=p}^q f_t(\w) 
        \leq  \left(2GD+\frac{G^2}{2\lambda}\right)h(q,T)b(p,q)+\frac{G^2}{\lambda} \left(1+\log (q-p+1)\right)b(p,q)
\end{equation*}
which implies that $\SAReg (T,\tau)=O(\frac{1}{\lambda}\log \tau\log T)$ for $\lambda$-strongly convex functions.

Finally, we focus on general convex functions. When functions are convex, we have
\begin{equation}\label{eqn:UMA2-opt1:cvx:meta-regret}
    \begin{aligned}
    &\sum_{t=r}^s f_t(\w_t) - \sum_{t=r}^s f_t(\w_{t,i}) 
    \leq  \sum_{t=r}^s \langle\nabla f_t(\w_t),\w_t-\w_{t,i}\rangle  \\
    \overset{\eqref{eqn:UMA2-opt1:meta-regret:1}}{\leq} & 2GD\left(\frac{\Gamma_i}{\sqrt{\gamma_i}}+2\Gamma_i \right) + \sqrt{\frac{\Gamma_i^2}{\gamma_i}\sum_{t=r}^s \langle\nabla f_t(\w_t),\w_t-\w_{t,i}\rangle^2}  \\
    \overset{\eqref{eqn:hsT}}{\leq} & 2GDh(s,T)+GD\sqrt{c(s)(s-r+1)}.
    \end{aligned}
\end{equation}
Recall that we create an instance of OGD over each interval $I=[r,s]\in \I$. Therefore, we can bound the expert-regret by the theoretical guarantee of OGD~\citep[Theorem~1]{zinkevich-2003-online}: 
\begin{equation}
    \label{eqn:OGD:expert-regret}
    \sum_{t=r}^s f_t(\w_{t,i}) -\sum_{t=r}^s f_t(\w) \leq  \frac{D^2}{2\eta} + \frac{\eta (s-r+1)G^2}{2}\leq GD\sqrt{s-r+1} 
\end{equation}
where we set $\eta = D/(G\sqrt{s-r+1})$. Combining \eqref{eqn:UMA2-opt1:cvx:meta-regret} and \eqref{eqn:OGD:expert-regret}, we attain
\begin{equation*}
    \sum_{t=r}^s f_t(\w_t) - \sum_{t=r}^s f_t(\w) \leq 2GD h(s,T) +  GD\left(\sqrt{h(s,T)}+1  \right)\sqrt{s-r+1}. 
\end{equation*}
Next, we extend the above bound to any interval $[p,q]\subseteq [T]$. Let $J=[p,q]$. According to Lemma~\ref{lem:GC:intervals}, we have \citep[Theorem 1]{Adaptive:ICML:15}
\begin{equation} \label{eqn:UMA:inequality:5}
\sum_{i=-m}^n \sqrt{|I_i|} \leq 2 \sum_{i=0}^\infty \sqrt{2^{-i} |J|} \leq \frac{2\sqrt{2}}{\sqrt{2}-1} \sqrt{|J|} \leq 7 \sqrt{|J|}=7 \sqrt{q-p+1}.
\end{equation}
By applying this property, we have
\begin{equation*}
\begin{aligned}
    \sum_{t=p}^q f_t(\w_t) - \sum_{t=p}^q  f_t(\w) &\leq  \sum_{i=-m}^n \left( \sum_{t\in I_i} \left( f_t(\w_t)-f_t(\w)\right) \right) \\
    &\leq  2GD (m+1+n) h(q,T) + GD (\sqrt{h(q,T)}+1)\sum_{i=-m}^n \sqrt{|I_i|} \\
        &\overset{\eqref{eqn:UMA:inequality:5}}{\leq}  2GD h(q,T)b(p,q) +  7GD\left(\sqrt{h(q,T)} +1 \right)\sqrt{q-p+1},  
\end{aligned}
\end{equation*}
which implies $\SAReg (T,\tau)=O(\sqrt{\tau\log T})$ for general convex functions. 

\subsection{Proof of Lemma~\ref{lem:Ns:logT}}
Recall that we construct experts over GC intervals. According to the structure of GC intervals in Figure~\ref{fig:interval:saol}, we can find an integer $k$ that 
\begin{equation*}
    2^k \leq s \leq 2^{k+1}-1, 
\end{equation*}
which means that we construct $(k+1)$ types of intervals with lengths $1,2,\cdots,2^k$ till round $s$. As a result, the total number of intervals is at most $s\times (k+1) \leq 
 s(\lfloor \log_2 s \rfloor +1)$. According to the definition of $\P_{exp}$ and $\P_{str}$, the number of experts in each interval is $3+2\lceil \log_2 T\rceil$. Therefore, the number of experts created till round $s$ is bounded by
\begin{equation*}
    N_s \leq s\left(\lfloor \log_2 s \rfloor +1\right)\left(3+2\left\lceil \log_2 T\right\rceil\right). 
\end{equation*}

\subsection{Proof of Theorem~\ref{thm:main}}
Let $\w_{t,I}^\eta$ and $\ell_t^\eta (\cdot)$ be the output and the surrogate loss of the expert $E_I^\eta$ in the $t$-th round, and $\widehat{\w}_{t,I}^\eta$ and $\hat{\ell}_t^\eta (\cdot)$ be the output and the surrogate loss of the the expert $\widehat{E}_I^\eta$ in the $t$-th round. First, we start with the meta-regret of UMA$2$ with Algorithm~\ref{alg:expert:option2}. 
\begin{lemma}\label{lem:UMA:meta-regret}
    Under Assumptions~\ref{ass:1} and \ref{ass:2}, for any interval $I=[r,s]\in \I$ and any $\eta\in \mathcal{S}(s-r+1)$, the meta-regret of UMA$2$ with Algorithm~\ref{alg:expert:option2} satisfies
    \begin{equation*}
        \begin{aligned}
            \sum_{t=r}^s \ell_t^\eta (\w_t) - \sum_{t=r}^s \ell_t^\eta (\w^\eta_{t,I}) &\leq 2GD \eta c(s) + \frac{c(s)}{4}, \\
            \sum_{t=r}^s \hat{\ell}_t^\eta (\w_t) - \sum_{t=r}^s \hat{\ell}_t^\eta (\wh^\eta_{t,I}) &\leq 2GD \eta c(s) + \frac{c(s)}{4}, 
        \end{aligned}
    \end{equation*}
    where $c(\cdot)$ is defined in \eqref{eqn:c}. 
\end{lemma}
Then, combining with the expert-regret of $E_I^\eta$ and $\hat{E}_I^\eta$, we prove the following second-order regret of UMA$2$ over any interval. 
\begin{lemma}\label{lem:UMA:second-order:GC}
    Under Assumptions~\ref{ass:1} and \ref{ass:2}, for any interval $I=[r,s]\in \I$ and any $\w\in\Omega$, UMA$2$ with Algorithm~\ref{alg:expert:option2} satisfies
    \begin{align}
        \sum_{t=r}^s \langle \nabla f_t(\w_t),\w_t-\w\rangle &\leq \frac{3}{2}\sqrt{a(r,s)\sum_{t=r}^s \langle \nabla f_t(\w_t),\w_t-\w\rangle^2}+2GD (5a(r,s)+2c(s)), \label{eqn:UMA:second-order:GC:1} \\
        \sum_{t=r}^s \langle \nabla f_t(\w_t),\w_t-\w\rangle &\leq \frac{3}{2}G\sqrt{\hat{a}(r,s)\sum_{t=r}^s \Vert\w_t-\w\Vert^2}+2GD (5\hat{a}(r,s)+2c(s)) \label{eqn:UMA:second-order:GC:2} 
    \end{align}
    where $a(\cdot,\cdot)$, $\hat{a}(\cdot,\cdot)$, and $c(\cdot)$ are defined in \eqref{eqn:a}, \eqref{eqn:a:hat} and \eqref{eqn:c}, respectively. 
\end{lemma}
We define $\tau (r,s)=2GD (5a(r,s)+2c(s))$ to simplify the following analysis. Based on Lemma~\ref{lem:GC:intervals}, we extend Lemma~\ref{lem:UMA:second-order:GC} to any interval $[p,q] \subseteq [T]$. 

For any interval $[p,q] \subseteq [T]$, let  $I_{-m},\ldots,I_0 \in \I$ and $I_1,\ldots,I_n \in \I$ be the partition described in Lemma~\ref{lem:GC:intervals}. Then, we have
\begin{equation}\label{eqn:decom:regret}
\sum_{t=p}^q\langle \nabla f_t(\w_t),  \w_t -\w \rangle = \sum_{i=-m}^n \sum_{t\in I_i}\langle \nabla f_t(\w_t),  \w_t -\w \rangle.
\end{equation}
Combining with \eqref{eqn:UMA:second-order:GC:1}, we have
\begin{equation} \label{eqn:thm1:tmp}
\begin{split}
&\sum_{t=p}^q\langle \nabla f_t(\w_t),  \w_t -\w \rangle \\
\leq {} &   \sum_{i=-m}^n  \left(\frac{3}{2} \sqrt{a(p,q)\sum_{t\in I_i} \langle \nabla f_t(\w_t),  \w_t -\w \rangle^2 }+\tau (p,q)   \right) \\
= {} & (m+1+n) \tau (p,q)+ \frac{3}{2} \sqrt{a(p,q)}  \sum_{i=-m}^n \sqrt{\sum_{t\in I_i} \langle \nabla f_t(\w_t),  \w_t -\w \rangle^2 } \\
\leq {} & (m+1+n) \tau(p,q) + \frac{3}{2} \sqrt{(m+1+n)a(p,q)} \sqrt{\sum_{i=-m}^n \sum_{t\in I_i} \langle \nabla f_t(\w_t),  \w_t -\w \rangle^2}\\
= {} & (m+1+n) \tau(p,q) + \frac{3}{2} \sqrt{(m+1+n)a(p,q)} \sqrt{\sum_{t=p}^q\langle \nabla f_t(\w_t),  \w_t -\w \rangle^2}\\
\leq {} & \tau(p,q) b(p,q) + \frac{3}{2} \sqrt{a(p,q) b(p,q)} \sqrt{\sum_{t=p}^q\langle \nabla f_t(\w_t),  \w_t -\w \rangle^2}.
\end{split}
\end{equation}

Next, we  define $\hat{\tau} (r,s)=2GD (5\hat{a}(r,s)+2c(s))$ and prove \eqref{eqn:UMA:inequality:1} in a similar way. Combining \eqref{eqn:UMA:second-order:GC:2} with \eqref{eqn:decom:regret}, we have
\begin{equation} \label{eqn:UMA:inequality:3}
\begin{split}
&\sum_{t=p}^q\langle \nabla f_t(\w_t),  \w_t -\w \rangle \\
\leq {} &  \sum_{i=-m}^n  \left(\frac{3}{2} G\sqrt{\hat{a}(p,q)\sum_{t\in I_i} \|\w_t -\w \|^2 }+ \hat{\tau}(p,q) \right) \\
= {} & (m+1+n) \hat{\tau}(p,q) + \frac{3}{2} G\sqrt{\hat{a}(p,q)}  \sum_{i=-m}^n \sqrt{\sum_{t\in I_i} \|\w_t -\w \|^2 } \\
\leq & (m+1+n)  \hat{\tau}(p,q) + \frac{3}{2}G \sqrt{(m+1+n)\hat{a}(p,q)} \sqrt{\sum_{i=-m}^n \sum_{t\in I_i} \|\w_t -\w \|^2}\\
= {} & (m+1+n)  \hat{\tau}(p,q)+ \frac{3}{2}G \sqrt{(m+1+n)\hat{a}(p,q)} \sqrt{\sum_{t=p}^q\|\w_t -\w \|^2}\\
\leq {} & \ \hat{\tau}(p,q) b(p,q) + \frac{3}{2}G \sqrt{\hat{a}(p,q) b(p,q)} \sqrt{\sum_{t=p}^q\|\w_t -\w \|^2}.
\end{split}
\end{equation}
Following the analysis of Theorem~\ref{thm:UMA2-op1} for general convex functions, we move to prove (\ref{eqn:UMA:inequality:2}) as follows. \eqref{eqn:UMA:inequality:3} implies that
\begin{equation} \label{eqn:UMA:inequality:4}
\begin{split}
&\sum_{t=p}^q\langle \nabla f_t(\w_t),  \w_t -\w \rangle 
\leq {} \sum_{i=-m}^n  \left(\frac{3}{2} G\sqrt{\hat{a}(p,q)\sum_{t\in I_i} \|\w_t -\w \|^2 }+ \hat{\tau}(p,q) \right) \\
= {} &  (m+1+n) \hat{\tau}(p,q) + \frac{3}{2} G\sqrt{\hat{a}(p,q)}  \sum_{i=-m}^n \sqrt{\sum_{t\in I_i} \|\w_t -\w \|^2 } \\
\leq {} & \hat{\tau}(p,q) b(p,q)+ \frac{3}{2} DG\sqrt{\hat{a}(p,q)}   \sum_{i=-m}^n \sqrt{|I_i|} \\
 \overset{\eqref{eqn:UMA:inequality:5}}{\leq} {} &\hat{\tau}(p,q) b(p,q)+ \frac{21}{2} DG\sqrt{\hat{a}(p,q) (q-p+1)}
\end{split}
\end{equation}
When all the online functions are $\alpha$-exp-concave, Lemma~\ref{lem:exp} implies
\[
\begin{split}
&\sum_{t=p}^q f_t(\w_t) -  \sum_{t=p}^q f_t(\w) 
\leq   \sum_{t=p}^q  \langle \nabla f_t(\w_t), \w_t -\w \rangle  - \frac{\beta}{2} \sum_{t=p}^q \langle \nabla f_t(\w_t), \w_t -\w \rangle^2\\
\overset{\text{(\ref{eqn:thm1:tmp})}}{\leq} & \tau (p,q) b(p,q) + \frac{3}{2} \sqrt{a(p,q) b(p,q)} \sqrt{\sum_{t=p}^q\langle \nabla f_t(\w_t),  \w_t -\w \rangle^2}
 -\frac{\beta}{2} \sum_{t=p}^q \langle \nabla f_t(\w_t), \w_t -\w \rangle^2\\
\leq {} & \left(\frac{9}{8\beta}a(p,q)+ \tau (p,q)\right)  b(p,q) \\
= {} & O\left(\frac{d\log q\log (q-p)}{\alpha}\right)
\end{split}
\]
where the last inequality is due to $\sqrt{ab}\leq \frac{a}{2}+\frac{b}{2}$. 

When all the online functions are $\lambda$-strongly convex, Definition~\ref{def:strong} implies
\[
\begin{split}
&\sum_{t=p}^q f_t(\w_t) -  \sum_{t=p}^q f_t(\w) 
\leq  \sum_{t=p}^q  \langle \nabla f_t(\w_t), \w_t -\w \rangle  - \frac{\lambda}{2} \sum_{t=p}^q \| \w_t -\w \|^2\\
\overset{\text{(\ref{eqn:UMA:inequality:3})}}{\leq} {} &  \hat{\tau}(p,q) b(p,q) + \frac{3}{2}G \sqrt{\hat{a}(p,q) b(p,q)} \sqrt{\sum_{t=p}^q\|\w_t -\w \|^2}  -\frac{\lambda}{2} \sum_{t=p}^q \| \w_t -\w \|^2\\
\leq {} & \left(\frac{9G^2}{8\lambda}\hat{a}(p,q)+\hat{\tau}(p,q) \right)  b(p,q)=O\left(\frac{\log q\log (q-p)}{\lambda}\right).
\end{split}
\]

\subsection{Proof of Lemma~\ref{lem:UMA:meta-regret}}
Let $\ell_{t,I}^\eta$, $\gamma_I^\eta$ and $\Gamma_I^\eta$ be the updating parameters of the expert $E_I^\eta$. The following theoretical guarantee is a special case of Lemma~\ref{lem:TUMA-Comp:meta-regret} when $m_{t,I}=0$. 
\begin{lemma}\label{lem:UMA:meta-regret:first}
    Under Assumptions~\ref{ass:1} and \ref{ass:2}, for any interval $I=[r,s]\in \I$ and any $\eta\in \mathcal{S}(s-r+1)$, the meta-regret of UMA$2$ with Algorithm~\ref{alg:expert:option2} satisfies
    \begin{equation*}
        \sum_{t=r}^s \ell_{t} - \sum_{t=r}^s \ell^\eta_{t,I} \leq \frac{\Gamma^\eta_I}{\sqrt{\gamma^\eta_I}} \sqrt{1+\sum_{t=r}^s (\ell_t-\ell^\eta_{t,I})^2} + 2\Gamma^\eta_I
    \end{equation*}
    where $\Gamma^\eta_I = 2\gamma^\eta_I+\ln N_s + \ln \ln \left(9+36s\right)$ and $N_s$ is the number of experts created till round $s$.  
\end{lemma}
According to the definition of $\ell_t$ and $\ell^\eta_{t,I}$, we have
\begin{equation}\label{eqn:UMA:meta-regret:01}
\begin{aligned}
    \sum_{t=r}^s \langle\nabla f_t(\w_t),\w_t-\w^\eta_{t,I}\rangle &\leq  \frac{\Gamma^\eta_I}{\sqrt{\gamma^\eta_I}} \sqrt{4G^2D^2+\sum_{t=r}^s \langle\nabla f_t(\w_t),\w_t-\w^\eta_{t,I}\rangle^2} + 4GD\Gamma^\eta_I \\
    &\leq 2GD\left(\frac{\Gamma^\eta_I}{\sqrt{\gamma^\eta_I}}+2\Gamma^\eta_I \right) + \sqrt{\frac{{\Gamma^\eta_I}^2}{\gamma^\eta_I}\sum_{t=r}^s \langle\nabla f_t(\w_t),\w_t-\w^\eta_{t,I}\rangle^2} \\
    &\leq 2GD\left(\frac{\Gamma^\eta_I}{\sqrt{\gamma^\eta_I}}+2\Gamma^\eta_I \right) + \frac{{\Gamma^\eta_I}^2}{4\eta\gamma^\eta_I} + \eta \sum_{t=r}^s \langle\nabla f_t(\w_t),\w_t-\w^\eta_{t,I}\rangle^2 
\end{aligned}
\end{equation}
where the second inequality is due to $\sqrt{a+b}\leq \sqrt{a}+\sqrt{b}$, and the last inequality is due to $\sqrt{ab}\leq \frac{a}{2}+\frac{b}{2}$. To bound $N_s$, we present the following lemma. 
\begin{lemma}\label{lem:Ns}
    Due to the construction of experts, UMA$2$ with Algorithm~\ref{alg:expert:option2}, UMA$3$, and UMA-Comp satisfy 
    \begin{equation*}
    N_s\leq 2s (\lfloor \log_2 s\rfloor+1) \left(1+\left\lceil \frac{1}{2} \log_2 s\right\rceil\right)\leq 4s^2
\end{equation*}
where $N_s$ is the number of experts created till round $s$. 
\end{lemma}
According to the definition of $\Gamma_I^\eta$ and $\gamma_I^\eta=\ln 4s^2\geq 1$,  we derive the following upper bound 
\begin{equation}\label{eqn:UMA2-opt1:gamma-bound}
    \begin{aligned}
        \frac{\Gamma_I^\eta}{\sqrt{\gamma_I^\eta}} + 2\Gamma_I^\eta &= \Gamma_I^\eta \cdot \left(2+\frac{1}{\sqrt{\gamma_I^\eta}}\right) \leq 3 \Gamma_I^\eta = 6 \gamma^\eta_I+3\ln N_s + 3\ln \ln \left(9+36s\right) \\
        &\leq 9 \gamma^\eta_I + 3\ln N_s \leq 24\ln 2s\leq c(s) \\
        \frac{{\Gamma_I^\eta}^2}{\gamma_I^\eta} &= \frac{(2\gamma^\eta_I+\ln N_s+\ln \ln \left(9+36s\right))^2}{\gamma_I^\eta} \leq \frac{(3\gamma^\eta_I+\ln N_s)^2}{\gamma_I^\eta}  \\
        & = 9 \gamma^\eta_I + 6\ln N_s + \frac{(\ln N_s)^2}{\gamma_I^\eta} \leq 32 \ln 2s =c(s)
    \end{aligned}
\end{equation}
where we utilize $3\ln \ln \left(9+36s\right)\leq \gamma_I^\eta$, and set
\begin{equation*}
    c(s) = 32\ln 2s. 
\end{equation*}
Therefore, \eqref{eqn:UMA:meta-regret:01} implies 
\begin{equation}\label{eqn:UMA:meta-regret:1}
    \sum_{t=r}^s \langle\nabla f_t(\w_t),\w_t-\w^\eta_{t,I}\rangle \leq \left(2GD+\frac{1}{4\eta}\right)c(s) + \eta \sum_{t=r}^s \langle\nabla f_t(\w_t),\w_t-\w^\eta_{t,I}\rangle^2 . 
\end{equation}
Rearranging the second-order term in \eqref{eqn:UMA:meta-regret:1} and multiplying both sides of the inequality by $\eta$, we have 
\begin{equation*}
    \ell_{t}^{\eta} (\w^\eta_{t,I}) = \eta \sum_{t=r}^s \langle\nabla f_t(\w_t),\w_t-\w^\eta_{t,I}\rangle - \eta^2 \sum_{t=r}^s \langle\nabla f_t(\w_t),\w_t-\w^\eta_{t,I}\rangle^2 
    \leq \left( 2\eta GD+\frac{1}{4} \right) c(s). 
\end{equation*}
For the expert $\widehat{E}^\eta_I$, we can obtain a similar bound
\begin{equation}\label{eqn:UMA:meta-regret:2}
\begin{aligned}
    \sum_{t=r}^s \langle\nabla f_t(\w_t),\w_t-\wh^\eta_{t,I}\rangle 
    &\leq \left( 2GD+\frac{1}{4\eta} \right) c(s) +\eta \sum_{t=r}^s \langle\nabla f_t(\w_t),\w_t-\wh^\eta_{t,I}\rangle^2 \\
    &\leq \left( 2GD+\frac{1}{4\eta} \right) c(s) +\eta \sum_{t=r}^s \Vert \nabla f_t(\w_t)\Vert^2 \Vert \w_t-\wh^\eta_{t,I}\Vert^2
\end{aligned}
\end{equation}
which implies that
\begin{equation*}
    \hat{\ell}_{t}^{\eta} (\w^\eta_{t,I}) = \eta \sum_{t=r}^s \langle\nabla f_t(\w_t),\w_t-\wh^\eta_{t,I}\rangle - \eta^2 \sum_{t=r}^s \Vert\nabla f_t(\w_t)\Vert \Vert\w_t-\wh^\eta_{t,I}\Vert^2 \leq \left( 2\eta GD+\frac{1}{4} \right) c(s). 
\end{equation*}

\subsection{Proof of Lemma~\ref{lem:UMA:second-order:GC}}
The analysis is similar to the proofs of Theorem 7 of \citet{NIPS2016_6268} and Theorem 1 of \citet{Adaptive:Maler}. From Lemma 5 of \citet{NIPS2016_6268}, we have the following expert-regret of $E_I^\eta$ .
\begin{lemma}\label{lem:expert} Under Assumptions~\ref{ass:1} and \ref{ass:2}, for any interval $I=[r,s] \in \I$, any  $\w \in \Omega$ and any $\eta \in \S(s-r+1)$, the expert-regret of $E_I^\eta$ satisfies
\[
\begin{split}
\sum_{t=r}^s \ell_t^\eta(\w_{t,I}^\eta) - \sum_{t=r}^s \ell_t^\eta(\w)  \leq & \frac{\|\w_{r,I}^\eta-\w\|^{2}}{2 D^{2}}+\frac{1}{2} \ln \operatorname{det}\left(I+2 \eta^{2} D^{2} \sum_{t=r}^{s} M_{t}\right), \ \end{split}
\]
\end{lemma}
where $M_{t}=\g_{t} \g_{t}^{\top}$ and $\g_{t}=\nabla f_{t}\left(\w_{t}\right)$. Based on Lemma~\ref{lem:expert}, we have
\[
\begin{split}
\sum_{t=r}^s \ell_t^\eta(\w_{t,I}^\eta) - \sum_{t=r}^s \ell_t^\eta(\w)
 \overset{\text{(\ref{eqn:domain})}}{\leq}  & \frac{1}{2}+\frac{1}{2} \sum_{i=1}^{d} \ln \left(1+2 \eta^{2} D^{2} \lambda_{i}\left(\sum_{t=r}^{s} \g_{t} \g_{t}^{\top}\right)\right) \\
\leq & \frac{1}{2}+\frac{d}{2} \ln \left(1+\frac{2 \eta^{2} D^{2}}{d} \sum_{i=1}^{d} \lambda_{i}\left(\sum_{t=r}^{s} \g_{t} \g_{t}^{\top}\right)\right) \\
= & \frac{1}{2}+\frac{d}{2} \ln \left(1+\frac{2 \eta^{2} D^{2}}{d} \tr\left(\sum_{t=r}^{s} \g_{t} \g_{t}^{\top}\right)\right) \\
= & \frac{1}{2}+\frac{d}{2} \ln \left(1+\frac{2 \eta^{2} D^{2}}{d} \sum_{t=r}^{s}\left\|g_{t}\right\|_{2}^{2}\right) \\
\leq & \frac{1}{2}+\frac{d}{2} \ln \left(1+\frac{2}{25 d}(s-r+1)\right)
\end{split}
\]
where the second inequality is by the concavity of the function $\ln x$ and Jensen's inequality and the last inequality is due to $\eta \leq \frac{1}{5 D G} $.
Combining the regret bounds in Lemmas~\ref{lem:UMA:meta-regret} and \ref{lem:expert}, we have
\[
\begin{split}
- \sum_{t=r}^s \ell_t^\eta(\w) = &  \eta \sum_{t=r}^s \langle \nabla f_t(\w_t),  \w_t -\w \rangle - \eta^2 \sum_{t=r}^s \langle \nabla f_t(\w_t),  \w_t -\w \rangle^2 \\
\leq & 2GD\eta c(s)+\frac{c(s)}{4}+\frac{1}{2}+\frac{d}{2} \ln \left(1+\frac{2}{25 d}(s-r+1)\right)
\end{split}
\]
for any $\eta \in \S(s-r+1)$. Therefore, we have
\begin{equation} \label{eqn:lem:second:1}
    \sum_{t=r}^s\langle \nabla f_t(\w_t),  \w_t -\w \rangle \leq  \frac{a (r,s)}{\eta} + \eta \sum_{t=r}^s \langle \nabla f_t(\w_t),  \w_t -\w \rangle^2 +2GD c(s)
\end{equation}
for any $\eta \in \S(s-r+1)$, where $a(r,s)=\frac{c(s)}{4}+\frac{1}{2}+\frac{d}{2} \ln \left(1+\frac{2}{25 d}(s-r+1)\right)$.

Note that the optimal $\eta_*$ that minimizes the R.H.S.~of (\ref{eqn:lem:second:1}) is
\[
\eta_*=\sqrt{\frac{a(r,s)}{\sum_{t=r}^s \langle \nabla f_t(\w_t),  \w_t -\w \rangle^2}} \geq \frac{\sqrt{2}}{GD \sqrt{s-r+1}}.
\]
where the inequality is due to $c(s)/4\geq 2$. Recall that
\[
\S(s-r+1)= \left\{ \frac{2^{-i}}{5 DG} \ \left |\ i=0,1,\ldots, \left\lceil \frac{1}{2} \log_2 (s-r+1)\right\rceil \right. \right\}.
\]
If $\eta_* \leq \frac{1}{5DG}$, there must exist an $\eta \in \S(s-r+1)$ such that
\[
\eta \leq \eta_* \leq 2\eta.
\]
Then, (\ref{eqn:lem:second:1}) implies
\begin{equation} \label{eqn:lem:second:2}
\begin{split}
\sum_{t=r}^s\langle \nabla f_t(\w_t),  \w_t -\w \rangle \leq &  \frac{ a(r,s)}{2\eta_*} + \eta_* \sum_{t=r}^s \langle \nabla f_t(\w_t),  \w_t -\w \rangle^2+2GD c(s) \\
 =& \frac{3}{2} \sqrt{a(r,s)\sum_{t=r}^s \langle \nabla f_t(\w_t),  \w_t -\w \rangle^2 }+2GD c(s).
\end{split}
\end{equation}
On the other hand, if $\eta_* \geq \frac{1}{5DG}$,  we have
\[
\sum_{t=r}^s \langle \nabla f_t(\w_t),  \w_t -\w \rangle^2 \leq  25 D^2 G^2 \Xi (r,s).
\]
Then, (\ref{eqn:lem:second:1}) with $\eta=\frac{1}{5DG}$ implies
\begin{equation} \label{eqn:lem:second:3}
\sum_{t=r}^s\langle \nabla f_t(\w_t),  \w_t -\w \rangle \leq   10 DG a(r,s)+2GD c(s).
\end{equation}
We complete the proof of \eqref{eqn:UMA:second-order:GC:1} by combining \eqref{eqn:lem:second:2} and \eqref{eqn:lem:second:3}. Next, we prove \eqref{eqn:UMA:second-order:GC:2} in a similar way.

As proven in Lemma~2 of \citet{Adaptive:Maler}, the surrogate loss $\hat{\ell}_t^\eta (\cdot)$ in \eqref{eqn:ell:hat} is $2\eta^2G^2$-strongly convex, and its gradient is bounded by
\begin{equation*}
    \Vert \nabla \hat{\ell}_t^\eta(\w) \Vert^2 = \Vert \eta \nabla f_t(\w_t) + 2\eta^2 G^2 (\w-\w_t) \Vert^2 \leq \eta^2G^2 (1+2\eta GD)^2 \overset{\eqref{eqn:metagrad:4}}{\leq} 4\eta^2 G^2. 
\end{equation*}
According to the theoretical guarantee of OGD for strongly convex functions~\citep[Theorem~1]{ML:Hazan:2007}, the expert-regret of $\widehat{E}_I^\eta$ satisfies
\begin{equation}\label{eqn:UMA2:surr:str:expert-regret}
    \sum_{t=r}^s \hat{\ell}_t^\eta(\wh_{t,I}^\eta) - \sum_{t=r}^s \hat{\ell}_t^\eta(\w)   \leq 1+\log (s-r+1). 
\end{equation}
Combining the regret bound in Lemmas~\ref{lem:UMA:meta-regret} and \eqref{eqn:UMA2:surr:str:expert-regret}, we have
\[
\begin{split}
- \sum_{t=r}^s \hat{\ell}_t^\eta(\w) = & \eta \sum_{t=r}^s \langle \nabla f_t(\w_t),  \w_t -\w \rangle - \eta^2 \|\nabla f_t(\w_t)\|^2 \sum_{t=r}^s \|\w_t -\w \|^2 \\
\leq & 2GD \eta c(s)+\frac{c(s)}{4}+1 + \log (s-r+1)
\end{split}
\]
for any $\eta \in \S(s-r+1)$. Thus,
\begin{equation} \label{eqn:lem:uma:1}
\begin{split}
\sum_{t=r}^s\langle \nabla f_t(\w_t),  \w_t -\w \rangle \leq & \frac{\hat{a}(r,s)}{\eta} + \eta \| \nabla f_t(\w_t)\|^2 \sum_{t=r}^s \|\w_t -\w \|^2 + 2GD c(s) \\
\overset{\text{(\ref{eqn:gradient})}}{\leq} & \frac{\hat{a}(r,s)}{\eta} + \eta G^2 \sum_{t=r}^s \|\w_t -\w \|^2 + 2GD c(s)
\end{split}
\end{equation}
for any $\eta \in \S(s-r+1)$, where $\hat{a}(r,s)= \frac{c(s)}{4}+1 + \log (s-r+1)$.

Note that the optimal $\eta_*$ that minimizes the R.H.S.~of (\ref{eqn:lem:uma:1}) is
\[
\eta_*=\sqrt{\frac{\hat{a}(r,s)}{G^2 \sum_{t=r}^s \|\w_t -\w \|^2}} \geq \frac{\sqrt{2}}{GD \sqrt{s-r+1}}.
\]
Recall that
\[
\S(s-r+1)= \left\{ \frac{2^{-i}}{5 DG} \ \left |\ i=0,1,\ldots, \left\lceil \frac{1}{2} \log_2 (s-r+1)\right\rceil \right. \right\}.
\]
If $\eta_* \leq \frac{1}{5DG}$, there must exist an $\eta \in \S(s-r+1)$ such that $\eta \leq \eta_* \leq 2\eta$. 
Then, (\ref{eqn:lem:uma:1}) implies
\begin{equation} \label{eqn:lem:uma:2}
\begin{aligned}
    \sum_{t=r}^s\langle \nabla f_t(\w_t),  \w_t -\w \rangle &\leq  \frac{ \hat{a}(r,s)}{2\eta_*} + \eta_* G^2\sum_{t=r}^s \|  \w_t -\w \|^2 +2GD c(s) \\
    &= \frac{3}{2} G\sqrt{\hat{a}(r,s)\sum_{t=r}^s \|\w_t -\w \|^2 }+2GD c(s).
\end{aligned}
\end{equation}
On the other hand, if $\eta_* \geq \frac{1}{5DG}$,  we have
\[
\sum_{t=r}^s \|  \w_t -\w \|^2 \leq  25  D^2  \hat{a}(r,s).
\]
Then, (\ref{eqn:lem:uma:1}) with $\eta=\frac{1}{5DG}$ implies
\begin{equation} \label{eqn:lem:uma:3}
\sum_{t=r}^s\langle \nabla f_t(\w_t),  \w_t -\w \rangle \leq  10 GD \hat{a}(r,s)+2GD c(s).
\end{equation}
We obtain \eqref{eqn:UMA:second-order:GC:2} by combining (\ref{eqn:lem:uma:2}) and (\ref{eqn:lem:uma:3}). 

\subsection{Proof of Lemma~\ref{lem:Ns}}
The analysis is similar to Lemma~\ref{lem:Ns:logT}, and the total number of intervals is at most $s(\lfloor \log_2 s \rfloor +1)$. For the construction of experts, UMA$3$ and UMA-Comp create one expert over each GC interval, while UMA$2$ with Algorithm~\ref{alg:expert:option2} create multiple experts. According to the definition of $\S(I)$ in \eqref{eqn:metagrad:4}, the number of experts in each GC interval is bounded by $2(1+\lceil \frac{1}{2}\log_2 s\rceil) $. 
Therefore, the number of experts created till round $s$ is bounded by
\begin{equation*}
    N_s \leq 2s\left(\lfloor \log_2 s \rfloor +1\right)\left(1+\left\lceil \frac{1}{2}\log_2 s\right\rceil\right)\leq 4s^2. 
\end{equation*}

\subsection{Proof of Theorem~\ref{thm:TUMA}}
Let $\w_{t,I}$ be the output of the expert $E_I$ in the $t$-th round, and $\ell_{t,I}$, $\gamma_I$ and $\Gamma_I$ be the updating parameters of the expert $E_I$. We start with the meta-regret of UMA$3$. The following theoretical guarantee is a special case of Lemma~\ref{lem:TUMA-Comp:meta-regret} when $m_{t,I}=0$. 
\begin{lemma}\label{lem:TUMA:meta-regret}
    Under Assumptions~\ref{ass:1} and \ref{ass:2}, for any interval $I=[r,s]\in \I$, the meta-regret of UMA$3$ with respect to the expert $E_I$ satisfies
    \begin{equation*}
        \sum_{t=r}^s \ell_{t} - \sum_{t=r}^s \ell_{t,I} \leq \frac{\Gamma_I}{\sqrt{\gamma_I}} \sqrt{1+\sum_{t=r}^s (\ell_t-\ell_{t,I})^2} + 2\Gamma_I
    \end{equation*}
    where $\Gamma_I = 2\gamma_I+\ln N_s + \ln \ln \left(9+36s\right)$ and $N_s$ is the number of experts created till round $s$.  
\end{lemma}
According to the definition of $\ell_t$ and $\ell_{t,I}$, we have
\begin{equation}\label{eqn:TUMA:meta-regret:1}
\begin{aligned}
    \sum_{t=r}^s \langle\nabla f_t(\w_t),\w_t-\w_{t,I}\rangle &\leq  \frac{\Gamma_I}{\sqrt{\gamma_I}} \sqrt{4G^2D^2+\sum_{t=r}^s \langle\nabla f_t(\w_t),\w_t-\w_{t,I}\rangle^2} + 4GD\Gamma_I \\
    &\leq 2GD\left(\frac{\Gamma_I}{\sqrt{\gamma_I}}+2\Gamma_I \right) + \sqrt{\frac{\Gamma_I^2}{\gamma_I}\sum_{t=r}^s \langle\nabla f_t(\w_t),\w_t-\w_{t,I}\rangle^2}
\end{aligned}
\end{equation}
where the last step is due to $\sqrt{a+b}\leq \sqrt{a}+\sqrt{b}$. Recall that we employ Maler to minimize $f_t(\cdot)$ during each interval $I=[r,s]\in \I$. Therefore, we can directly use the theoretical guarantee of Maler to bound the expert-regret \citep[Theorem~1 and Corollary~2]{Adaptive:Maler}. 
\begin{lemma}
    Under Assumptions~\ref{ass:1} and \ref{ass:2}, for any interval $I=[r,s]\in \I$ and any $\w\in\Omega$, if functions are general convex, the expert-regret of UMA$3$ satisfies
    \begin{equation}\label{eqn:Maler:convex}
        \sum_{t=r}^s f_t(\w_{t,I}) - \sum_{t=r}^s f_t(\w) \leq \left(2\ln 3 +\frac{3}{2}\right)GD \sqrt{s-r+1}. 
    \end{equation}
    If functions are $\lambda$-strongly convex, the expert-regret of UMA$3$ satisfies 
    \begin{equation}\label{eqn:Maler:str-cvx}
        \sum_{t=r}^s f_t(\w_{t,I}) - \sum_{t=r}^s f_t(\w) \leq \left(10GD+\frac{9G^2}{2\lambda}\right)\left(\Xi(r,s) + 1+ \log (s-r+1)\right) 
    \end{equation}
    where $\Xi(r,s) = 2\ln (\frac{\sqrt{3}}{2}\log_2 (s-r+1)+3\sqrt{3})$. If functions are $\alpha$-exp-concave, the expert-regret of UMA$3$ satisfies
    \begin{equation}\label{eqn:Maler:exp}
        \sum_{t=r}^s f_t(\w_{t,I}) - \sum_{t=r}^s f_t(\w) \leq \left(10GD+\frac{9}{2\beta}\right)\left(\Xi(r,s) + 10d\log (s-r+1)\right) 
    \end{equation}
    where $\beta=\frac{1}{2}\min \{\frac{1}{4GD},\alpha\}$. 
\end{lemma}
When functions are $\alpha$-exp-concave during the interval $[r,s]$, we have
\begin{equation}\label{eqn:TUMA:meta-regret:exp}
    \begin{aligned}
    &\sum_{t=r}^s f_t(\w_t) - \sum_{t=r}^s f_t(\w_{t,I}) 
    \overset{\eqref{eqn:lem:exp}}{\leq} \sum_{t=r}^s \langle\nabla f_t(\w_t),\w_t-\w_{t,I}\rangle - \frac{\beta}{2} \sum_{t=r}^s \langle\nabla f_t(\w_t),\w_t-\w_{t,I}\rangle^2 \\
    \overset{\eqref{eqn:TUMA:meta-regret:1}}{\leq} {} & 2GD\left(\frac{\Gamma_I}{\sqrt{\gamma_I}}+2\Gamma_I \right) + \sqrt{\frac{\Gamma_I^2}{\gamma_I}\sum_{t=r}^s \langle\nabla f_t(\w_t),\w_t-\w_{t,I}\rangle^2} - \frac{\beta}{2} \sum_{t=r}^s \langle\nabla f_t(\w_t),\w_t-\w_{t,I}\rangle^2 \\
    \leq {}& 2GD\left(\frac{\Gamma_I}{\sqrt{\gamma_I}}+2\Gamma_I \right) + \frac{\Gamma_I^2}{2\beta\gamma_I} \\
    \overset{\eqref{eqn:UMA2-opt1:gamma-bound}}{\leq} {} &  \left(2GD +\frac{1}{2\beta}\right)c(s)
    \end{aligned}
\end{equation}
where the third inequality is due to $\sqrt{ab}\leq \frac{a}{2}+\frac{b}{2}$ and $c(\cdot)$ is defined in \eqref{eqn:c}. Combining \eqref{eqn:Maler:exp} and \eqref{eqn:TUMA:meta-regret:exp}, we obtain
\begin{equation*}
        \sum_{t=r}^s f_t(\w_t) - \sum_{t=r}^s f_t(\w) 
        \leq \left(2GD +\frac{1}{2\beta}\right)c(s) + \left(10GD+\frac{9}{2\beta}\right)\left(\Xi(r,s) + 10d\log (s-r+1)\right)
\end{equation*}
where $\Xi(r,s) = 2\ln (\frac{\sqrt{3}}{2}\log_2 (s-r+1)+3\sqrt{3})$. Finally, we follow the analysis of Theorem~\ref{thm:main} to extend the above bound to any interval $[p,q]\subseteq [T]$. Based on Lemma~\ref{lem:GC:intervals}, we have 
\begin{equation*}
    \begin{aligned}
        &\sum_{t=p}^q f_t(\w_t) - \sum_{t=p}^q f_t(\w) \\
        \leq & \left(2GD +\frac{1}{2\beta}\right)c(q)b(p,q) + \left(10GD+\frac{9}{2\beta}\right)\left(\Xi(q,p) + 10d\log (q-p+1)\right)b(p,q)\\
        =& O\left(\frac{d\log q\log (q-p)}{\alpha}\right). 
    \end{aligned}
\end{equation*}

When functions are $\lambda$-strongly convex during the interval $[r,s]$, the meta-regret can be bounded by
\begin{equation*}
    \begin{aligned}
    &\sum_{t=r}^s f_t(\w_t) - \sum_{t=r}^s f_t(\w_{t,I}) \\
    \overset{\eqref{eqn:def:str}}{\leq} & \sum_{t=r}^s \langle\nabla f_t(\w_t),\w_t-\w_{t,I}\rangle - \frac{\lambda}{2} \sum_{t=r}^s \Vert\w_t-\w_{t,I}\Vert^2 \\
    \overset{\eqref{eqn:TUMA:meta-regret:1}}{\leq} & 2GD\left(\frac{\Gamma_I}{\sqrt{\gamma_I}}+2\Gamma_I \right) + \sqrt{\frac{\Gamma_I^2}{\gamma_I}\sum_{t=r}^s \langle\nabla f_t(\w_t),\w_t-\w_{t,I}\rangle^2} - \frac{\lambda}{2} \sum_{t=r}^s \Vert\w_t-\w_{t,I}\Vert^2 \\
    \leq& 2GD\left(\frac{\Gamma_I}{\sqrt{\gamma_I}}+2\Gamma_I \right) + \frac{\Gamma_I^2G^2}{2\lambda\gamma_I} \\
    \overset{\eqref{eqn:UMA2-opt1:gamma-bound}}{\leq} & \left(2GD+\frac{G^2}{2\lambda}\right)c(s).
    \end{aligned}
\end{equation*}
Combining with \eqref{eqn:Maler:str-cvx}, we have
\begin{equation*}
    \sum_{t=r}^s f_t(\w_t) - \sum_{t=r}^s f_t(\w)\leq \left(2GD+\frac{G^2}{2\lambda}\right)c(s)+\left(10GD+\frac{9G^2}{2\lambda}\right)\left(\Xi(r,s) + 1+ \log (s-r+1)\right). 
\end{equation*}
Next, we extend the above bound to any interval $[p,q]\subseteq [T]$ by applying Lemma~\ref{lem:GC:intervals},
\begin{equation*}
    \begin{aligned}
        &\sum_{t=p}^q f_t(\w_t) - \sum_{t=p}^q f_t(\w) \\
        \leq & \left(2GD+\frac{G^2}{2\lambda}\right)c(q)b(p,q)+\left(10GD+\frac{9G^2}{2\lambda}\right)\left(\Xi(p,q) + 1+ \log (q-p+1)\right)b(p,q) \\
        =& O\left(\frac{\log q\log (q-p)}{\lambda}\right). 
    \end{aligned}
\end{equation*}

Finally, we focus on general convex functions. When functions are convex, we have
\begin{equation*}
    \begin{aligned}
    &\sum_{t=r}^s f_t(\w_t) - \sum_{t=r}^s f_t(\w_{t,I}) 
    \leq  \sum_{t=r}^s \langle\nabla f_t(\w_t),\w_t-\w_{t,I}\rangle  \\
    \overset{\eqref{eqn:TUMA:meta-regret:1}}{\leq} & 2GD\left(\frac{\Gamma_I}{\sqrt{\gamma_I}}+2\Gamma_I \right) + \sqrt{\frac{\Gamma_I^2}{\gamma_I}\sum_{t=r}^s \langle\nabla f_t(\w_t),\w_t-\w_{t,I}\rangle^2}  \\
    \leq& 2GD\left(\frac{\Gamma_I}{\sqrt{\gamma_I}}+2\Gamma_I \right) + GD \sqrt{\frac{\Gamma_I^2}{\gamma_I}(s-r+1)} \\
    \overset{\eqref{eqn:UMA2-opt1:gamma-bound}}{\leq} & 2GDc(s)+GD\sqrt{c(s)(s-r+1)}.
    \end{aligned}
\end{equation*}
Combining with \eqref{eqn:Maler:convex}, we attain
\begin{equation*}
    \sum_{t=r}^s f_t(\w_t) - \sum_{t=r}^s f_t(\w) \leq 2GDc(s) + GD \left(\sqrt{c(s)} +2\ln 3 + \frac{3}{2} \right)\sqrt{s-r+1}. 
\end{equation*}
Next, we follow the analysis of Theorem~\ref{thm:UMA2-op1} to extend the above bound to any interval $[p,q]\subseteq [T]$. By applying Lemma~\ref{lem:GC:intervals}, we have
\begin{equation*}
        \sum_{t=p}^q f_t(\w_t) - \sum_{t=p}^q  f_t(\w) 
        \overset{\eqref{eqn:UMA:inequality:5}}{\leq}  2GD c(q)b(p,q) + GD \left(\sqrt{c(q)} +2\ln 3 + \frac{3}{2} \right)\sqrt{q-p+1}.  
\end{equation*}

\subsection{Proof of Theorem~\ref{thm:UMS-Comp}}
Let $\u^\eta_{t}$ be the output of the expert $E^{\eta}$ in the $t$-th round. According to the theoretical guarantee of Optimistic-Adapt-ML-Prod~\citep[Theorem~10]{zhang2024universal}, we have
\begin{equation}\label{eqn:UMS-Comp:meta-regret:0}
\begin{aligned}
&\sum_{t=1}^T \langle \nabla f_t(\u_t), \u_t - \u^\eta_{t} \rangle + \sum_{t=1}^T r(\u_t) - \sum_{t=1}^T r(\u_t^\eta) \\
\leq {} & GD\left(\Xi + \frac{\Psi}{\sqrt{\ln \vert\EC\vert}}\right)+ \frac{\Psi}{\sqrt{\ln \vert\EC\vert}} \sqrt{\sum_{t=1}^T \langle \nabla f_t(\u_t), \u_t - \u^\eta_{t} \rangle^2}
\end{aligned}
\end{equation}
where $\Xi$ and $\Psi$ are defined as
\begin{equation}
    \begin{aligned}
        \Psi &= \ln \vert\EC\vert + \ln \left( 1+\frac{\vert\EC\vert}{e}(1+\ln (T+1)) \right), \\
        \Xi &= \frac{1}{4} \Psi +2\sqrt{\ln \vert\EC\vert} +16\ln \vert \EC\vert.  
    \end{aligned}
\end{equation}
Then, we apply AM-GM inequality and rearrange the second-order term to attain
\begin{equation}\label{eqn:UMS-Comp:meta-regret}
\begin{aligned}
&\sum_{t=1}^T \langle \nabla f_t(\u_t), \u_t - \u^\eta_{t} \rangle - \eta \sum_{t=1}^T \langle \nabla f_t(\u_t), \u_t - \u^\eta_{t} \rangle^2+ \sum_{t=1}^T r(\u_t) - \sum_{t=1}^T r(\u_t^\eta) \\
\leq {} & GD\left(\Xi + \frac{\Psi}{\sqrt{\ln \vert\EC\vert}}\right)+ \frac{\Psi^2}{4\eta\ln \vert\EC\vert}. 
\end{aligned}
\end{equation}
Recall that we employ ProxONS to minimize $\ell_t^\eta (\u)$ in \eqref{eqn:comp:surr:exp}. The time-varying function in $\ell_t^\eta (\u)$ is $1$-exp-concave, and its gradient is bounded by
\begin{equation*}
    \Vert\nabla \ell_t^\eta (\u)\Vert^2=\Vert \eta \nabla f_t(\u_t) + \eta^2 \langle \nabla f_t(\u_t),\u-\u_t\rangle \nabla f_t(\u_t) \Vert^2 \leq (1+\eta GD)^2 \eta^2 G^2 \leq \frac{4}{25 D^2}.
\end{equation*}
According the theoretical guarantee of ProxONS~\citep[Theorem~1]{DBLP:journals/ijon/YangTCWS24}, we have
\begin{equation*}
    \sum_{t=1}^T \ell_t^\eta (\u_t) - \sum_{t=1}^T \ell_t^\eta (\u_{t}^\eta) \leq 4+ 4 d\ln \left(T +1\right)
\end{equation*}
which implies that
\begin{equation}\label{eqn:UMS-Comp:expert-regret:exp}
\begin{aligned}
 &- \sum_{t=1}^T \langle \nabla f_t(\u_t),\u_t-\u_{t}^\eta\rangle + \eta \sum_{t=1}^T\langle \nabla f_t(\u_t),\u_t-\u_{t}^\eta\rangle^2 +  \sum_{t=1}^T r(\u_{t}^\eta) \\
 \leq {} & - \sum_{t=1}^T\langle \nabla f_t(\u_t),\u_t-\u\rangle + \eta \sum_{t=1}^T\langle \nabla f_t(\u_t),\u_t-\u\rangle^2 + \sum_{t=1}^T r(\u) 
  + \frac{4+ 4 d\ln \left(T +1\right)}{\eta}. 
\end{aligned}
\end{equation}
Combining \eqref{eqn:UMS-Comp:meta-regret} with \eqref{eqn:UMS-Comp:expert-regret:exp}, we have
\begin{equation}\label{eqn:UMS-Comp:second-order:eta}
\begin{aligned}
    &\sum_{t=1}^T\langle \nabla f_t(\u_t),\u_t-\u\rangle +\sum_{t=1}^T r(\u_t) - \sum_{t=1}^T r(\u) \\
    \leq {} &   \eta \sum_{t=1}^T\langle \nabla f_t(\u_t),\u_t-\u\rangle^2 + \frac{\psi(T)}{\eta}+GD\left(\Xi + \frac{\Psi}{\sqrt{\ln \vert\EC\vert}}\right)
\end{aligned}
\end{equation}
where $\psi(T)= \Psi^2/(4\ln \vert\EC\vert)+4+ 4 d\ln \left(T +1\right)$. 

Note that the optimal $\eta^*$ minimizes the R.H.S of \eqref{eqn:UMS-Comp:second-order:eta} is 
\begin{equation*}
    \eta^* = \sqrt{\frac{\psi (T)}{\sum_{t=1}^T\langle \nabla f_t(\u_t),\u_t-\u\rangle^2}} \geq \frac{1}{GD\sqrt{T}}. 
\end{equation*}
According to the construction of $\S(T)$ in \eqref{eqn:ST}, if $\eta_* \leq \frac{1}{5DG}$, there must exist an $\eta \in \S(T)$ such that
\[
\eta \leq \eta_* \leq 2\eta.
\]
Then, \eqref{eqn:UMS-Comp:second-order:eta} implies that 
\begin{equation}\label{eqn:UMS-Comp:second-order:1}
\begin{aligned}
&\sum_{t=1}^T \langle \nabla f_t(\u_t),\u_t-\u\rangle +\sum_{t=1}^T r(\u_t) - \sum_{t=1}^T r(\u) \\
    \leq {} &  \eta_* \sum_{t=1}^T \langle \nabla f_t(\u_t),\u_t-\u\rangle^2 + \frac{\psi(T)}{2\eta_*}+GD\left(\Xi + \frac{\Psi}{\sqrt{\ln \vert\EC\vert}}\right)\\
    \leq {} & \frac{3}{2} \sqrt{\psi(T)\sum_{t=1}^T\langle \nabla f_t(\u_t),\u_t-\u\rangle^2}+GD\left(\Xi + \frac{\Psi}{\sqrt{\ln \vert\EC\vert}}\right). 
\end{aligned}
\end{equation}
On the other hand, if $\eta_* \geq \frac{1}{5DG}$, we have
\begin{equation*}
    \sum_{t=1}^T\langle \nabla f_t(\u_t),\u_t-\u\rangle^2 \leq 25 G^2D^2 \psi (T). 
\end{equation*}
Then, \eqref{eqn:UMS-Comp:second-order:eta} with $\eta=\frac{1}{5GD}$ implies that 
\begin{equation}\label{eqn:UMS-Comp:second-order:2}
    \sum_{t=1}^T\langle \nabla f_t(\u_t),\u_t-\u\rangle +\sum_{t=1}^T r(\u_t) - \sum_{t=1}^T r(\u) \leq 10GD \psi (T) + GD\left(\Xi + \frac{\Psi}{\sqrt{\ln \vert\EC\vert}}\right). 
\end{equation}
Combining \eqref{eqn:UMS-Comp:second-order:1} and \eqref{eqn:UMS-Comp:second-order:2}, we obtain a second-order bound 
\begin{equation}\label{eqn:UMS-Comp:second-order}
\begin{aligned}
    &\sum_{t=1}^T\langle \nabla f_t(\u_t),\u_t-\u\rangle +\sum_{t=1}^T r(\u_t) - \sum_{t=1}^T r(\u) \\
    \leq {} & \frac{3}{2} \sqrt{\psi(T)\sum_{t=1}^T\langle \nabla f_t(\u_t),\u_t-\u\rangle^2}+10GD \psi (T)+2GD\left(\Xi + \frac{\Psi}{\sqrt{\ln \vert\EC\vert}}\right). 
\end{aligned}
\end{equation}
When the time-varying function $f_t(\cdot)$ is $\alpha$-exp-concave, we have
\begin{equation*}
    \begin{aligned}
        & \sum_{t=1}^T \left(f_t(\u_t)+r(\u_t)\right) - \sum_{t=1}^T \left(f_t(\u)+r(\u)\right) \\
        \overset{\eqref{eqn:lem:exp}}{\leq} {} & \sum_{t=1}^T\langle \nabla f_t(\u_t),\u_t-\u\rangle +\sum_{t=1}^T r(\u_t) - \sum_{t=1}^T r(\u) - \frac{\beta}{2} \sum_{t=1}^T\langle \nabla f_t(\u_t),\u_t-\u\rangle^2 \\ 
        \overset{\eqref{eqn:UMS-Comp:second-order}}{\leq} {} & \frac{3}{2} \sqrt{\psi(T)\sum_{t=1}^T\langle \nabla f_t(\u_t),\u_t-\u\rangle^2}+10GD \psi (T)+2GD\left(\Xi + \frac{\Psi}{\sqrt{\ln \vert\EC\vert}}\right) \\
        &- \frac{\beta}{2} \sum_{t=1}^T\langle \nabla f_t(\u_t),\u_t-\u\rangle^2 \\
        \leq {} &  \left(\frac{9}{8\beta} + 10 GD \right) \psi (T)+2GD\left(\Xi + \frac{\Psi}{\sqrt{\ln \vert\EC\vert}}\right)
    \end{aligned}
\end{equation*}
where the last inequality is due to $\sqrt{ab}\leq \frac{a}{2}+\frac{b}{2}$. Due to our construction of experts, the number of experts is $\vert\EC\vert = 3+\lceil \log_2 T\rceil$. Finally, plugging $\vert \EC\vert$ into the above bound yields the desired result. 

Next, we focus on strongly convex functions. Let $\widehat{\u}^\eta_{t}$ be the output of the expert $\widehat{E}^{\eta}$ in the $t$-th round. For the expert $\widehat{E}^{\eta}$, \eqref{eqn:UMS-Comp:meta-regret} also holds, and due to Assumption~\ref{ass:2}, we obtain
\begin{equation}\label{eqn:UMS-Comp:meta-regret:2}
\begin{aligned}
& \sum_{t=1}^T \langle \nabla f_t(\u_t), \u_t - \widehat{\u}^\eta_{t} \rangle - \eta G^2 \sum_{t=1}^T \Vert \u_t - \widehat{\u}^\eta_{t} \Vert^2+  \sum_{t=1}^T r(\u_t) -  \sum_{t=1}^T r(\widehat{\u}_t^\eta) \\
\leq {} &  GD\left(\Xi + \frac{\Psi}{\sqrt{\ln \vert\EC\vert}}\right)+ \frac{\Psi^2}{4\eta \ln \vert\EC\vert}. 
\end{aligned}
\end{equation}
Recall that we use FOBOS to minimize  $\hat{\ell}_t^\eta (\u)$ in  \eqref{eqn:comp:surr:str}. 
We know  that the time-varying function in $\hat{\ell}_t^\eta (\u)$ is $2\eta^2 G^2$-strongly convex, and its gradient is bounded by
\begin{equation*}
    \Vert\nabla \hat{\ell}_t^\eta (\u)\Vert^2=\Vert \eta \nabla f_t(\u_t) + 2\eta^2 G^2 (\u-\u_t) \Vert^2 \leq G^2 \eta^2 (1+2\eta GD)^2\leq 4\eta^2 G^2. 
\end{equation*}
According to the theoretical guarantee of FOBOS~\citep[Theorem~8]{DBLP:journals/jmlr/DuchiS09}, we have
\begin{equation*}
    \sum_{t=1}^T \hat{\ell}_t^\eta (\u_t) - \sum_{t=1}^T \hat{\ell}_t^\eta (\widehat{\u}_t^\eta) \leq 8+7\log T.
\end{equation*}
which implies that
\begin{equation}\label{eqn:UMS-Comp:expert-regret:str}
\begin{aligned}
 &- \sum_{t=1}^T \langle \nabla f_t(\u_t),\u_t-\widehat{\u}_{t}^\eta\rangle + \eta G^2 \sum_{t=1}^T\Vert\u_t-\widehat{\u}_{t}^\eta\Vert^2 +  \sum_{t=1}^T r(\widehat{\u}_{t}^\eta) \\
 \leq {} & - \sum_{t=1}^T\langle \nabla f_t(\u_t),\u_t-\u\rangle + \eta G^2 \sum_{t=1}^T\Vert \u_t-\u\Vert^2 + \sum_{t=1}^T r(\u) 
  + \frac{8+7\log T}{\eta}. 
\end{aligned}    
\end{equation}
Combining \eqref{eqn:UMS-Comp:expert-regret:str} with \eqref{eqn:UMS-Comp:meta-regret:2}, we arrive at
\begin{equation}\label{eqn:UMS-Comp:second-order:eta:2}
\begin{aligned}
    & \sum_{t=1}^T \langle \nabla f_t(\u_t),\u_t-\u\rangle+ \sum_{t=1}^T r(\u_t) -  \sum_{t=1}^T r(\u)  \\
    \leq {} & \eta G^2 \sum_{t=1}^T \Vert \u_t-\u\Vert^2 + \frac{\hat{\psi} (T)}{\eta} +  GD\left(\Xi + \frac{\Psi}{\sqrt{\ln \vert\EC\vert}}\right)
\end{aligned}
\end{equation}
where $\hat{\psi} (T) = \frac{\Psi^2}{4\ln \vert\EC\vert}+8+7\log T$. The optimal learning  rate of the right side in \eqref{eqn:UMS-Comp:second-order:eta:2} is 
\begin{equation*}
    \eta_* = \sqrt{\frac{\hat{\psi} (T)}{G^2 \sum_{t=1}^T \Vert \u_t-\u\Vert^2 }} \geq \frac{\sqrt{2}}{GD\sqrt{T}}. 
\end{equation*}
If $\eta_* \leq \frac{1}{5GD}$, then \eqref{eqn:UMS-Comp:second-order:eta:2} implies that
\begin{equation}\label{eqn:UMS-Comp:second-order:2:1}
    \begin{aligned}
        & \sum_{t=1}^T\langle \nabla f_t(\u_t),\u_t-\u\rangle+ \sum_{t=1}^T r(\u_t) -  \sum_{t=1}^T r(\u)  \\
    \leq {} & \eta_* G^2 \sum_{t=1}^T\Vert \u_t-\u\Vert^2 + \frac{2\hat{\psi} (T)}{\eta_*} +  GD\left(\Xi + \frac{\Psi}{\sqrt{\ln \vert\EC\vert}}\right) \\
    \leq {} & 3G\sqrt{\hat{\psi} (T)\sum_{t=1}^T \Vert \u_t-\u\Vert^2 } + GD\left(\Xi + \frac{\Psi}{\sqrt{\ln \vert\EC\vert}}\right). 
    \end{aligned}
\end{equation}
On the other hand, if $\eta_*\geq \frac{1}{5GD}$, we have $\sum_{t=1}^T \Vert \u_t-\u\Vert^2 \leq 25D^2 \hat{\psi} (T)$. 
Then, \eqref{eqn:UMS-Comp:second-order:eta:2} with $\eta_* = \frac{1}{5GD}$ implies
\begin{equation}\label{eqn:UMS-Comp:second-order:2:2}
         \sum_{t=1}^T\langle \nabla f_t(\u_t),\u_t-\u\rangle+ \sum_{t=1}^T r(\u_t) -  \sum_{t=1}^T r(\u) \leq 10GD \hat{\psi} (T) + GD\left(\Xi + \frac{\Psi}{\sqrt{\ln \vert\EC\vert}}\right).
\end{equation}
Combining \eqref{eqn:UMS-Comp:second-order:2:1} and \eqref{eqn:UMS-Comp:second-order:2:2}, we have
\begin{equation}\label{eqn:UMS-Comp:second-order:2:final}
    \begin{aligned}
       & \sum_{t=1}^T\langle \nabla f_t(\u_t),\u_t-\u\rangle+ \sum_{t=1}^T r(\u_t) -  \sum_{t=1}^T r(\u) \\
        \leq {} & 3G\sqrt{\hat{\psi} (T)\sum_{t=1}^T \Vert \u_t-\u\Vert^2 } + 10GD \hat{\psi} (T)+ 2GD\left(\Xi + \frac{\Psi}{\sqrt{\ln \vert\EC\vert}}\right). 
    \end{aligned}
\end{equation}
When the time-varying function $f_t(\cdot)$ is $\lambda$-strongly convex, we have
\begin{equation*}
    \begin{aligned}
        & \sum_{t=1}^T \left(f_t(\u_t)+r(\u_t)\right) - \sum_{t=1}^T \left(f_t(\u)+r(\u)\right) \\
        \overset{\eqref{eqn:def:str}}{\leq} {} & \sum_{t=1}^T\langle \nabla f_t(\u_t),\u_t-\u\rangle +\sum_{t=1}^T r(\u_t) - \sum_{t=1}^T r(\u) - \frac{\lambda}{2} \sum_{t=1}^T\Vert  \u_t-\u\Vert^2 \\ 
        \overset{\eqref{eqn:UMS-Comp:second-order:2:final}}{\leq} {} & 3G\sqrt{\hat{\psi} (T)\sum_{t=1}^T \Vert \u_t-\u\Vert^2 } + 10GD \hat{\psi} (T)+ 2GD\left(\Xi + \frac{\Psi}{\sqrt{\ln \vert\EC\vert}}\right) - \frac{\lambda}{2} \sum_{t=1}^T\Vert  \u_t-\u\Vert^2 \\
        \leq {} & \left( \frac{9G^2}{\lambda} + 10GD \right) \hat{\psi} (T) + 2GD\left(\Xi + \frac{\Psi}{\sqrt{\ln \vert\EC\vert}}\right) 
    \end{aligned}
\end{equation*}
where the last step is due to $\sqrt{ab}\leq \frac{a}{2}+\frac{b}{2}$.  

Finally, we focus on general convex functions. Let $\widetilde{\u}_{t}$ be the output of the expert $\widetilde{E}$ in the $t$-th round. According to Assumptions~\ref{ass:1} and \ref{ass:2}, \eqref{eqn:UMS-Comp:meta-regret:0} implies
\begin{equation}\label{eqn:UMS-Comp:meta-regret:cvx}
\begin{aligned}
&\sum_{t=1}^T f_t(\u_t) - \sum_{t=1}^T f_t(\widetilde{\u}_{t}) + \sum_{t=1}^T r(\u_t) - \sum_{t=1}^T r(\u) \\
\leq {} & \sum_{t=1}^T \langle \nabla f_t(\u_t), \u_t - \widetilde{\u}_{t} \rangle + \sum_{t=1}^T r(\u_t) - \sum_{t=1}^T r(\widetilde{\u}_t) \\
\leq {} &   GD\left(\Xi + \frac{\Psi}{\sqrt{\ln \vert\EC\vert}}\right)+ \frac{GD\Psi}{\sqrt{\ln \vert\EC\vert}} \sqrt{T}. 
\end{aligned}
\end{equation}
where the first inequality is due to the convexity. According to the theoretical guarantee of FOBOS~\citep[Theorem~6]{DBLP:journals/jmlr/DuchiS09}, we have 
\begin{equation}\label{eqn:UMS-Comp:expert-regret:cvx}
    \sum_{t=1}^T f_t(\widetilde{\u}_t) - \sum_{t=1}^T f_t(\u) + \sum_{t=1}^T r(\widetilde{\u}_t) - \sum_{t=1}^T r(\u) \leq GD+GD\sqrt{7T}
\end{equation}
where we set $\eta=\frac{D}{G\sqrt{7T}}$. Combining \eqref{eqn:UMS-Comp:meta-regret:cvx} and \eqref{eqn:UMS-Comp:expert-regret:cvx}, we finish the proof. 

\subsection{Proof of Theorem~\ref{thm:UMA-Comp}}
Let $\w_{t,I}$, $\ell_{t,I}$ and $m_{t,I}$ be the output, normalized loss and optimism of the expert $E_I$ in the $t$-th round. We start with the meta-regret in terms of $\ell_{t,I}$. 
\begin{lemma}\label{lem:TUMA-Comp:meta-regret}
    Under Assumptions~\ref{ass:1}, \ref{ass:2}, \ref{ass:3} and \ref{ass:4}, for any interval $I=[r,s]\in \I$, the meta-regret of UMA-Comp satisfies
    \begin{equation*}
        \sum_{t=r}^s \ell_{t} - \sum_{t=r}^s \ell_{t,I} \leq \frac{\Gamma_I}{\sqrt{\gamma_I}} \sqrt{1+\sum_{t=r}^s (\ell_t-\ell_{t,I}-m_{t,I})^2} + 2\Gamma_I
    \end{equation*}
    where $\Gamma_I = 2\gamma_I+\ln N_s + \ln \ln \left(9+36s\right)$ and $N_s$ is the number of experts created till round $s$. 
\end{lemma}
According to the definition of $\ell_t$, $\ell_{t,I}$, and $m_{t,I}$, we have 
\begin{equation*}
    \begin{aligned}
    & \sum_{t=r}^s \langle \nabla f_t(\w_t), \w_t-\w_{t,I}\rangle + \sum_{t=r}^s \sum_{E_J\in\A_t} p_{t,J} r(\w_{t,J}) - \sum_{t=r}^s r(\w_{t,I}) \\
    \leq & \frac{\Gamma_I}{\sqrt{\gamma_I}} \sqrt{G^2D^2+\sum_{t=r}^s \langle \nabla f_t(\w_t), \w_t-\w_{t,I}\rangle^2} + 2GD\Gamma_I. 
    \end{aligned}
\end{equation*}
According to Jensen’s inequality that $r(\w_t) = r(\sum_{E_J\in\A_t} p_{t,J}\w_{t,J})\leq \sum_{E_J\in\A_t} p_{t,J} r(\w_{t,J})$, the above inequality can be rewritten as
\begin{equation}\label{eqn:TUMA-Comp:meta-regret}
\begin{aligned}
    &\sum_{t=r}^s \langle \nabla f_t(\w_t), \w_t-\w_{t,I}\rangle + \sum_{t=r}^s r(\w_t) - \sum_{t=r}^s r(\w_{t,I}) \\
    \leq &  \frac{\Gamma_I}{\sqrt{\gamma_I}} \sqrt{G^2D^2+\sum_{t=r}^s \langle \nabla f_t(\w_t), \w_t-\w_{t,I}\rangle^2} + 2GD\Gamma_I \\
    \leq & GD \left( \frac{\Gamma_I}{\sqrt{\gamma_I}} + 2\Gamma_I\right) + \sqrt{\frac{\Gamma^2_I}{\gamma_I}\sum_{t=r}^s \langle \nabla f_t(\w_t), \w_t-\w_{t,I}\rangle^2}
\end{aligned}
\end{equation}
where the last step is due to $\sqrt{a+b}\leq \sqrt{a}+\sqrt{b}$. Recall that we utilize UMS-Comp to minimize $F_t(\cdot)=f_t(\cdot)+r(\cdot)$ during each interval $I=[r,s]\in\I$. Therefore, we can directly use the theoretical guarantee of UMS-Comp to bound the expert-regret. 

When functions are $\alpha$-exp-concave during the interval $[r,s]$, the meta-regret in terms of the composite function is bounded by
\begin{equation}\label{eqn:UMA-Comp:meta:exp}
    \begin{aligned}
        & \sum_{t=r}^s F_t(\w_t) - \sum_{t=r}^s F_t(\w_{t,I}) \\
        \overset{\eqref{eqn:lem:exp}}{\leq} {} & \sum_{t=r}^s \langle \nabla f_t(\w_t),\w_t-\w_{t,I}\rangle - \frac{\beta}{2} \sum_{t=r}^s \langle \nabla f_t(\w_t),\w_t-\w_{t,I}\rangle^2+ \sum_{t=r}^s \left(r(\w_t)-r(\w_{t,I})\right) \\
        \overset{\eqref{eqn:TUMA-Comp:meta-regret}}{\leq} {} & GD \left( \frac{\Gamma_I}{\sqrt{\gamma_I}} + 2\Gamma_I\right) + \sqrt{\frac{\Gamma^2_I}{\gamma_I}\sum_{t=r}^s \langle \nabla f_t(\w_t), \w_t-\w_{t,I}\rangle^2} - \frac{\beta}{2} \sum_{t=r}^s \langle \nabla f_t(\w_t),\w_t-\w_{t,I}\rangle^2 \\
        \leq {} & GD \left( \frac{\Gamma_I}{\sqrt{\gamma_I}} + 2\Gamma_I\right) + \frac{\Gamma_I^2}{2\beta\gamma_I} \\
        \overset{\eqref{eqn:UMA2-opt1:gamma-bound}}{\leq}  {} & \left(GD+\frac{1}{2\beta}\right) c(s)
    \end{aligned}
\end{equation}
where $c(\cdot)$ is defined in \eqref{eqn:c}. According to Theorem~\ref{thm:UMS-Comp}, we have 
\begin{equation*}
    \sum_{t=r}^s F_t(\w_{t,I}) - \sum_{t=r}^s F_t(\w) \leq \left( \frac{9}{8\beta }+10GD \right) \cdot \left(4d\ln (s-r+2)+\phi_1+4\right) + 2GD \phi_2 = \varphi (r,s). 
\end{equation*}
Combining the above bound with \eqref{eqn:UMA-Comp:meta:exp}, we have
\begin{equation*}
    \sum_{t=r}^s F_t(\w_t) -  \sum_{t=r}^s F_t(\w) \leq \left(GD+\frac{1}{2\beta}\right) c(s)+\varphi(r,s). 
\end{equation*}
Finally, we extend it to any interval $[p,q]\subseteq [T]$, and the analysis is similar to that of Theorem~\ref{thm:TUMA}. Based on Lemma~\ref{lem:GC:intervals}, we have
\begin{equation*}
         \sum_{t=p}^q F_t(\w_t) -  \sum_{t=p}^q F_t(\w)
        \leq  \left(GD+\frac{1}{2\beta}\right) c(q)b(p,q)+ \varphi (p,q) b(p,q).
\end{equation*}

When functions are $\lambda$-strongly convex, the meta-regret in terms of the composite function can be bounded by
\begin{equation*}
    \begin{aligned}
        & \sum_{t=r}^s F_t(\w_t) -\sum_{t=r}^s F_t(\w_{t,I})  \\
        \overset{\eqref{eqn:def:str}}{\leq} {} & \sum_{t=r}^s \langle \nabla f_t(\w_t),\w_t-\w_{t,I}\rangle - \frac{\lambda}{2} \sum_{t=r}^s \Vert\w_t-\w_{t,I}\Vert^2+ \sum_{t=r}^s \left(r(\w_t)-r(\w_{t,I})\right) \\
        \overset{\eqref{eqn:TUMA-Comp:meta-regret}}{\leq} {} & GD \left( \frac{\Gamma_I}{\sqrt{\gamma_I}} + 2\Gamma_I\right) + \sqrt{\frac{\Gamma^2_I}{\gamma_I}\sum_{t=r}^s \langle \nabla f_t(\w_t), \w_t-\w_{t,I}\rangle^2} - \frac{\lambda}{2} \sum_{t=r}^s \Vert\w_t-\w_{t,I}\Vert^2 \\
        \leq {} & GD \left( \frac{\Gamma_I}{\sqrt{\gamma_I}} + 2\Gamma_I\right) + \frac{\Gamma_I^2G^2}{2\lambda\gamma_I} \\
        \overset{\eqref{eqn:UMA2-opt1:gamma-bound}}{\leq} {} & \left(GD+\frac{G^2}{2\lambda}\right) c(s). 
    \end{aligned}
\end{equation*}
Next, we combine the above bound with the expert-regret in Theorem~\ref{thm:UMS-Comp} to attain
\begin{equation*}
    \sum_{t=r}^s F_t(\w_t) -  \sum_{t=r}^s F_t(\w)\leq \left(GD+\frac{G^2}{2\lambda}\right) c(s) + \hat{\varphi} (r,s)
\end{equation*}
where 
\begin{equation*}
    \hat{\varphi} (r,s) = \left( \frac{9G^2}{\lambda }+10GD \right) \cdot \left(7\log (s-r+1)+8+\phi_1 \right) + 2GD \phi_2. 
\end{equation*}
Then, we  extend it to any interval $[p,q]\subseteq [T]$. Based on Lemma~\ref{lem:GC:intervals}, we have
\begin{equation*}
         \sum_{t=p}^q F_t(\w_t) -\sum_{t=p}^q F_t(\w)  
        \leq  \left(GD+\frac{G^2}{2\lambda}\right) c(q)b(p,q)+\hat{\varphi}(p,q) b(p,q) 
        =  O\left(\frac{\log q\log (q-p)}{\lambda}\right). 
\end{equation*}

Finally, we focus on general convex functions. When functions are convex, we have
\begin{equation*}
    \begin{aligned}
        & \sum_{t=r}^s F_t(\w_t) - \sum_{t=r}^s F_t(\w_{t,I}) 
        \leq  \sum_{t=r}^s \langle \nabla f_t(\w_t),\w_t-\w_{t,I}\rangle + \sum_{t=r}^s \left(r(\w_t)-r(\w)\right) \\
        \overset{\eqref{eqn:TUMA-Comp:meta-regret}}{\leq} & GD \left( \frac{\Gamma_I}{\sqrt{\gamma_I}} + 2\Gamma_I\right) + \sqrt{\frac{\Gamma^2_I}{\gamma_I}\sum_{t=r}^s \langle \nabla f_t(\w_t), \w_t-\w_{t,I}\rangle^2}  \\
        \leq & GD \left( \frac{\Gamma_I}{\sqrt{\gamma_I}} + 2\Gamma_I\right) + GD\sqrt{\frac{\Gamma^2_I}{\gamma_I} (s-r+1)} \\
        \overset{\eqref{eqn:UMA2-opt1:gamma-bound}}{\leq} & GDc(s)+GD\sqrt{c(s)(s-r+1)}. 
    \end{aligned}
\end{equation*}
Then, we combine the meta-regret and the expert-regret in Theorem~\ref{thm:UMS-Comp} to arrive at
\begin{equation*}
\begin{aligned}
    \sum_{t=r}^s F_t(\w_t) - \sum_{t=r}^s F_t(\w) 
    \leq GDc(s)+GD \left( \sqrt{c(s)} +\phi_3\right) \sqrt{s-r+1}+ GD(\phi_2+1) 
\end{aligned}
\end{equation*}
Finally, we extend it to any interval $[p,q]\subseteq [T]$. Based on Lemma~\ref{lem:GC:intervals}, we have
\begin{equation*}
\begin{aligned}
     \sum_{t=p}^q F_t(\w_t) - \sum_{t=p}^q F_t(\w) 
        &\overset{\eqref{eqn:UMA:inequality:5}}{\leq}   GD c(q)b(p,q) + GD \left(\sqrt{c(q)}+\phi_3\right) \sqrt{q-p+1} +GD(\phi_2+1) \\
        &= O(\sqrt{(q-p)\log q}). 
\end{aligned}
\end{equation*}

\subsection{Proof of Lemma~\ref{lem:TUMA-Comp:meta-regret}}
We extend Lemma~E.1 of \citet{NIPS:2016:Wei} to support sleeping experts, achieving the following lemma. 
\begin{lemma}\label{lem:TUMA-Comp:adaptive-meta}
    Under Assumptions~\ref{ass:1}, \ref{ass:2}, \ref{ass:3} and \ref{ass:4}, for any interval $I=[r,s]\in\I$, the meta-regret of UMA-Comp satisfies
    \begin{equation*}
    \begin{aligned}
        \sum_{t=r}^s \ell_{t} - \sum_{t=r}^s \ell_{t,I} &\leq \frac{1}{\up_{r-1,I}} \ln \frac{1}{x_{r-1,I}} + \sum_{t=r}^s \up_{t-1,I} (\ell_{t}-\ell_{t,I}-m_{t,I})^2 \\
        + &\frac{1}{\up_{s,I}}\ln \left(N_s+ \frac{1}{e} \sum_{t=1}^s\sum_{E_J\in \A_t} \left(\frac{\up_{t-1,J}}{\up_{t,J}}-1\right)\right)
    \end{aligned}
    \end{equation*}
    where $N_s$ denotes the number of experts created till round $s$. 
\end{lemma}
The following analysis is similar to \citet[
Corollary~4]{pmlr-v35-gaillard14}. According to the definition of $\up_{t-1,I}$, we have
\begin{equation}\label{eqn:TUMA-Comp:term1-}
    \sum_{t=r}^s \up_{t-1,I} (r_{t,I}-m_{t,I})^2 \leq \sqrt{\gamma_I} \sum_{t=r}^s \frac{(r_{t,I}-m_{t,I})^2}{\sqrt{1+L_{t-1,I}}}
\end{equation}
where $r_{t,I}=\ell_t-\ell_{t,I}$. Then, we introduce the following lemma \citep[Lemma~14]{pmlr-v35-gaillard14} to bound the above term. 
\begin{lemma}\label{lem:fa}
    Let $a_0>0$ and $a_1,\cdots,a_m\in [0,1]$ be real numbers and let $f\colon (0,+\infty)\rightarrow [0,+\infty)$ be a non-increasing function. Then
    \begin{equation*}
        \sum_{i=1}^m a_i f(a_0+\cdots+a_{i-1}) \leq f(a_0) + \int_{a_0}^{a_0+a_1+\cdots+a_m} f(u) du.
    \end{equation*}
\end{lemma}
By applying Lemma~\ref{lem:fa} with $f(x)=\frac{1}{\sqrt{x}}$, we have
\begin{equation*}
\begin{aligned}
    \sum_{t=r}^s \frac{(r_{t,I}-m_{t,I})^2}{\sqrt{1+L_{t-1,I}}} &\leq \frac{1}{\sqrt{1+L_{r-1,I}}}+ \int_{L_{r-1,I}}^{L_{s,I}} \frac{1}{\sqrt{1+u}} du \\
    &\leq 1-2\sqrt{1} + 2\sqrt{1+\sum_{t=r}^s (r_{t,I}-m_{t,I})^2}. 
\end{aligned}
\end{equation*}
By substituting the above term into \eqref{eqn:TUMA-Comp:term1-}, we have
\begin{equation}\label{eqn:TUMA-Comp:term1}
    \sum_{t=r}^s \up_{t-1,I} (r_{t,I}-m_{t,I})^2 \leq 2\sqrt{\gamma_I} \sqrt{1+\sum_{t=r}^s (r_{t,I}-m_{t,I})^2}. 
\end{equation}

Next, we proceed to bound the following term
\begin{equation}\label{eqn:TUMA-Comp:term2-}
    \begin{aligned}
        \sum_{t=1}^s\sum_{E_J\in \A_t} \left(\frac{\up_{t-1,J}}{\up_{t,J}}-1\right) &\leq  \sum_{t=1}^s\sum_{E_J\in \A_t} \left(\sqrt{\frac{1+L_{t,J}}{1+L_{t-1,J}}}-1\right) \\
        &\leq \sum_{t=1}^s\sum_{E_J\in \A_t} \left(\sqrt{1+\frac{(r_{t,J}-m_{t,J})^2}{1+L_{t-1,J}}}-1\right) \\
        &\leq \frac{1}{2} \sum_{t=1}^s\sum_{E_J\in \A_t} \frac{(r_{t,J}-m_{t,J})^2}{1+L_{t-1,J}}
    \end{aligned}
\end{equation}
where the last inequality is due to $g(1+z)\leq g(1)+zg^\prime (1),z\geq 0$ for any concave function~$g(\cdot)$. Denote $e_J$ be the ending time of the expert $E_J$. We can rewrite \eqref{eqn:TUMA-Comp:term2-} to arrive at
\begin{equation*}
    \begin{aligned}
        \frac{1}{2} \sum_{t=1}^s\sum_{E_J\in \A_t} \frac{(r_{t,J}-m_{t,J})^2}{1+L_{t-1,J}} &= \frac{1}{2} \sum_{E_J\in \cup_{i=1}^s \A_i} \sum_{t=\min J}^{s\wedge e_J} \frac{(r_{t,J}-m_{t,J})^2}{1+L_{t-1,J}} \\
        &\leq \frac{1}{2} \sum_{E_J\in \cup_{i=1}^s \A_i} \left(1+\ln \left( 1+\sum_{t=\min J}^{s\wedge e_J} (r_{t,J}-m_{t,J})^2\right)-\ln (1)\right) \\
        &\leq \frac{1}{2} \sum_{E_J\in \cup_{i=1}^s \A_i} \left(1+\ln (1+4s)\right) \\
        &\leq \frac{N}{2} \left(1+\ln (1+4s)\right)
    \end{aligned}
\end{equation*}
where the first inequality is because we apply Lemma~\ref{lem:fa} with $f(x)=\frac{1}{x}$, and the second inequality is due to $\vert r_{t,J}-m_{t,J}\vert\leq 2$. 

Furthermore, we have 
\begin{equation}\label{eqn:TUMA-Comp:term2}
    \ln \left(N+ \frac{1}{e} \sum_{t=1}^s\sum_{E_J\in \A_t} \left(\frac{\up_{t-1,J}}{\up_{t,J}}-1\right)\right) \leq \ln N + \ln \ln \left(9+36s\right)=\G(N,s). 
\end{equation}
Substituting \eqref{eqn:TUMA-Comp:term1} and \eqref{eqn:TUMA-Comp:term2} into Lemma~\ref{lem:TUMA-Comp:adaptive-meta}, we have
\begin{equation}\label{eqn:TUMA-Comp:meta-regret-1}
    \sum_{t=r}^s \ell_{t} - \sum_{t=r}^s \ell_{t,I} \leq \frac{1}{\up_{s,I}} \left(\ln \frac{1}{x_{r-1,I}} + \G(N,s)\right)+ 2\sqrt{\gamma_I \left(1+\sum_{t=r}^s (r_{t,I}-m_{t,I})^2\right)}. 
\end{equation}
Now if $\sqrt{1+\sum_{t=r}^s (r_{t,I}-m_{t,I})^2}> 2\sqrt{\gamma_I}$ then $\up_{s,I}<\frac{1}{2}$, \eqref{eqn:TUMA-Comp:meta-regret-1} is bounded by 
\begin{equation}\label{eqn:TUMA-Comp:meta-regret-2}
    \sum_{t=r}^s \ell_{t} - \sum_{t=r}^s \ell_{t,I} \leq \sqrt{1+\sum_{t=r}^s (r_{t,I}-m_{t,I})^2} \left(2\sqrt{\gamma_I}+ \frac{\ln \frac{1}{x_{r-1,I}} + \G(N,s)}{\sqrt{\gamma_I}}\right).
\end{equation}
Alternatively, if $\sqrt{1+\sum_{t=r}^s (r_{t,I}-m_{t,I})^2}\leq 2\sqrt{\gamma_I}$ then $\up_{s,I}=\frac{1}{2}$, \eqref{eqn:TUMA-Comp:meta-regret-1} is bounded by
\begin{equation}\label{eqn:TUMA-Comp:meta-regret-3}
    \sum_{t=r}^s \ell_{t} - \sum_{t=r}^s \ell_{t,I} \leq 2\ln \frac{1}{x_{r-1,I}} + 2\G(N,s) + 4\gamma_I. 
\end{equation}
Combining \eqref{eqn:TUMA-Comp:meta-regret-2} and \eqref{eqn:TUMA-Comp:meta-regret-3}, we finish the proof. 

\subsection{Proof of Lemma~\ref{lem:TUMA-Comp:adaptive-meta}}
This lemma is an extension of Lemma~E.1 of \citet{NIPS:2016:Wei} to sleeping experts. We first introduce the following inequality. 
\begin{lemma}\label{lem:xalpha}
    For all $x>0$ and $\alpha\geq 1$, we have $x\leq x^\alpha +(\alpha-1)/e$. 
\end{lemma}
We start to analyze the meta-regret over any interval $I=[r,s]\in\I$. Let $X_s = \sum_{E_J\in\A_s} x_{s,J}$ and $r_{t,I} = \ell_t - \ell_{t,I}$, we aim to bound $\ln X_s$ from below and above. 

For the lower bound, according to the definition of $x_{t,J}$, we have
\begin{equation}\label{eqn:TUMA-Comp:meta-regret:lower-bound}
        \ln X_s \geq \ln x_{s,I} = \frac{\up_{s,I}}{\up_{r-1,I}} \ln x_{r-1,I} + \up_{s,I} \sum_{t=r}^s \left( r_{t,I} - \up_{t-1,I} (r_{t,I}-m_{t,I})^2\right). 
\end{equation}

Then, we derive its upper bound
\begin{equation*}
    \begin{aligned}
        (x_{t,J})^{\frac{\up_{t-1,J}}{\up_{t,J}}} &= x_{t-1,J} \exp\left( \up_{t-1,J}r_{t,J}-\up_{t-1,J}^2 (r_{t,J}-m_{t,J})^2\right) \\
        &= \widetilde{x}_{t-1,J} \exp\left( \up_{t-1,J}(r_{t,J}-m_{t,J})-\up_{t-1,J}^2 (r_{t,J}-m_{t,J})^2\right) \\
        &\leq \widetilde{x}_{t-1,J} \left(1+ \up_{t-1,J}(r_{t,J}-m_{t,J})\right)
    \end{aligned}
\end{equation*}
where the equality is due to the definition of $x_{t,J}$ and $\widetilde{x}_{t-1,J}$, and the last inequality is due to $\ln (1+z)\geq z-z^2$ for all $z\geq -1/2$. Next, we sum the above over all the experts $E_J\in\A_t$ to arrive at
\begin{equation*}
\begin{aligned}
    \sum_{E_J\in \A_t} (x_{t,J})^{\frac{\up_{t-1,J}}{\up_{t,J}}} &\leq \sum_{E_J\in \A_t} \widetilde{x}_{t-1,J} \left(1+ \up_{t-1,J}(r_{t,J}-m_{t,J})\right) \\
    & = \sum_{E_J\in \A_t} \widetilde{x}_{t-1,J} + \sum_{E_J\in \A_t} \widetilde{x}_{t-1,J} \up_{t-1,J}r_{t,J} - \sum_{E_J\in \A_t} \widetilde{x}_{t-1,J}\up_{t-1,J}m_{t,J}.
\end{aligned}
\end{equation*}
Next, we proceed to prove that the second term in the above inequality is always equal to $0$, because
\begin{equation*}
        \sum_{E_J\in \A_t} \widetilde{x}_{t-1,J} \up_{t-1,J}r_{t,J} = \left(\sum_{E_J\in \A_t} \widetilde{x}_{t-1,J} \up_{t-1,J}\right)\ell_{t} - \sum_{E_J\in \A_t} \widetilde{x}_{t-1,J} \up_{t-1,J}\ell_{t,J} 
        = 0. 
\end{equation*}
Then, we use the fact that $1-x\leq \exp(-x)$ for any $x$ to obtain
\begin{equation*}
    \sum_{E_J\in \A_t} (x_{t,J})^{\frac{\up_{t-1,J}}{\up_{t,J}}} \leq \sum_{E_J\in \A_t} \widetilde{x}_{t-1,J} \exp \left(-\up_{t-1,J}m_{t,J}\right) = \sum_{E_J\in \A_t} x_{t-1,J}. 
\end{equation*}
Due to $x_{t,J}>0$ and $\frac{\up_{t-1,J}}{\up_{t,J}}\geq 1$, Lemma~\ref{lem:xalpha} implies 
\begin{equation}\label{eqn:TUMA-Comp:sumAtxtJ}
\begin{aligned}
    \sum_{E_J\in \A_t} x_{t,J} &\leq \sum_{E_J\in \A_t} (x_{t,J})^{\frac{\up_{t-1,J}}{\up_{t,J}}} + \frac{1}{e} \sum_{E_J\in \A_t} \left(\frac{\up_{t-1,J}}{\up_{t,J}}-1\right) \\
    &\leq \sum_{E_J\in \A_t} x_{t-1,J}+ \frac{1}{e} \sum_{E_J\in \A_t} \left(\frac{\up_{t-1,J}}{\up_{t,J}}-1\right)
\end{aligned}
\end{equation}
Summing \eqref{eqn:TUMA-Comp:sumAtxtJ} over $t=1,2,\cdots,s$, we have
\begin{equation*}
    \sum_{t=1}^s \sum_{E_J\in \A_t} x_{t,J} \leq \sum_{t=1}^s\sum_{E_J\in \A_t} x_{t-1,J}+ \frac{1}{e} \sum_{t=1}^s\sum_{E_J\in \A_t} \left(\frac{\up_{t-1,J}}{\up_{t,J}}-1\right)
\end{equation*}
which can be rewritten as
\begin{equation*}
    \begin{aligned}
        &\sum_{E_J\in\A_s} x_{s,J} + \sum_{t=1}^{s-1} \left(\sum_{E_J\in \A_t\setminus \A_{t+1}} x_{t,J}+ \sum_{E_J\in \A_t\cap \A_{t+1}}x_{t,J} \right) \\
        \leq & \sum_{E_J\in \A_{1}} x_{0,J} + \sum_{t=2}^s \left(\sum_{E_J\in \A_t\setminus \A_{t-1}} x_{t-1,J}+ \sum_{E_J\in \A_t\cap \A_{t-1}}x_{t-1,J} \right)+ \frac{1}{e} \sum_{t=1}^s\sum_{E_J\in \A_t} \left(\frac{\up_{t-1,J}}{\up_{t,J}}-1\right)
    \end{aligned}
\end{equation*}
implying
\begin{equation*}
    \begin{aligned}
        &\sum_{E_J\in\A_s} x_{s,J} + \sum_{t=1}^{s-1}\sum_{E_J\in \A_t\setminus \A_{t+1}} x_{t,J} \\
        \leq & \sum_{E_J\in \A_{1}} x_{0,J} + \sum_{t=2}^s\sum_{E_J\in \A_t\setminus \A_{t-1}} x_{t-1,J}+ \frac{1}{e} \sum_{t=1}^s\sum_{E_J\in \A_t} \left(\frac{\up_{t-1,J}}{\up_{t,J}}-1\right) \\
        = & \vert\A_1\vert + \sum_{t=2}^s \vert \A_t\setminus \A_{t-1}\vert + \frac{1}{e} \sum_{t=1}^s\sum_{E_J\in \A_t} \left(\frac{\up_{t-1,J}}{\up_{t,J}}-1\right). 
    \end{aligned}
\end{equation*}
Note that $\vert\A_1\vert + \sum_{t=2}^s \vert \A_t\setminus \A_{t-1}\vert$ is the total number of experts created till round $s$. 
\begin{equation*}
    \vert\A_1\vert + \sum_{t=2}^s \vert \A_t\setminus \A_{t-1}\vert \leq N
\end{equation*}
where $N$ denotes the number of experts created till round $s$. Therefore, the upper bound of $\ln X_s$ is
\begin{equation}\label{eqn:TUMA-Comp:meta-regret:upper-bound}
    \ln X_s = \ln \sum_{E_J\in\A_s} x_{s,J} \leq \ln \left(N+ \frac{1}{e} \sum_{t=1}^s\sum_{E_J\in \A_t} \left(\frac{\up_{t-1,J}}{\up_{t,J}}-1\right)\right).
\end{equation}
Combining the lower bound in \eqref{eqn:TUMA-Comp:meta-regret:lower-bound} and the upper bound in \eqref{eqn:TUMA-Comp:meta-regret:upper-bound}, we finish the proof.

\section{Conclusion and Future Work}\label{sec:conclusion}
In this paper, we develop a meta-expert framework for dual adaptive algorithms, where multiple experts are created dynamically and aggregated by a meta-algorithm. Specifically, we require the meta-algorithm to equip with a second-order bound, and utilize the linearized loss to evaluate the performance of experts. Based on this framework, we propose two kinds of universal algorithms to deal with changing environments, including two-layer algorithms where we increase the number of experts, and three-layer algorithms where we enhance experts' capabilities. In addition, our meta-expert framework can be extended to  online composite optimization. In the composite setting, we first introduce a novel universal algorithm for static regret of composite functions. By employing it as the expert-algorithm, we propose a universal algorithm that delivers strongly adaptive regret bounds for multiple types of convex functions. 

To equip our universal algorithms with dual adaptivity to function types and changing environments, they all maintain $O(\log^2 T)$ expert-algorithms (in the last layer) for a $T$-round online problem, which means that they need to conduct $O(\log^2 T)$ projections onto the feasible domain in each round. Such a large number of projections can be time-consuming in practical scenarios, especially when the  domain is complicated. Notice that recent developments in online learning utilize the black-box reduction \citep{pmlr-v75-cutkosky18a,pmlr-v119-cutkosky20a} to reduce the number of projections from $O(\log T)$ to $1$ per round in non-stationary OCO \citep{NeurIPS'22:efficient} and universal OCO \citep{NeurIPS:2024:Yang}. In the future, we will investigate whether this technique can be utilized to reduce the projection complexity of our methods.

\bibliography{ref}

\begin{thebibliography}{47}
\providecommand{\natexlab}[1]{#1}
\providecommand{\url}[1]{\texttt{#1}}
\expandafter\ifx\csname urlstyle\endcsname\relax
  \providecommand{\doi}[1]{doi: #1}\else
  \providecommand{\doi}{doi: \begingroup \urlstyle{rm}\Url}\fi

\bibitem[Abernethy et~al.(2008)Abernethy, Bartlett, Rakhlin, and Tewari]{Minimax:Online}
Jacob Abernethy, Peter~L. Bartlett, Alexander Rakhlin, and Ambuj Tewari.
\newblock Optimal strategies and minimax lower bounds for online convex games.
\newblock In \emph{Proceedings of the 21st Annual Conference on Learning Theory}, pages 415--423, 2008.

\bibitem[Adamskiy et~al.(2012)Adamskiy, Koolen, Chernov, and Vovk]{Adamskiy2012}
Dmitry Adamskiy, Wouter~M. Koolen, Alexey Chernov, and Vladimir Vovk.
\newblock A closer look at adaptive regret.
\newblock In \emph{Proceedings of the 23rd International Conference on Algorithmic Learning Theory}, pages 290--304, 2012.

\bibitem[Arora et~al.(2012)Arora, Hazan, and Kale]{v008a006}
Sanjeev Arora, Elad Hazan, and Satyen Kale.
\newblock The multiplicative weights update method: a meta-algorithm and applications.
\newblock \emph{Theory of Computing}, 8\penalty0 (6):\penalty0 121--164, 2012.

\bibitem[Bartlett et~al.(2008)Bartlett, Hazan, and Rakhlin]{NIPS2007_3319}
Peter~L. Bartlett, Elad Hazan, and Alexander Rakhlin.
\newblock Adaptive online gradient descent.
\newblock In \emph{Advances in Neural Information Processing Systems 20}, pages 65--72, 2008.

\bibitem[Boyd and Vandenberghe(2004)]{Convex-Optimization}
Stephen Boyd and Lieven Vandenberghe.
\newblock \emph{Convex Optimization}.
\newblock Cambridge University Press, 2004.

\bibitem[Cesa-Bianchi and Lugosi(2006)]{bianchi-2006-prediction}
Nicol\`{o} Cesa-Bianchi and G{\'a}bor Lugosi.
\newblock \emph{Prediction, Learning, and Games}.
\newblock Cambridge University Press, 2006.

\bibitem[Cutkosky(2020)]{pmlr-v119-cutkosky20a}
Ashok Cutkosky.
\newblock Parameter-free, dynamic, and strongly-adaptive online learning.
\newblock In \emph{Proceedings of the 37th International Conference on Machine Learning}, pages 2250--2259, 2020.

\bibitem[Cutkosky and Orabona(2018)]{pmlr-v75-cutkosky18a}
Ashok Cutkosky and Francesco Orabona.
\newblock Black-box reductions for parameter-free online learning in {Banach} spaces.
\newblock In \emph{Proceedings of the 31st Conference On Learning Theory}, pages 1493--1529, 2018.

\bibitem[Daniely et~al.(2015)Daniely, Gonen, and Shalev-Shwartz]{Adaptive:ICML:15}
Amit Daniely, Alon Gonen, and Shai Shalev-Shwartz.
\newblock Strongly adaptive online learning.
\newblock In \emph{Proceedings of the 32nd International Conference on Machine Learning}, pages 1405--1411, 2015.

\bibitem[Do et~al.(2009)Do, Le, and Foo]{icml2009_033}
Chuong Do, Quoc Le, and Chuan-Sheng Foo.
\newblock Proximal regularization for online and batch learning.
\newblock In \emph{Proceedings of the 26th International Conference on Machine Learning}, pages 257--264, 2009.

\bibitem[Duchi and Singer(2009)]{DBLP:journals/jmlr/DuchiS09}
John~C. Duchi and Yoram Singer.
\newblock Efficient online and batch learning using forward backward splitting.
\newblock \emph{Journal of Machine Learning Research}, 10:\penalty0 2899--2934, 2009.

\bibitem[Duchi et~al.(2010)Duchi, Shalev{-}Shwartz, Singer, and Tewari]{DBLP:conf/colt/DuchiSST10}
John~C. Duchi, Shai Shalev{-}Shwartz, Yoram Singer, and Ambuj Tewari.
\newblock Composite objective mirror descent.
\newblock In \emph{Proceedings of the 23rd Annual Conference on Learning Theory}, pages 14--26, 2010.

\bibitem[Freund et~al.(1997)Freund, Schapire, Singer, and Warmuth]{Freund:1997:UCP}
Yoav Freund, Robert~E. Schapire, Yoram Singer, and Manfred~K. Warmuth.
\newblock Using and combining predictors that specialize.
\newblock In \emph{Proceedings of the 29th Annual ACM Symposium on Theory of Computing}, pages 334--343, 1997.

\bibitem[Gaillard et~al.(2014)Gaillard, Stoltz, and van Erven]{pmlr-v35-gaillard14}
Pierre Gaillard, Gilles Stoltz, and Tim van Erven.
\newblock A second-order bound with excess losses.
\newblock In \emph{Proceedings of the 27th Conference on Learning Theory}, pages 176--196, 2014.

\bibitem[Gy\"{o}rgy et~al.(2012)Gy\"{o}rgy, Linder, and Lugosi]{Track_Large_Expert}
Andr\'{a}s Gy\"{o}rgy, Tam\'{a}s Linder, and G\'{a}bor Lugosi.
\newblock Efficient tracking of large classes of experts.
\newblock \emph{IEEE Transactions on Information Theory}, 58\penalty0 (11):\penalty0 6709--6725, 2012.

\bibitem[Hazan(2016)]{Intro:Online:Convex}
Elad Hazan.
\newblock Introduction to online convex optimization.
\newblock \emph{Foundations and Trends in Optimization}, 2\penalty0 (3-4):\penalty0 157--325, 2016.

\bibitem[Hazan and Seshadhri(2007)]{Adaptive:Hazan}
Elad Hazan and C.~Seshadhri.
\newblock Adaptive algorithms for online decision problems.
\newblock \emph{Electronic Colloquium on Computational Complexity}, 88, 2007.

\bibitem[Hazan and Seshadhri(2009)]{Hazan:2009:ELA}
Elad Hazan and C.~Seshadhri.
\newblock Efficient learning algorithms for changing environments.
\newblock In \emph{Proceedings of the 26th Annual International Conference on Machine Learning}, pages 393--400, 2009.

\bibitem[Hazan et~al.(2007)Hazan, Agarwal, and Kale]{ML:Hazan:2007}
Elad Hazan, Amit Agarwal, and Satyen Kale.
\newblock Logarithmic regret algorithms for online convex optimization.
\newblock \emph{Machine Learning}, 69\penalty0 (2-3):\penalty0 169--192, 2007.

\bibitem[Herbster and Warmuth(1998)]{Herbster1998}
Mark Herbster and Manfred~K. Warmuth.
\newblock Tracking the best expert.
\newblock \emph{Machine Learning}, 32\penalty0 (2):\penalty0 151--178, 1998.

\bibitem[Jun et~al.(2017)Jun, Orabona, Wright, and Willett]{Improved:Strongly:Adaptive}
Kwang-Sung Jun, Francesco Orabona, Stephen Wright, and Rebecca Willett.
\newblock Improved strongly adaptive online learning using coin betting.
\newblock In \emph{Proceedings of the 20th International Conference on Artificial Intelligence and Statistics}, pages 943--951, 2017.

\bibitem[Littlestone and Warmuth(1994)]{LITTLESTONE1994212}
Nick Littlestone and Manfred~K. Warmuth.
\newblock The weighted majority algorithm.
\newblock \emph{Information and Computation}, 108\penalty0 (2):\penalty0 212--261, 1994.

\bibitem[Luo and Schapire(2015)]{pmlr-v40-Luo15}
Haipeng Luo and Robert~E. Schapire.
\newblock Achieving all with no parameters: Adanormalhedge.
\newblock In \emph{Proceedings of the 28th Conference on Learning Theory}, pages 1286--1304, 2015.

\bibitem[Mhammedi et~al.(2019)Mhammedi, Koolen, and Van~Erven]{pmlr-v99-mhammedi19a}
Zakaria Mhammedi, Wouter~M Koolen, and Tim Van~Erven.
\newblock Lipschitz adaptivity with multiple learning rates in online learning.
\newblock In \emph{Proceedings of the 32nd Conference on Learning Theory}, pages 2490--2511, 2019.

\bibitem[Ordentlich and Cover(1998)]{Lower:bound:Portfolio}
Erik Ordentlich and Thomas~M. Cover.
\newblock The cost of achieving the best portfolio in hindsight.
\newblock \emph{Mathematics of Operations Research}, 23\penalty0 (4):\penalty0 960--982, 1998.

\bibitem[Shalev-Shwartz(2011)]{Online:suvery}
Shai Shalev-Shwartz.
\newblock Online learning and online convex optimization.
\newblock \emph{Foundations and Trends in Machine Learning}, 4\penalty0 (2):\penalty0 107--194, 2011.

\bibitem[Shalev-Shwartz et~al.(2007)Shalev-Shwartz, Singer, and Srebro]{ICML_Pegasos}
Shai Shalev-Shwartz, Yoram Singer, and Nathan Srebro.
\newblock Pegasos: primal estimated sub-gradient solver for {SVM}.
\newblock In \emph{Proceedings of the 24th International Conference on Machine Learning}, pages 807--814, 2007.

\bibitem[Tibshirani(1996)]{tibshirani1996regression}
Robert Tibshirani.
\newblock Regression shrinkage and selection via the lasso.
\newblock \emph{Journal of the Royal Statistical Society Series B: Statistical Methodology}, 58\penalty0 (1):\penalty0 267--288, 1996.

\bibitem[Toh and Yun(2010)]{toh2010accelerated}
Kim-Chuan Toh and Sangwoon Yun.
\newblock An accelerated proximal gradient algorithm for nuclear norm regularized linear least squares problems.
\newblock \emph{Pacific Journal of optimization}, 6\penalty0 (3):\penalty0 615--640, 2010.

\bibitem[van Erven and Koolen(2016)]{NIPS2016_6268}
Tim van Erven and Wouter~M Koolen.
\newblock {MetaGrad}: Multiple learning rates in online learning.
\newblock In \emph{Advances in Neural Information Processing Systems 29}, pages 3666--3674, 2016.

\bibitem[van Erven et~al.(2021)van Erven, Koolen, and van~der Hoeven]{JMLR:v22:20-1444}
Tim van Erven, Wouter~M. Koolen, and Dirk van~der Hoeven.
\newblock {MetaGrad}: Adaptation using multiple learning rates in online learning.
\newblock \emph{Journal of Machine Learning Research}, 22\penalty0 (161):\penalty0 1--61, 2021.

\bibitem[Veness et~al.(2013)Veness, White, Bowling, and Gy\"{o}rgy]{6543068}
Joel Veness, Martha White, Michael Bowling, and Andr\'{a}s Gy\"{o}rgy.
\newblock Partition tree weighting.
\newblock In \emph{Proceedings of the 2013 Data Compression Conference}, pages 321--330, 2013.

\bibitem[Wan et~al.(2021)Wan, Tu, and Zhang]{Adaptive:Short}
Yuanyu Wan, Wei-Wei Tu, and Lijun Zhang.
\newblock Strongly adaptive online learning over partial intervals.
\newblock \emph{Science China Information Sciences}, 2021.

\bibitem[Wang et~al.(2018)Wang, Zhao, and Zhang]{Adaptive:One:Gradient}
Guanghui Wang, Dakuan Zhao, and Lijun Zhang.
\newblock Minimizing adaptive regret with one gradient per iteration.
\newblock In \emph{Proceedings of the 27th International Joint Conference on Artificial Intelligence}, pages 2762--2768, 2018.

\bibitem[Wang et~al.(2019)Wang, Lu, and Zhang]{Adaptive:Maler}
Guanghui Wang, Shiyin Lu, and Lijun Zhang.
\newblock Adaptivity and optimality: A universal algorithm for online convex optimization.
\newblock In \emph{Proceedings of the 35th Conference on Uncertainty in Artificial Intelligence}, pages 659--668, 2019.

\bibitem[Wang et~al.(2020)Wang, Lu, Hu, and Zhang]{AAAI:2020:Wang}
Guanghui Wang, Shiyin Lu, Yao Hu, and Lijun Zhang.
\newblock Adapting to smoothness: A more universal algorithm for online convex optimization.
\newblock In \emph{Proceedings of the 34th AAAI Conference on Artificial Intelligence}, pages 6162--6169, 2020.

\bibitem[Wei et~al.(2016)Wei, Hong, and Lu]{NIPS:2016:Wei}
Chen{-}Yu Wei, Yi{-}Te Hong, and Chi{-}Jen Lu.
\newblock Tracking the best expert in non-stationary stochastic environments.
\newblock In \emph{Advances in Neural Information Processing Systems 29}, pages 3972--3980, 2016.

\bibitem[Xiao(2009)]{DBLP:conf/nips/Xiao09}
Lin Xiao.
\newblock Dual averaging method for regularized stochastic learning and online optimization.
\newblock In Yoshua Bengio, Dale Schuurmans, John~D. Lafferty, Christopher K.~I. Williams, and Aron Culotta, editors, \emph{Advances in Neural Information Processing Systems 22}, pages 2116--2124, 2009.

\bibitem[Yang et~al.(2018)Yang, Li, and Zhang]{AISTATS:2018:Yang}
Tianbao Yang, Zhe Li, and Lijun Zhang.
\newblock A simple analysis for exp-concave empirical minimization with arbitrary convex regularizer.
\newblock In \emph{Proceedings of the 21st International Conference on Artificial Intelligence and Statistics (AISTATS)}, pages 445--453, 2018.

\bibitem[Yang et~al.(2024{\natexlab{a}})Yang, Wang, Zhao, and Zhang]{NeurIPS:2024:Yang}
Wenhao Yang, Yibo Wang, Peng Zhao, and Lijun Zhang.
\newblock Universal online convex optimization with $1 $ projection per round.
\newblock In \emph{Advances in Neural Information Processing Systems 35}, 2024{\natexlab{a}}.

\bibitem[Yang et~al.(2024{\natexlab{b}})Yang, Tian, Cheng, Wan, and Song]{DBLP:journals/ijon/YangTCWS24}
Xu~Yang, Peng Tian, Xiao Cheng, Yuanyu Wan, and Mingli Song.
\newblock Regularized online exponentially concave optimization.
\newblock \emph{Neurocomputing}, 595:\penalty0 1--8, 2024{\natexlab{b}}.

\bibitem[Zhang et~al.(2018)Zhang, Yang, Jin, and Zhou]{Dynamic:Regret:Adaptive}
Lijun Zhang, Tianbao Yang, Rong Jin, and Zhi-Hua Zhou.
\newblock Dynamic regret of strongly adaptive methods.
\newblock In \emph{Proceedings of the 35th International Conference on Machine Learning}, pages 5882--5891, 2018.

\bibitem[Zhang et~al.(2021)Zhang, Wang, Tu, Jiang, and Zhou]{NEURIPS2021_d1588e68}
Lijun Zhang, Guanghui Wang, Wei-Wei Tu, Wei Jiang, and Zhi-Hua Zhou.
\newblock Dual adaptivity: A universal algorithm for minimizing the adaptive regret of convex functions.
\newblock In \emph{Advances in Neural Information Processing Systems}, pages 24968--24980, 2021.

\bibitem[Zhang et~al.(2022)Zhang, Wang, Yi, and Yang]{ICML:2022:Zhang}
Lijun Zhang, Guanghui Wang, Jinfeng Yi, and Tianbao Yang.
\newblock A simple yet universal strategy for online convex optimization.
\newblock In \emph{Proceedings of the 39th International Conference on Machine Learning}, pages 26605--26623, 2022.

\bibitem[Zhang et~al.(2024)Zhang, Wang, Wang, Yi, and Yang]{zhang2024universal}
Lijun Zhang, Yibo Wang, Guanghui Wang, Jinfeng Yi, and Tianbao Yang.
\newblock Universal online convex optimization meets second-order bounds.
\newblock \emph{arXiv preprint arXiv:2105.03681}, 2024.

\bibitem[Zhao et~al.(2022)Zhao, Xie, Zhang, and Zhou]{NeurIPS'22:efficient}
Peng Zhao, Yan-Feng Xie, Lijun Zhang, and Zhi-Hua Zhou.
\newblock Efficient methods for non-stationary online learning.
\newblock In \emph{Advances in Neural Information Processing Systems 35}, pages 11573--11585, 2022.

\bibitem[Zinkevich(2003)]{zinkevich-2003-online}
Martin Zinkevich.
\newblock Online convex programming and generalized infinitesimal gradient ascent.
\newblock In \emph{Proceedings of the 20th International Conference on Machine Learning}, pages 928--936, 2003.

\end{thebibliography}

\end{document}